\documentclass[final,3p,times]{elsarticle}
\usepackage{graphicx}
\usepackage[caption=false]{subfig}
\usepackage{amsmath,bm}
\usepackage{amsthm,amsmath,amssymb} \usepackage{mathrsfs}
\usepackage{url}
\usepackage{lineno}
\usepackage{color}
\usepackage{algorithmic}
\usepackage{algorithm}
\usepackage{diagbox}
\usepackage{makecell}
\usepackage{booktabs}
\usepackage{multirow}
\usepackage[figuresright]{rotating}
\usepackage[normalem]{ulem} 
\usepackage{siunitx} 
\usepackage{xcolor}
\usepackage{colortbl} 
\usepackage{hyperref}

\definecolor{bestblue}{RGB}{0, 119, 187}
\definecolor{rowgray}{gray}{0.92} 
\definecolor{impgreen}{RGB}{0, 150, 0} 

\newcommand{\best}[1]{\textbf{\tablenum[table-format=1.2e-1]{#1}}}
\newcommand{\second}[1]{\color{bestblue}\bfseries\tablenum[table-format=1.2e-1]{#1}}
\newcommand{\third}[1]{\uline{\tablenum[table-format=1.2e-1]{#1}}}

\newcommand{\imp}[1]{\multicolumn{1}{c}{\cellcolor{impgreen!10}\bfseries\color{impgreen!80!black}$\uparrow$#1\%}}
\newcommand{\noimp}{\multicolumn{1}{c}{-}}

\newcommand{\na}{\multicolumn{1}{c}{--}}

\newtheorem{definition}{Definition}%

\journal{Journal of Computational Physics}

\begin{document}

\begin{frontmatter}

  \title{DyMixOp: A Neural Operator Designed from a Complex Dynamics Perspective with Local-Global Mixing for Solving PDEs}

  \author[aff1]{Pengyu Lai}
  \author[aff1]{Yixiao Chen}
  \author[aff1]{Dewu Yang}
  \author[aff1]{Rui Wang}
  \author[aff1]{Feng Wang}
  \author[aff1]{Hui Xu\corref{cor1}}

  \cortext[cor1]{Corresponding author: dr.hxu@sjtu.edu.cn}

  \affiliation[aff1]{%
    organization={School of Aeronautics and Astronautics, Shanghai Jiao Tong University},
    city={Shanghai},
    postcode={200240},
    country={China}
  }

  \begin{abstract}
    A primary challenge in using neural networks to approximate nonlinear dynamical systems governed by partial differential equations (PDEs) lies in recasting these systems into a tractable representation—particularly when the dynamics are inherently non-linearizable or require infinite-dimensional spaces for linearization.
    To address this challenge, we introduce DyMixOp, a novel neural operator framework for PDEs that integrates theoretical insights from complex dynamical systems. Grounded in dynamics-aware priors and inertial manifold theory, DyMixOp projects the original infinite-dimensional PDE dynamics onto a finite-dimensional latent space. This reduction preserves both essential linear structures and dominant nonlinear interactions, thereby establishing a physically interpretable and computationally structured foundation.
    Central to this approach is the local–global mixing (LGM) transformation, a key architectural innovation inspired by the convective nonlinearity in turbulent flows (e.g., $u\cdot\nabla u$). By multiplicatively coupling local fine-scale features with global spectral information, LGM effectively captures high-frequency details and complex nonlinear couplings while mitigating the spectral bias that plagues many existing neural operators.
    The framework is further enhanced by a dynamics-informed architecture that stacks multiple LGM layers in a hybrid configuration, incorporating timescale-adaptive gating and parallel aggregation of intermediate dynamics. This design enables robust approximation of general evolutionary dynamics across diverse physical regimes.
    Extensive experiments on seven benchmark PDE systems—spanning 1D to 3D, elliptic to hyperbolic types—demonstrate that DyMixOp achieves state-of-the-art performance on six of them, significantly reducing prediction errors (by up to 94.3$\%$ in chaotic regimes) while maintaining computational efficiency and strong scalability.
  \end{abstract}


  \begin{highlights}

  \item Introduces DyMixOp, a novel neural operator grounded in inertial manifold theory and complex dynamics priors, which effectively reduces infinite-dimensional PDE dynamics to a finite-dimensional latent space while preserving essential nonlinear interactions and enhancing physical interpretability.
  \item Proposes the local-global mixing (LGM) transformation, inspired by the convective nonlinearity in turbulence (e.g., $u\cdot\nabla u$), that multiplicatively fuses local fine-scale features with global spectral information to mitigate spectral bias and recover high-frequency components commonly lost in existing neural operators.
  \item Demonstrates state-of-the-art performance across six diverse PDE benchmarks (1D–3D, elliptic to hyperbolic), achieving significant error reductions (up to 94.3\% on chaotic systems) while maintaining computational efficiency and scalability---establishing DyMixOp as a robust, general-purpose framework for data-driven PDE solving.

  \end{highlights}

  \begin{keyword}
    Complex Systems \sep Neural Operator \sep Nonlinear Dynamics \sep PDE Solving \sep Inertial Manifold
  \end{keyword}

\end{frontmatter}


\section{Introduction}\label{sec1}
Partial differential equations (PDEs) are the mathematical backbone for describing chaotic behaviors and understanding underlying mechanics in dynamical systems. They are distributed in various fields including climate\cite{bi2023accurate,lam2023learning}, molecular dynamics\cite{rapaport2004art}, ecological modeling\cite{blasius1999complex}, brain activity\cite{breakspear2017dynamic}, chemistry\cite{jensen2017introduction}, heat transfer\cite{howell2020thermal} and turbulent flows\cite{mukherjee2023intermittency}. Predicting the dynamics of complex systems by solving PDEs is crucial for scientific and engineering applications. As a result, a variety of numerical methods have been developed\cite{dennis1996numerical, moin1998direct}, including the finite difference method\cite{smith1985numerical}, finite volume method\cite{versteeg2007introduction}, lattice Boltzmann method\cite{succi2001lattice} and finite element method\cite{zienkiewicz1971finite}. Despite the high fidelity of simulations produced by traditional approaches, they become inefficient with frequent recalculations whenever initial conditions or equation parameters are altered. For example, during the initial design phase, numerous experiments need to be conducted to optimize the design. This greatly raises computational costs and significantly increases the time required. When faced with practical challenges such as noisy, damaged, or low-fidelity initial conditions, it is difficult to produce high-quality solutions using solely traditional techniques. Additionally, these conventional methods are not equipped to handle situations where PDEs are partially known or entirely unknown.

In recent years, data-driven methods have thrived across various disciplines, with neural networks emerging as powerful tools to overcome the limitations of traditional numerical methods \cite{jordan2015machine}.
Neural networks aim to approximate the mapping between input variables and solution fields, either as coordinate-based approximators which directly map spatiotemporal coordinates to solution values \cite{yu2018deep, raissi2019physics}, surrogate evolutionary operators which predict future states based on historical data and effectively replace traditional time-integration \cite{li2020fourier, sanchez2020learning}, and auxiliary components which infer physical parameters or correction terms to enhance classical solvers \cite{duraisamy2019turbulence, kochkov2021machine, ling2016reynolds, buaria2023forecasting}.
However, the rapidly expanding zoo of network architectures (e.g., CNNs, Transformers, UNets) often obscures the fundamental mathematical operations at play. To address this, we revisit these methodologies through a unified perspective of integral transformations. Rather than focusing on network topology, we classify existing approaches based on the domain of their integral operator—specifically, whether the transformation spans the entire input domain or is restricted to a local neighborhood.

The general transformation of a single neural layer or block can be defined as an integral operator in the form of:
\begin{equation}\label{eq:general_transformation}
  \left(\mathcal{G}_\theta v\right)(x)=\int_{D_\tau} g_\theta(x, \tau) v(\tau) d \tau, \quad x \in D_x
\end{equation}
where $\mathcal{G}_\theta: \mathcal{F}(D_\tau, \mathbb{R}^{d_u}) \rightarrow \mathcal{F}(D_x, \mathbb{R}^{d_v})$ denotes the transformation parameterized by $\theta \in \Theta$, and $g_\theta: D_x \times D_\tau \rightarrow \mathbb{R}^{d_u \times d_v}$ represents the learnable integral kernel. Here, $D_\tau$ and $D_x$ are bounded domains in $\mathbb{R}^d$ ($d\in\mathbb{Z}^+$ denotes the spatial dimension), representing the input and output physical domains, respectively.
Let $E_\tau \subset \mathbb{R}^d$ denote the entire physical domain and $P_{\tau} \subset E_\tau$ denote a partial local domain (a proper subset). The classification is determined by the integration domain $D_\tau$. It is a \textbf{global transformation} when $D_\tau = E_\tau$, meaning the output at $x$ depends on the global input state. It is a \textbf{local transformation} when $D_\tau = P_\tau$ (typically a neighborhood around $x$), meaning the output depends only on local features.
To describe composite interaction, we define the interaction mechanisms-\textbf{adding} and \textbf{mixing}-between transformations, where mixing refers to the element-wise product (Hadamard product) of two transformations:
\begin{equation}
  \left(\mathcal{G}^{mix}_\theta v\right)(x) = \left(\int_{D_\tau} g_\theta(x, \tau) v(\tau) d \tau\right) \odot \left(\int_{\bar{D}_\tau} \bar{g}_{\theta}(x, \tau) v(\tau) d \tau\right), \quad x \in D_x,
\end{equation}
and adding refers to the element-wise summation of two transformations:
\begin{equation}
  \left(\mathcal{G}^{add}_\theta v\right)(x) = \int_{D_\tau} g_\theta(x, \tau) v(\tau) d \tau + \int_{\bar{D}_\tau} \bar{g}_{\theta}(x, \tau) v(\tau) d \tau, \quad x \in D_x
\end{equation}
where $\bar{g}_{\theta}$ and $\bar{D}_\tau$ represent a distinct kernel and domain from $g_\theta$ and $D_\tau$. For instance, in an LGA transformation, one domain would be local ($P_\tau$) and the other global ($E_\tau$).

The implementation of neural networks for solving ordinary differential equations (ODEs) and partial differential equations (PDEs) dates back two decades. In the 1990s, Psichogios et al. \cite{psichogios1992hybrid} utilized a shallow fully connected neural network (FCNN), combined with first-principles modeling for the partial terms of ODEs, to estimate unmeasured process parameters. Subsequently, Lagaris et al. \cite{lagaris1998artificial} trained a shallow FCNN to satisfy differential equations, pioneering a method in comparison to traditional numerical techniques. With the recent revival of deep learning \cite{lecun2015deep}, Raissi et al. \cite{raissi2019physics} extended these ideas using modern deep learning techniques, proposing physics-informed neural networks (PINNs) to solve forward and inverse problems in complex PDEs. This work demonstrated the remarkable potential of neural networks in solving PDE problems and positioned neural networks at the center of the scientific computing community \cite{pang2019fpinns, lu2021physics, karniadakis2021physics}. However, due to the point-wise connections in FCNNs, applying them to high-dimensional inputs (arising from computational domain discretization) results in an extensive number of parameters. To maintain computational feasibility, the aforementioned studies limited inputs to a few points at a time and treated the number of points as the batch size. Consequently, as a classical architecture, these instances of FCNNs fall within the scope of local transformations.

To address the curse of dimensionality in image-based inputs, convolutional neural networks (CNNs) were subsequently proposed. CNNs utilize convolutional kernels to extract local structures by considering the aggregation of neighborhood information and sharing parameters across the entire input domain. Since image inputs can represent discrete solutions in PDEs, many studies have employed CNNs to approximate evolution operators \cite{qu2022learning, gao2021phygeonet, list2022learned}. Various convolutional architectures have also been developed to solve PDE problems, including generative adversarial networks for two-dimensional turbulence\cite{kim2024prediction, kim2021unsupervised}, and autoencoder for three-dimensional turbulence\cite{xuan2023reconstruction}.
Naturally, these convolutional architectures are categorized as local transformations. Considering the effectiveness of residual connections in deep networks \cite{he2016deep}, adding mechanisms are incorporated into most architectures to further improve their performance \cite{han2018solving}. Therefore, the modern local-transformations-featured architectures almost locates at \textbf{local-local adding} (LLA) transformation.
However, with the rapid evolution of deep learning, more complex architectures have emerged, which explored the interaction mechanism between convolutional transformation. These methods aim for \textbf{local-local-mixing} (LLM) transformations to obtain more powerful nonlinear representations. For instance, Shi et al. combined convolutional layers with Long Short-Term Memory networks (LSTMs) for precipitation nowcasting problems, where the cell and latent states are updated via element-wise products \cite{shi2015convolutional}. Long et al. integrated traditional numerical schemes with LLM transformations to approximate high-order differential operators \cite{long2018pde}. Similarly, Rao et al. developed a framework to encode the physical law and traditional numerical schemes, where the nonlinear terms in PDEs are approximated through LLM transformations \cite{rao2023encoding}.
Nevertheless, pure convolutional transformation heavily depend on mesh discretization, effectively acting as approximators for mesh-specific kernels. Consequently, changes in the mesh resolution or geometry typically lead to a significant degradation in performance, often falling short of acceptable accuracy.

To overcome the inherent mesh dependency of convolutional architectures, research converged on the development of neural operators. The fundamental objective is to approximate a kernel function $g$ that is globally consistent across arbitrary spatial coordinates, thereby learning a mapping between infinite-dimensional function spaces rather than finite-dimensional Euclidean vectors.
Two primary strategies emerged in neural operators to achieve this objective. The first involves designing networks in which the discretization of the kernel is decoupled from the underlying mesh, such as point-cloud-based MLPs or one-size-kernel convolutions. The second, more robust approach approximates the kernel function within an auxiliary latent space (e.g., a spectral/frequency domain).
As a seminal implementation, Li et al. introduced the graph neural operator (GNO) \cite{li2020neural, li2020multipole}. GNO parameterized the kernel as a continuous function (typically an MLP) and formulated the operator as a Nyström-discretized integral over a graph structure. By evaluating the kernel on the relative coordinates of graph edges, GNO theoretically achieved discretization invariance. To obtain global receptive fields with computational efficiency, subsequent research shifted toward spectral methods. In the spectral domain, complex global convolutions are transformed into point-wise multiplications, leveraging the fact that spectral modes provide a compact representation of the operator's eigenfunctions \cite{trefethen2000spectral}.
Building on this advantage, Li et al. further proposed the Fourier neural operator (FNO) \cite{li2020fourier}, which replaces graph-based message passing with a Fourier integral layer. By parameterizing the kernel in the frequency domain and truncating high-frequency modes, FNO achieves a global transformation that is both computationally efficient and highly generalizable across various physical regimes \cite{azizzadenesheli2024neural}. This spectral paradigm inspired a family of operators, including wavelet neural operators \cite{tripura2022wavelet, gupta2021multiwavelet} and the Laplace neural operator \cite{cao2024laplace}, which utilized different basis functions to project the kernel into corresponding transform spaces.
Despite the efficiency of global transformations, truncating high-frequency modes intensifies the spectral bias in neural networks \cite{rahaman2019spectral}.
Spectral operator architectures such as FNO and its variants often combine global spectral transformations with a parallel one-size-kernel convolution in an additive way to mitigate this issue and could therefore be classified as \textbf{local-global adding} (LGA) transformations, where local and high-frequency features are partially preserved in local transformations while global dependencies are resolved in the spectral global transformation.

Even though one-size-kernel convolutions were adopted in above LGA transformations, the powerful local feature extraction capabilities of large-kernel convolutions are appealing. They can aggregate the information from neighbors to accurately establish the interaction mechanism between fine-scale dynamical structures. To this end, several approaches have been proposed to achieve mesh-independent local transformations. Ocampo et al. \cite{ocampo2022scalable} introduced discrete-continuous (DISCO) convolutions, which treated the input discretely while keeping the convolution kernel continuous, enabling scalable and equivariant operations across arbitrary discretizations. Raonic et al. \cite{raonic2023convolutional} proposed the convolutional neural operator (CNO), designed to preserve continuous-discrete equivalence (CDE) by strictly enforcing aliasing-free constraints, thereby allowing standard CNN architectures to function as resolution-independent operators. Bartolucci et al. \cite{bartolucci2023representation} established the theoretical framework of representation equivalent neural operators (ReNO) to characterize and mitigate operator aliasing, ensuring consistency between continuous operators and their discrete realizations. Liu-Schiaffini et al. integrated localized integral and differential kernels into the FNO architecture \cite{liu2024neural}, which preserved operator-theoretic consistency across resolutions. Furthermore, Gao et al. \cite{gao2025discretizationinvariance} identified the discretization mismatch error (DME) inherent in multi-resolution training and proposed the cross-resolution operator-learning pipeline (CROP), which leverages a band-limited latent space to robustly capture local features while maintaining invariance to input resolution.
These architectures promote the local transformation and extend one-size convolutional kernel to large-size kernel while maintain the mesh invariance.

Beyond spectral methods, the rapid evolution of large language models has introduced Transformer \cite{vaswani2017attention} and Mamba \cite{gu2024mamba} architectures to the field of operator learning. When viewed through the lens of integral transformations, these models naturally exemplify the LGA transformations. A standard block in these architectures combines a global transformation, such as self-attention or selective state spaces, with a local transformation, such as point-wise feed-forward networks. This classification is theoretically grounded by that self-attention has been demonstrated to approximate kernel integration \cite{kovachki2023neural}, while the theoretical equivalence between Mamba and structured attention \cite{dao2024transformers} confirms their shared capacity for global information aggregation.
Leveraging this structure, Transformer-based frameworks have evolved from theoretical adaptations to robust geometric solvers. Cao first bridged the gap by formulating linear attention as a Galerkin projection \cite{cao2021choose}. Subsequent studies focused on handling complex physical domains, for illustrations, Li et al. developed OFormer to predict spatiotemporal dynamics \cite{li2022transformer}, Hao et al. proposed GNOT to introduce geometric gating for processing multiple inputs in support of a novel attention mechanism \cite{hao2023gnot}, and Wu et al. proposed Transolver to achieve efficiency on general geometries via low-rank decomposition with physically-aware tokens \cite{wu2024transolver}. To overcome the quadratic complexity of attention, recent research has pivoted to Mamba-based frameworks, which offer linear-time sequence modeling. These approaches have been successfully adapted for PDEs, including spatiotemporal modeling in DeepOMamba \cite{hu2025deepomamba}, efficiency enhancements via latent space projections \cite{tiwari2025latent}, and extensions to non-Euclidean unstructured grids in GeoMaNO \cite{han2025geomano}.

\begin{figure}[htbp]
  \centering
  \includegraphics[width=\linewidth]{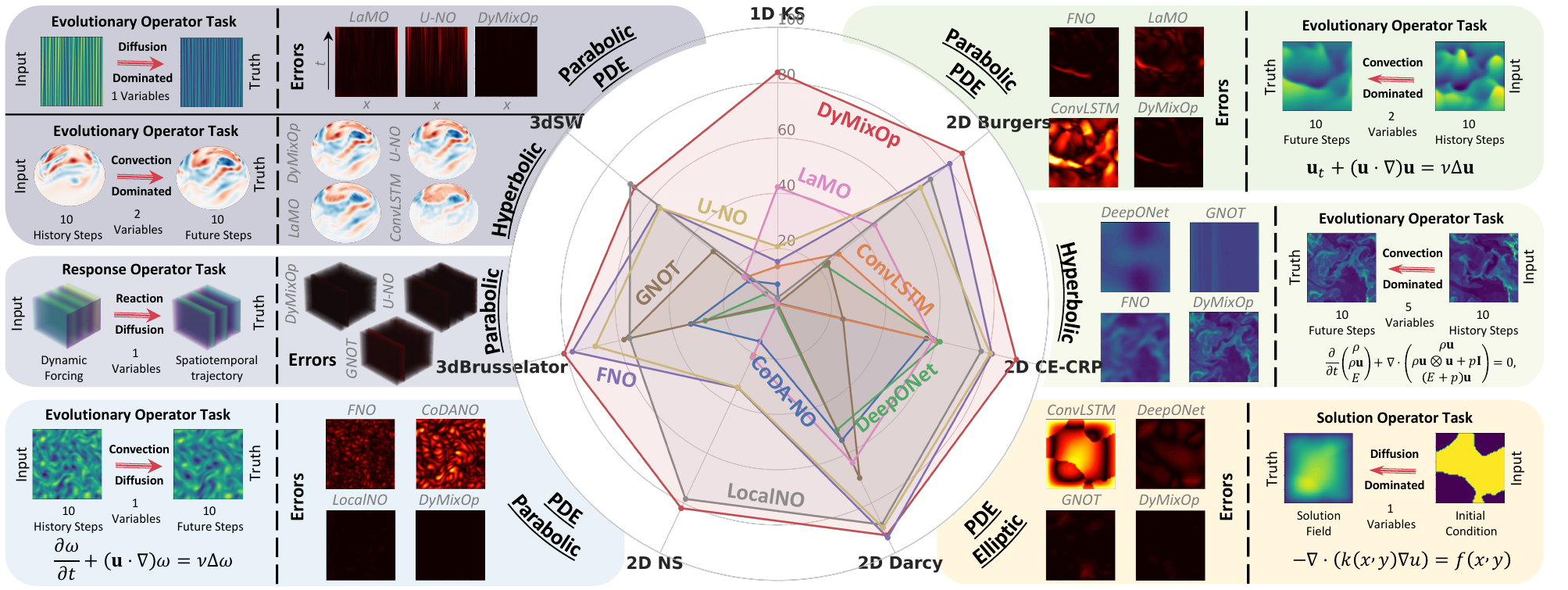}
  \caption{Overview of DyMixOp’s performance across seven PDE benchmarks. Each axis corresponds to a representative task, covering elliptic, parabolic, and hyperbolic systems. DyMixOp demonstrates consistently strong performance across heterogeneous physical regimes, highlighting its robustness and scalability compared to existing neural operator baselines.}
  \label{fig:radar_overview}
\end{figure}

Despite their success, global transformations suffer from a critical limitation in capturing intricate interactions and fine-scale dynamics.
To mitigate computational complexity and reduce parameter counts, these architectures typically rely on dimensionality reduction strategies, such as truncating high-frequency modes in spectral methods or partitioning the domain into coarse patches in Transformer- and Mamba-based models.
This compression inevitably attenuates fine-scale information and nonlinear couplings. Although LGA transformations attempt to compensate for this loss through a parallel local pathway, simple linear summation is often insufficient to reconstruct the complex, nonlinear high-frequency interactions that are essential for accurately modeling chaotic and multiscale systems.
As a result, existing approaches frequently produce over-smoothed solutions and struggle to maintain consistent performance across diverse PDE regimes.
Motivated by this limitation, we seek a neural operator that can preserve fine-scale expressiveness while remaining robust and scalable across heterogeneous physical systems.
To this end, we propose \textbf{DyMixOp}, a dynamics-aware neural operator designed to systematically integrate local and global interactions within a unified framework.
Fig. \ref{fig:radar_overview} provides an overview of DyMixOp’s performance across seven representative PDE benchmarks ranging from one to three dimensions, spanning elliptic, parabolic, and hyperbolic systems.
Rather than optimizing for a single task, DyMixOp exhibits a markedly improved balance between global coherence and fine-scale fidelity when compared with existing neural operators.
The physical and dynamical principles underlying DyMixOp are discussed in the following motivation section.

\section{Motivation}
Complex physical phenomena inherent in dynamical systems have long inspired the construction of novel neural network architectures, often yielding superior performance. For instance, the design of the Hopfield neural network draws directly from the physics of magnetic spin systems and particle interactions, specifically leveraging the energy minimization principles of magnetic forces \cite{hopfield1982neural}. Similarly, the renowned generative diffusion models are grounded in non-equilibrium statistical physics, mimicking the process where data structure is systematically degraded via iterative forward diffusion \cite{sohl2015deep, ho2020denoising}. These successes suggest that tapping into the mechanics of intricate dynamical systems could offer a pathway to enhance global transformations and resolve their current limitations.

In this pursuit, we turn to the most sophisticated of dynamical systems: turbulence. As the Nobel laureate Richard Feynman famously remarked, turbulence remains "the most important unsolved problem of classical physics" \cite{feynman2015feynman}. Turbulence presents an optimal paradigm for harnessing multi-scale processes. Central to these processes is the \textbf{convective term}, which drives the energy cascade, transferring energy from large-scale structures to smaller scales. This mechanism facilitates the self-organization of turbulence, creating intricate vortex structures and giving rise to diverse dynamic behaviors such as intermittency, vorticity stretching, and compression. Indeed, this principle has already inspired recent research into approximating the mapping from large-scale to small-scale dynamics \cite{lai2024neural}.

Drawing inspiration from the mechanistic role of the convective term, we propose an innovative transformation designed to possess powerful and diverse expressive capabilities: the \textbf{local-global mixing} (LGM) transformation. Mathematically, the convective term comprises the nonlinear product of the velocity field and its gradient (e.g., $u \cdot \nabla u$). In the context of numerical approximation, the velocity field $u$ is typically defined point-wise, representing the local state. Conversely, accurately capturing the gradient $\nabla u$, particularly in pseudospectral methods, requires global information (differentiation in the frequency domain) to resolve derivatives across the domain. From this perspective, the physical act of convection can be reinterpreted as a structural interaction: the multiplicative mixing of a local quantity and a global quantity. This LGM transformation offers distinct advantages by merging the computational efficiency of spectral global transformations with the capacity to recover high-frequency components (HFCs) through nonlinear mixing with local features.
In the Appendix A.1$\sim$2, we provide a formal theoretical proof demonstrating that the lack of capability representing HFCs in global transformations and the mixing interaction expands the spectral support effectively, alongside simple numerical experiments validating the lack and recovery of high-frequency modes.
As a novel development, LGM offers a fresh approach to leveraging the multi-scale and nonlinear characteristics of convection, significantly enhancing the expressive power of the model and potentially serving as a foundation for the next generation of neural operators.

\section{Methodology}\label{Methodology}
\subsection{Operator Approximation}
In this work, we focus on the form of neural operators due to their mesh independence and generalization to unknownn parameters. Here, we first outline the concept of operator approximation.
Given a spatial domain \( D \subset \mathbb{R}^d \) with a spatial dimension \( d \), an operator mapping from the input space to the solution space is defined as:
\begin{equation}
  \mathcal{G}: \mathcal{I}(D; \mathbb{R}^{d_i}) \rightarrow \mathcal{U}(D; \mathbb{R}^{d_u}),
\end{equation}
where \( \mathcal{I}(D; \mathbb{R}^{d_i}) \) and \( \mathcal{U}(D; \mathbb{R}^{d_u}) \) are Banach spaces representing the input and solution spaces, respectively, and \( d_i, d_u \in \mathbb{N} \) denote their corresponding dimensions. Depending on the type of input, as referenced in the Introduction, the operator \( \mathcal{G} \) can be interpreted as either an evolutionary operator or a solution operator. In this work, we specifically focus on approximating the evolutionary operator, which adjusts the operator mapping to:
\begin{equation}
  \mathcal{G}: \mathcal{U}(D; \mathbb{R}^{d_u}) \rightarrow \mathcal{U}(D; \mathbb{R}^{d_u}).
\end{equation}
In the context of operator semi-groups, the evolutionary operator exhibits the semi-group property. For times \( s, t \geq 0 \), it can be expressed as:
\begin{equation}
  \mathcal{G}^s(u(t)) = \mathcal{G}^s \circ \mathcal{G}^t(u(0)) = \mathcal{G}^{s+t}(u(0)),
\end{equation}
where \( u(t) \in \mathcal{U}(D; \mathbb{R}^{d_u}) \) represents the state of the system at time \( t \), and \( u(0) \) is the initial state. This formulation highlights the key property of the evolutionary operator: \( \mathcal{G}^{s+t} = \mathcal{G}^s \circ \mathcal{G}^t \), which is essential for modeling the time evolution of the system.

Computationally, the operator \(\mathcal{G}\) must be discretized, resulting in data that resides in a finite-dimensional space. Specifically, consider a discrete time sequence \(\{t_i\}_{i=0}^T\).
Given states \(u^{t_i},u^{t_{i+1}} \in \mathcal{U}(D;\mathbb{R}^{d_u})\), sampled from a probability distribution \(P_u\) over \(\mathcal{U}(D;\mathbb{R}^{d_u})\), their relationship is defined as \(u^{t_{i+1}} = \mathcal{G}(u^{t_i})\) for \(i = 0, 1, \dots, T-1\).
The objective of neural operators is to approximate the evolutionary operator \(\mathcal{G}\) by designing a neural network architecture parameterized by \(\theta \in \Theta\), where \(\Theta\) is a parameter space whose dimensionality depends on the chosen architecture. Thus, the construction of the neural operator is formulated as the following optimization problem:
\begin{equation}
  \min_{\theta \in \Theta} \mathbb{E}_{(u^{t_i}, u^{t_{i+1}}) \sim P_u} \left\| \mathcal{G}^{\dagger}(u^{t_i}; \theta) - u^{t_{i+1}} \right\|_{\mathcal{U}},
\end{equation}
where \(\mathcal{G}^{\dagger}(u^{t_i}; \theta)\) represents the neural operator's prediction of \(u^{t_{i+1}}\), and \(\|\cdot\|_{\mathcal{U}}\) denotes the norm in the Banach space \(\mathcal{U}(D;\mathbb{R}^{d_u})\).
Universal approximation theorems (UATs) \cite{hornik1989multilayer} suggest that neural networks can theoretically approximate any nonlinear function to arbitrary precision. However, while UATs affirm this capability, they provide no practical guidance on designing, training, or deploying these networks for real-world applications.
In this work, we leverage data pairs of the form \(\{[u^{t_{i-k}}, \dots, u^{t_i}], u^{t_{i+1}}\}\) to account for temporal dependencies. The optimization problem is solved using empirical risk minimization, approximating the expectation with a finite dataset to effectively train the neural evolutionary operator.

\subsection{Dynamics on Finite-dimensional Space}
We consider an autonomous dynamical system of the form
\begin{equation}
  \frac{\partial u(t)}{\partial t} = F(u), \quad t \in [0, T],
\end{equation}
where $u \in \mathcal{U}(D; \mathbb{R}^{d_u})$ denotes the system state and
$F:\mathcal{U}(D; \mathbb{R}^{d_u}) \to \mathcal{U}(D; \mathbb{R}^{d_u})$ is a (generally nonlinear) operator governing the temporal evolution of $u$.
Depending on the underlying physical system, $F$ may consist of linear operators, nonlinear interactions, and external forcing terms.
Modeling and approximating such nonlinear dynamical systems is challenging due to the coupling between state variables and complex spatiotemporal interactions.

\subsubsection{Variable-dimension Lifting and Intrinsic Dimension}
In the perspective of state variables (or channels in neural networks), a common strategy is to reformulate the system in an alternative representation where the dynamics become more tractable.
This idea underlies a broad class of approaches, including Koopman operator theory \cite{lasota2013chaos}, normal form theory \cite{poincare1893methodes,guckenheimer2013nonlinear,murdock2003normal}, feedback linearization \cite{isidori1985nonlinear}, and data-driven modal representations \cite{berkooz1993proper,schmid2010dynamic}.
These approaches introduce a transformation operator
\begin{equation}
  \mathscr{T}:\mathcal{U}(D; \mathbb{R}^{d_u}) \rightarrow \mathcal{V}(D; \mathbb{R}^{d_v}),
\end{equation}
which maps the original state $u$ to a latent representation $v = \mathscr{T}(u)$.
Formally, the induced latent dynamics can be written as
\begin{align}\label{relationship_of_infinite_u_and_v}
  \frac{\partial v(t)}{\partial t}
  = (\mathscr{T}'u)[F(u)]
  = \tilde{F}(v),
\end{align}
where $\mathscr{T}'$ denotes the Fréchet derivative of $\mathscr{T}$ and
$\tilde{F}:\mathcal{V}(D; \mathbb{R}^{d_v}) \to \mathcal{V}(D; \mathbb{R}^{d_v})$ represents the latent-space dynamics.
In practice, both $F$ and $\mathscr{T}$ are typically unknown for complex systems, motivating data-driven and learning-based approximations.

By lifting the original state into a higher-dimensional latent space, nonlinear interactions may become more weakly coupled, more separable, or more amenable to approximation by simpler operators. This aligns with the common practice in machine learning of increasing channel dimensionality to enhance representational capacity.
However, such lifting often leads to very high-dimensional or even infinite-dimensional latent spaces.
This is undesirable in practice, as it hinders efficient computation and generalization.
Consequently, a central question is whether the essential dynamics can be represented using a significantly smaller number of effective variables.
This notion is closely related to the concept of intrinsic dimension, which characterizes the minimal number of degrees of freedom required to faithfully represent the underlying dynamics \cite{champion2019data, floryan2022data}.
These studies suggest that, despite the apparent high dimensionality of the state space, the effective dynamics often evolve on a low-dimensional structure \cite{xiong2024koopman, wu2024neural}.

Motivated by these observations, we assume that the latent state $v$ admits a reduced coordinate representation
\begin{equation}
  c = \mathscr{P}_m(v), \quad c \in \mathcal{C}(D; \mathbb{R}^{d_m}),
\end{equation}
where $d_m < d_v$.
Here, $\mathscr{P}_m$ denotes a dimension-reduction operator acting on the variable (channel) dimension. Importantly, $\mathscr{P}_m$ is not restricted to be a spectral projection or associated with predefined basis functions; instead, it represents a data-driven coordinate reduction, which in this work correspond to channel-wise compression implemented by neural network layers.

\subsubsection{Spatial Discretization and Inertial Manifold Inspired Reduction.}
In addition to variable-space considerations, the system state is defined over a spatial domain $D$ and is necessarily discretized in practical computations.
As a result, the continuous operator $F$ is approximated in a finite-dimensional spatial setting, leading to a high-dimensional but finite dynamical system.
Inertial manifold theory \cite{foias1988inertial, temam2012infinite} provides theoretical insight into this setting.
For a broad class of dissipative partial differential equations, the long-term dynamics are attracted exponentially fast to a finite-dimensional manifold embedded in the infinite-dimensional phase space.
Although the existence of inertial manifolds for general dynamical systems has not been established, it is widely believed that many physical systems admit global attractors of finite effective dimension.
Rather than explicitly constructing an inertial manifold or verifying the associated spectral gap conditions, we adopt this theory as a guiding principle.
Specifically, we assume that the essential spatiotemporal dynamics can be parameterized by a finite number of reduced coordinates, while the influence of unresolved components can be captured implicitly through effective nonlinear interactions.

Assuming that the latent dynamics admit a decomposition
\begin{equation}
  \tilde{F}(v) = \mathcal{L}(v) + \mathcal{N}(v),
\end{equation}
where $\mathcal{L}$ and $\mathcal{N}$ denote linear and nonlinear components, respectively.
We posit that the reduced state (coordinates) $c$ evolve according to an effective finite-dimensional dynamics of the form
\begin{align}\label{dynamics_of_c}
  \frac{\partial c(t)}{\partial t}
  = \mathcal{L}_c c + \mathcal{N}_c(c) + \mathcal{R}_c[\mathcal{N}_c(c)].
\end{align}
Here, $\mathcal{L}_c$ represents the reduced linear dynamics, $\mathcal{N}_c$ captures the dominant nonlinear interactions within the reduced space, and $\mathcal{R}_c$ accounts for residual effects arising from unresolved spatial and variable-scale interactions.
This reduced dynamical form is inspired by classical inertial manifold constructions, which suggest that the long-term dynamics of dissipative systems can be effectively parameterized on a finite-dimensional space. A heuristic derivation illustrating this connection is provided in Appendix A.6.
For practical modeling, we absorb the residual contribution into a single nonlinear operator by defining
\begin{equation}
  \mathcal{A}[\mathcal{N}_c(c)]
  := \mathcal{N}_c(c) + \mathcal{R}_c[\mathcal{N}_c(c)].
\end{equation}
This yields the compact reduced dynamical system
\begin{align}\label{compact_dynamics_of_c}
  \frac{\partial c(t)}{\partial t}
  \approx \mathcal{L}_c c + \mathcal{A}[\mathcal{N}_c(c)]
  = \mathscr{F}(c),
\end{align}
This formulation directly motivates the design of neural network hidden layers that parameterize $\mathcal{L}_c$, $\mathcal{N}_c$ and $\mathcal{A}$, as described in the following section.

\subsection{local-global mixing Transformation and Layer}
Inspired by the structure of convective nonlinearity in fluid dynamics (e.g., the term $u \cdot \nabla u$), which inherently couples a velocity field with its local gradients, we propose the \textbf{LGM transformation}. This architecture explicitly fuses local finer-grained features with global spectral information via an element-wise multiplicative gating mechanism.
\begin{definition}\textbf{(local-global mixing Transformation)} The LGM transformation $\mathscr{M}_{\theta}$ is defined as a parameterized operator that fuses local and global information of the reduced state $c$ via an element-wise product:
  \begin{align}
    \mathscr{M}_{\theta}(c) &= \mathscr{L}^{loc}{\theta}(c) \odot \mathscr{G}^{glob}{\theta}(c) \\
    &= \left(\int_{D} p_\theta(x, \tau) c(\tau) d\tau\right) \odot \left(\int_{D} e_\theta(x, \tau) c(\tau) d\tau\right), \quad x \in D,
  \end{align}
  where $\odot$ denotes the Hadamard product. The local branch $\mathscr{L}^{loc}_{\theta}$, parameterized by the kernel $p_\theta$, captures high-frequency, position-specific interactions. The global branch $\mathscr{G}^{glob}_{\theta}$, parameterized by $e_\theta$, extracts domain-wide correlations (e.g., via Fourier or Galerkin-type spectral transforms).
\end{definition}
The multiplicative structure allows $\mathscr{M}_{\theta}$ the flexibility to unify linear and nonlinear representations. If one branch acts as an identity or constant scaling (e.g., $e_\theta(x, \tau) \approx \text{const}$), $\mathscr{M}_{\theta}$ reduces to a linear integral operator (e.g. a pure local or global transformation). Conversely, when both kernels are active, $\mathscr{M}_{\theta}$ functions as a quadratic nonlinear operator. This duality allows us to employ LGM transformations as the building blocks.
Recall that the reduced dynamics in Eq. \eqref{compact_dynamics_of_c}, we approximate the linear component $\mathcal{L}_c$ using a set of linear LGM transformations, and the nonlinear component $\mathcal{N}_c$ using fully nonlinear LGM transformations. Furthermore, consistent with the analysis that the residual operator $\mathcal{A}$ (including $\mathcal{R}_c$) primarily accounts for addtional high-mode local corrections, we can model $\mathcal{A}$ using a composite local block $\mathscr{A}_{\theta}$ (e.g., a convolutional block) to enhance the ability capturing finer-grain details. Combining these components, we define the \textbf{LGM neural layer} to parameterize the effective dynamics $\mathscr{F}(c)$:
\begin{definition}
  \textbf{(Reduced Dynamics via LGM Layer)} The dynamics $\mathscr{F}_\theta(c)$ are approximated by summing $n_l$ linear branches and $n_n$ nonlinear branches, the latter being processed by a local refinement (residual) operator $\mathscr{A}_{\theta}$:
  \begin{align}\label{continuous_dynamics_in_LGM_layer}
    \mathscr{F}_\theta (c)&= \underbrace{\sum_{a=1}^{n_l} {\mathscr{M}^{\mathcal{L}^{a}}_{\theta}} (c)}_{\text{Approximation of } \mathcal{L}_c c}+ \underbrace{\mathscr{A}_{\theta} \left[\sum_{b=1}^{n_n} {\mathscr{M}^{\mathcal{N}^{b}}_{\theta}}(c)\right]}_{\text{Approximation of } \mathcal{A}[\mathcal{N}_c(c)]}.
  \end{align}
  Here, $\mathscr{M}^{\mathcal{L}}_{\theta}$ and $\mathscr{M}^{\mathcal{N}}_{\theta}$ represent LGM transformations dedicated to capturing linear dynamics and nonlinear dynamics, respectively. $\theta$ indicates parameters in neural networks. This multi-branch design allows the layer to resolve independently different components in complex dynamics.
\end{definition}

\subsection{Dynamics-informed Architecture}
The efficacy of modern deep learning relies heavily on network depth to approximate complex nonlinear mappings.
In operator learning for dynamical systems, such depth admits a natural physical interpretation: stacked neural layers can be viewed as discrete realizations of temporal evolution operators, analogous to numerical time integration schemes for differential equations \cite{ren2022phycrnet, weinan2017proposal}.
Motivated by this connection, we propose a dynamics-informed architecture that explicitly bridges continuous-time reduced dynamics and discrete neural network realizations.

Let the transformation $\mathscr{T}$ induce a linear mapping between time variables, $t_v = g(t_u)$, and assume that the projection $\mathscr{P}_m$ preserves the latent time scale such that $t_v = t_c$.
Then the reduced state $c(t_c) = \mathscr{P}_m \mathscr{T}(u)(t_u)$ satisfies
\begin{align}\label{new_relationship_infinite_u_and_v}
  \frac{\partial \mathscr{P}_m \mathscr{T}(u)(t_u)}{\partial t_u}
  = \frac{\partial c(t_c)}{\partial t_c} \frac{\partial t_c}{\partial t_u}
  = \zeta \, \mathscr{F}(c),
\end{align}
where $\zeta = \partial g(t_u) / \partial t_c$ is a time-scaling factor and $\mathscr{F}(c)$ denotes the effective reduced dynamics introduced in Eq. \eqref{compact_dynamics_of_c}.
When discretizing this evolution using a neural network of depth $L_d$, we interpret each layer as an effective discrete evolution operator, so that the overall time scaling is distributed across layers.
However, unlike standard residual networks that employ a fixed step size, real-world physical systems often possess intrinsic timescales that vary significantly across different scenarios. A uniform discretization fails to adapt to these different rates of evolution, making it suboptimal for modeling unknown and complex dynamics.

\subsubsection{Timescale-adaptive Evolution}
To address this issue, we introduce a timescale-adaptive gating mechanism that dynamically modulates the effective evolutionary strength of each layer based on the actual data.
We define a global temporal structural variation metric $\mathcal{K} \in \mathbb{R}$ that quantifies the discretized step size evolved within the dynamical system.
Given a discretized trajectory $\{u^{t_i}\}_{i=0}^{T}$ sampled from the data distribution $P_u$, $\mathcal{K}$ is computed as
\begin{equation}
  \mathcal{K}
  := \mathbb{E}_{u \sim P_u} \left[
    \frac{1}{T} \sum_{i=0}^{T-1}
    \left(
      1 - \frac{\langle u^{t_i}, u^{t_{i+1}} \rangle}
      {\|u^{t_i}\| \, \|u^{t_{i+1}}\|}
    \right)
  \right].
\end{equation}
A larger value of $\mathcal{K}$ indicates rapid dynamical structural changes between consecutive states, whereas small $\mathcal{K}$ reflects quasi-static evolution. Considering $\mathcal{K}$ is computed in the observation space, a learnable scalar $\lambda$ is introduced to bridge the evolutional difference mapping between the original physical space and reduced latent space.

Based on this metric, the reduced state evolves hierarchically according to
\begin{align}\label{hierarchical_update}
  c_l = c_{l-1} + \sigma \big( \lambda_l \, \mathcal{K} \big) \cdot \mathscr{F}_{\theta_l}(c_{l-1}),
  \quad l = 1, \dots, L_d - 1,
\end{align}
where $\mathscr{F}_{\theta_l}$ denotes the parameterized dynamics at layer $l$, $\sigma(\cdot)$ the sigmoid function and $\lambda_l$ a learnable scalar of layer $l$ for the temporal variation in the reduced latent space.
This adaptive gating mechanism ensures bounded residual updates and improves numerical stability, analogous to adaptive step-size control in time integration schemes.

\subsubsection{Parallel Aggregation of Multi-stage Dynamics}
In addition to the hierarchical evolution of intermediate states, we incorporate a parallel aggregation mechanism that combines the dynamical increments produced at different layers.
Each layer yields a intermediate dynamics $\mathscr{F}_{\theta_l}(c_{l-1})$, representing an approximation of the reduced vector field at a particular integration stage.
Rather than relying solely on the terminal hierarchical state, these dynamics are aggregated through learnable weights to form the final update:
\begin{align}
  c_{out} = c_{0} + \sum_{l=1}^{L_d} w_l \mathscr{F}_{\theta_{l}} (c_{l-1}),
\end{align}
where $w_l$ denotes a learnable weight.
This parallel aggregation can be interpreted as constructing a stabilized approximation of the time-integrated vector field, analogous to multi-stage integration schemes \cite{hairer1993solving}.
Such formulations are known to reduce sensitivity to local discretization errors by averaging over multiple intermediate approximations of the dynamics. Notably, the hierachical, parallel and hybrid connection between layers can be derived from the Eq. \eqref{new_relationship_infinite_u_and_v} (details refer to Appendix A.7)


\begin{figure*}[hbtp]
  \centering
  \includegraphics[width=0.85\textwidth]{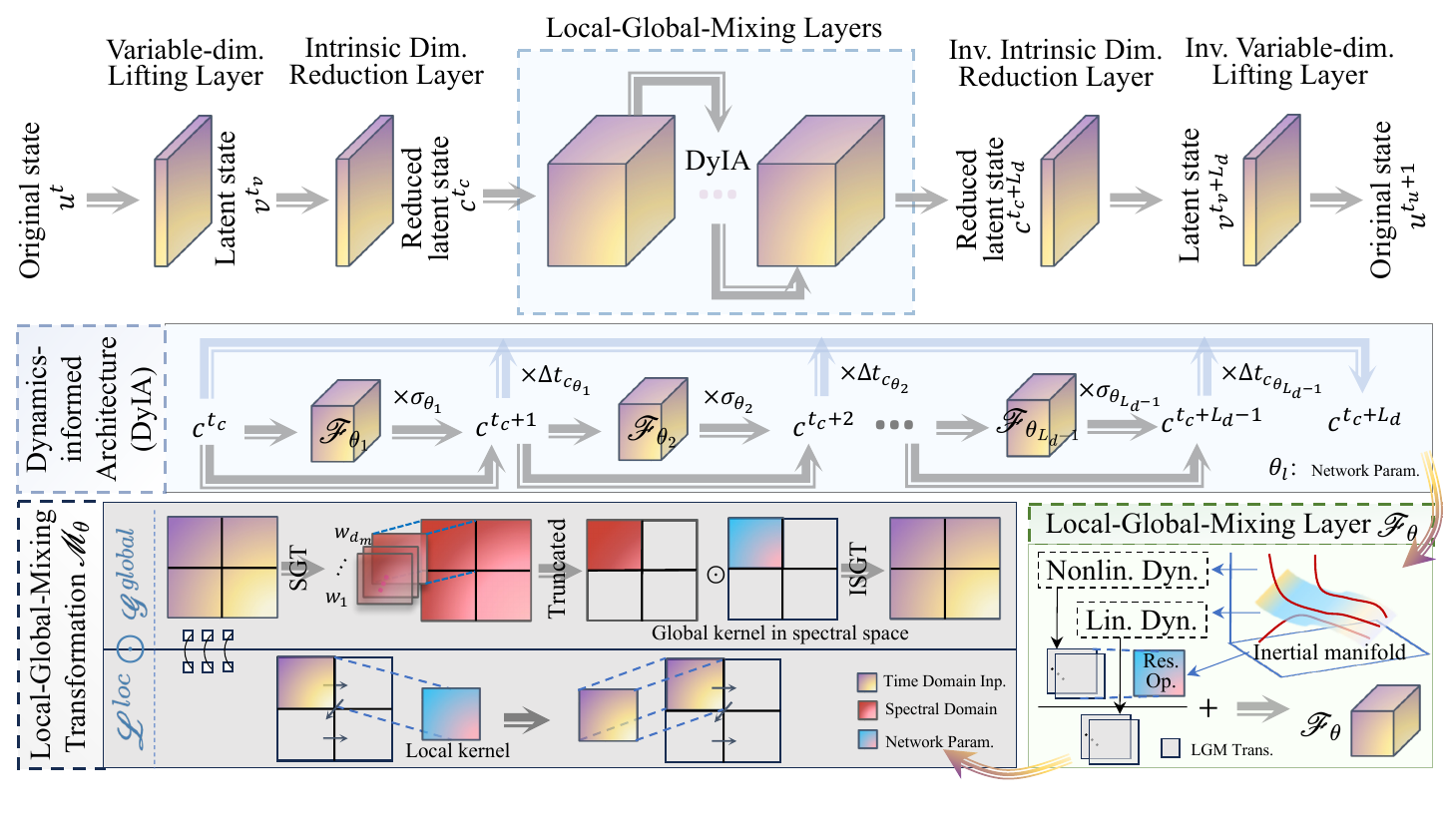}
  \caption{Illustration of the DyMixOp model composed of the variable-dimension lifting layer and its inverse, the intrinsic dimension reduction layer and its inverse, and the LGM layers that adopt LGM transformations to approximate dynamics within the dynamics-informed architecture.}
  \label{fig:arch}
\end{figure*}

\subsection{DyMixOp}
\subsubsection{Overall Architecture}
Based on the preceding analysis, we initiate from the perspective of dynamics and now summarize the proposed architecture as a unified neural evolutionary operator, termed \textbf{DyMixOp} shown in Fig. \ref{fig:arch}.
DyMixOp is designed as a structured approximation of the reduced evolutionary operator $\mathcal{G}$, explicitly respecting the reduced dynamical form introduced in Eq. \eqref{compact_dynamics_of_c}.
For simplicity, we consider the single-step setting ($k=0$) and denote the temporal input by $u^{t} \in \mathcal{U}(D;\mathbb{R}^{d_u})$.
Given a network depth $L_d \in \mathbb{N}$, DyMixOp is formulated as the following operator composition:
\begin{equation}\label{mixop_overall}
  \mathcal{G}^{\dagger}(u^{t}; \theta)
  =
  \mathscr{T}^{-1}
  \circ
  \mathscr{P}_m^{-1}
  \circ
  \mathscr{C}
  \circ
  \mathscr{P}_m
  \circ
  \mathscr{T}
  \,(u^{t}),
\end{equation}
where each component corresponds to a well-defined stage of the reduced dynamical modeling pipeline.

The operator $\mathscr{T}:\mathcal{U}(D;\mathbb{R}^{d_u}) \rightarrow \mathcal{V}(D;\mathbb{R}^{d_v})$ lifts the input state into a latent representation, while $\mathscr{P}_m:\mathcal{V}(D;\mathbb{R}^{d_v}) \rightarrow \mathcal{C}(D;\mathbb{R}^{d_m})$ performs an intrinsic coordinate reduction along the channel dimension. The inverse operators $\mathscr{P}_m^{-1}$ and $\mathscr{T}^{-1}$ reconstruct the solution back to the original state space. Together, these transformations realize a learnable change of variables consistent with the intrinsic low-dimensional manifold.

The core operator $\mathscr{C}:\mathcal{C}(D;\mathbb{R}^{d_m}) \rightarrow \mathcal{C}(D;\mathbb{R}^{d_m})$ approximates the time evolution of the reduced state. $\mathscr{C}$ is implemented as a depth-$L_d$ dynamics-informed architecture that combines timescale-adaptive evolution with parallel aggregation:
\begin{align}\label{mixop_dynamics}
  c_l &= c_{l-1}
  + \sigma\!\left(\lambda_l \mathcal{K}\right)\,
  \mathscr{F}_{\theta_l}(c_{l-1}),
  \quad l = 1,\dots,L_d-1, \\
  \mathscr{C}(c_0) &= c_0 + \sum_{l=1}^{L_d} w_l\, \mathscr{F}_{\theta_l}(c_{l-1}),
\end{align}
where $\mathcal{K}$ is the global temporal variation metric, $\lambda_l$ and $w_l$ are learnable scalars,
and $\sigma(\cdot)$ denotes the sigmoid function. This formulation admits a natural interpretation as an adaptive, multi-stage discretization of the reduced dynamics, analogous to stabilized time-integration schemes.

Each intermediate dynamics $\mathscr{F}_{\theta_l}$ is parameterized by an LGM layer, which approximates the finite-dimensional reduced dynamics:
\begin{align}\label{c_compact_dynamics_dymixop}
  \mathscr{F}(c)
  \approx \mathcal{L}_c c + \mathcal{A}[\mathcal{N}_c(c)],
\end{align}
with distinct linear and nonlinear components:
\begin{equation}\label{mixop_lgm}
  \mathscr{F}_{\theta}(c)
  =
  \sum_{a=1}^{n_l} \mathscr{M}^{\mathcal{L}^a}_{\theta}(c)
  +
  \mathscr{A}_{\theta}
  \left[
    \sum_{b=1}^{n_n} \mathscr{M}^{\mathcal{N}^b}_{\theta}(c)
  \right].
\end{equation}
Here, $\mathscr{M}^{\mathcal{L}}_{\theta}$ and $\mathscr{M}^{\mathcal{N}}_{\theta}$ denote linear and nonlinear LGM transformations, respectively, while $\mathscr{A}_{\theta}$ captures residual local corrections associated with unresolved scales.

Overall, DyMixOp can be viewed as a dynamics-consistent neural operator that factorizes the evolutionary mapping into
(i) latent lifting and intrinsic-dimension reduction,
(ii) inertial manifold inspired decomposition for linear and nonlinear dynamics,
(iii) timescale-adaptive reduced-order evolution incorporating parallel aggregation,
and (iv) inverse transformations for reconstruction to the physical space.
Consequently, DyMixOp provides a principled yet flexible framework for approximating nonlinear evolutionary operators across a wide range of dynamical systems.

\subsubsection{Practical Construction of DyMixOp}
While the preceding sections establish DyMixOp from a dynamics-consistent and operator-theoretic perspective, its concrete realization requires specific architectural parameterizations. These choices are primarily implementation-oriented, guided by empirical evidence from neural operator literature, yet they strictly adhere to the underlying reduced dynamical formulation derived in Eq. \ref{mixop_lgm}.

\paragraph{Mesh-Invariant Lifting and Projection}
A central requirement for neural operators is invariance to the spatial discretization of the domain. To ensure this property, the variable-dimension lifting $\mathscr{T}$, the intrinsic projection $\mathscr{P}_m$, and their inverses are implemented as unit-size convolutional operators.
In practice,  $\mathscr{T}$ lifts the input variable (channel) dimension to twice the intrinsic dimension $m$ of the reduced latent space.
Notably, for task-specific applications where strict mesh-invariance is not a constraint, larger kernel stencils or complicated encoder (decoder) architectures may be employed to enhance local feature extraction capabilities.

\paragraph{Specialization of LGM Transformation and Layer}
Empirically, we observe that the linear component of the reduced dynamics, $\mathcal{L}_c c$, is predominantly governed by local operators (e.g., Laplacian or gradient operators). Consequently, for the approximation of linear terms, we simplify the LGM architecture by constraining the global transformation to an identity operator. The linear LGM branch $\mathscr{M}^{\mathcal{L}}_{\theta}$ is thus specified as:
\begin{equation}
  \mathscr{M}^{\mathcal{L}}_{\theta}(c) = \mathscr{L}^{\mathrm{loc}}_{\theta}(c) \odot \mathbf{1}_D,
\end{equation}
where $\mathbf{1}_D$ denotes the indicator function on the spatial domain $D$. This specification reduces computational complexity while preserving sufficient expressivity to model linear dispersive and dissipative effects. Moreover, this realization naturally includes the effectiveness stemming from the adding interaction mechanism. In practice, the $\mathscr{L}^{\mathrm{loc}}_{\theta}$ is realized via unit-size convolutional operators and differential operators \cite{liu2024neural}, remaining invariant to spatial discretization and mesh resolution.

Conversely, for the approximation of nonlinear terms, both local and global transformations in the LGM branch $\mathscr{M}^{\mathcal{N}}_{\theta}$, are retained to capture the full spectrum of interactions:
\begin{equation}
  \mathscr{M}^{\mathcal{N}}_{\theta}(c) = \mathscr{L}^{\mathrm{loc}}_{\theta}(c) \odot \mathscr{G}^{\mathrm{glob}}_{\theta}(c).
\end{equation}
This obeys the initial intention for modeling quadratic and higher-order nonlinearities. In practice, the global operator $\mathscr{G}^{\mathrm{glob}}_{\theta}$ is realized via Fourier transforms with learnable, truncated spectral coefficients that implemented in FNO \cite{li2020fourier}, leveraging the efficiency of the Fast Fourier Transform (FFT) to resolve non-local dependencies. Meantime, the local operator $\mathscr{L}^{\mathrm{loc}}_{\theta}(c)$ is realized via unit-size convolutional operators. It is worth mentioning that global transformations are not limited to classical spectral methods and can also be implemented using Transformer- and Mamba-based architectures.

To further model the nonlinear dynamics, the residual operator $\mathscr{A}_{\theta}$ is implemented as a composite convolutional block comprising two convolutional layers. The intermediate channel dimension is expanded to $d_v$, thereby enhancing the operator’s capacity to recover unresolved or previously suppressed dynamical features from the latent space.

\paragraph{Dynamics Stabilization}
Finally, to stabilize the recursive evolution of the reduced state, nonlinear activation functions are applied directly to the computed dynamics $\mathscr{F}_{\theta}(c)$ rather than solely to the state variables as follows:
\begin{equation}
  c_l = c_{l-1} + \sigma(\lambda_l \mathcal{K}) \, \sigma_{\mathrm{GELU}}\!\left(\mathscr{F}_{\theta_l}(c_{l-1})\right).
\end{equation}
This placement of nonlinearities ensures bounded dynamical increments, thereby improving the numerical stability of the deep operator architecture.

\section{Results}
\subsection{Benchmarks}\label{benchmarks}
To demonstrate the scalability and versatility of our method across varying spatial dimensionalities and physical complexities, we conduct experiments on seven benchmark datasets. These cases cover elliptic, parabolic, and hyperbolic PDEs, ranging from steady-state problems to chaotic spatiotemporal evolutions: (i) \textbf{1D Kuramoto-Sivashinsky (KS)}: a fourth-order nonlinear parabolic PDE. Originally derived for flame front instabilities, it is a canonical PDE for studying complex, multiscale and spatiotemporal chaos.
(ii) \textbf{2D Darcy Flow:} An elliptic PDE governing steady-state flow in porous media with spatially heterogeneous permeability. This case tests the operator's capacity to map parameter fields (coefficients) to solution fields in a non-temporal setting.
(iii) \textbf{2D Burgers Equation:} A parabolic PDE combining nonlinear advection and viscous diffusion. It describes shock wave formation and viscous dissipation in fluid mechanics.
(iv) \textbf{2D Navier-Stokes:} Modeling incompressible viscous fluids in the vorticity-stream function formulation. This parabolic system exhibits rich flow structures and is a standard benchmark for capturing complex vortex dynamics.
(v) \textbf{2D Compressible Euler (CE-CRP):} A hyperbolic system representing inviscid flow, specifically the four-quadrant Riemann problem. This PDE features curved domain partitions and the formation of shock waves and contact discontinuities.
(vi) \textbf{3D Shallow Water Equations:} Modeling global atmospheric circulation and Rossby waves on a rotating sphere. This hyperbolic-viscous system involves zonal flows and barotropic instabilities, requiring the capability to handle spherical geometry and large-scale geophysical dynamics.
(vii) \textbf{3D Brusselator:} A nonlinear reaction-diffusion system (parabolic) describing autocatalytic chemical reactions. This case focuses on the formation of self-organized oscillating patterns, evaluating the ability to capture the interplay between diffusion and nonlinear reaction kinetics in a volumetric domain.
In summary, the detailed description of the generation and implementation for all benchmarks can be found in Appendix A.3.

\subsection{Baselines}\label{baselines}
The selected baselines are configured as follows: (i) \textbf{ConvLSTM} \cite{shi2015convolutional}: a recurrent architecture adapted with $1 \times 1$ convolutions to function as a mesh-invariant operator that mixes local features via gating mechanisms. (ii) \textbf{Fourier Neural Operator (FNO)} \cite{li2020fourier}: a classical spectral baseline combining a global Fourier integral branch with a local linear branch to capture frequency-domain dependencies. (iii) \textbf{DeepONet} \cite{lu2021learning}: a global operator that approximates solutions via the dot product of a coordinate-processing Trunk and an input-processing Branch, configured here with aggressive downsampling for efficiency. (iv) \textbf{U-NO} \cite{rahman2022uno}: a multi-scale model applying a U-Net-inspired encoder-decoder structure with skip connections, facilitating deep spectral processing through progressive domain contraction and expansion. (v) \textbf{Generalized Neural Operator Transformer (GNOT)} \cite{hao2023gnot}: a transformer-based model employing linear attention for efficient global processing, implemented here with a non-autoregressive sequence-to-sequence strategy to optimize memory for multi-step predictions. (vi) \textbf{LocalNO} \cite{liu2024neural}: an architecture that augments the FNO backbone with parallel differential and integral operator branches, explicitly designed to mitigate spectral over-smoothing and capture high-frequency local details. (vii) \textbf{CoDA-NO} \cite{rahman2024pretraining}: a multiphysics-oriented model that tokenizes variables along the codomain to model inter-variable dependencies via function-space attention within a sequence-to-sequence framework. (viii) \textbf{Latent Mamba Operator (LaMO)} \cite{tiwari2025latent}: a state-space model that utilizes Structured State-Space Models (SSMs) to capture long-range dependencies with linear complexity, processing latent patches via multi-directional scanning.
The detailed description for the implementation of baselines can be found in Appendix A.4.

\subsection{Training Details and Metrics}
In many existing studies, reported performance gains often stem from meticulous hyperparameter tuning or task-specific architectural adjustments, rather than from fundamental algorithmic superiority. This reliance on over-optimization creates significant challenges for making a fair comparison \cite{mcgreivy2024weak}. To address this issue, we evaluate all baselines using their standard configurations for regular grids, and we strictly avoid task-specific parameter engineering to minimize tuning bias.

Our scaling strategy focuses principally on adjusting the hidden dimension and network depth to vary the model capacity. Meanwhile, essential architecture-specific parameters, such as global modes in FNO or attention heads in GNOT and LaMO, are scaled proportionally. Crucially, we align these hyperparameter settings based on \textbf{peak GPU memory usage} rather than the parameter size. Since the parameter count often scales nonlinearly with actual computational costs, memory usage serves as a more practical metric for fairness across diverse hyperparameter spaces. Under such strict restrictions, the investigation allows for a more reliable assessment of the intrinsic potential of each model architecture. Ultimately, we construct three model variants—Tiny, Medium, and Large—spanning a diverse range of GPU memory usage.

For the setting for training hyperparameters, we take the uniform setting for all baselines. All of them are trained for $500$ epochs using the learning rate of $1e-3$ and the AdamW optimizer \cite{loshchilov2017decoupled} with $0.97$ gamma and $7$ step size. StepLR scheduler are utilized to modify the learning rate and a batch size of $32$ is used. Furthermore, to mitigate stochastic variations and evaluate robustness, all experiments are conducted using two distinct random seeds $\{123, 456\}$, and the averaged results are reported.

Suppose $u_i,u^{'}_i\in \mathbb{R}^{n}$ is the ground truth solution and the predicted solution for the $i$-th sample, and $D$ is the size of the dataset.
All datasets are normalized in a min-max normalization as follows,
$$u^* = (u-u_{\text{min}})/u_{\text{max}}$$
where $u_{\text{max}}$ is the maximum in the dataset and $u_{\text{min}}$ the minimum in the dataset.
For all datasets, we adopt the relative mean squared error (MSE) as the metric for training and evaluation, aiming to address extremely minimal solutions and prevent gradient vanishing. The relative MSE error is computed as follows,
$$\epsilon=\frac{1}{D}\sum_{i=1}^{D}\frac{(u^{'}_i-u_i)^2}{u_i^2}.$$

To enable a more intuitive and fair comparison across diverse benchmarks—each of which may exhibit vastly different loss magnitudes due to variations in physical scales, discretization resolutions, or problem complexity—we introduce a Log-Min-Max normalized score mapped to the \([0, 100]\) interval. This scoring scheme mitigates polarization caused by orders-of-magnitude differences in raw error values, ensuring that performance gains are not disproportionately dominated by benchmarks with inherently larger losses.

Specifically, let \(\epsilon_{\text{min}}\) and \(\epsilon_{\text{max}}\) denote the smallest and largest relative MSE values observed among all evaluated models. The Log-Min-Max normalized score for a model with error \(\epsilon\) is computed as:
\[
  \text{Score} = 100 \cdot \left(1 - \frac{\log(\epsilon) - \log(\epsilon_{\text{min}})}{\log(\epsilon_{\text{max}}) - \log(\epsilon_{\text{min}})}\right),
\]
where the logarithmic transformation ensures uniform sensitivity across multiple orders of magnitude. By construction, a score of 100 corresponds to the best-performing model on that benchmark (\(\epsilon = \epsilon_{\text{min}}\)), and a score of 0 corresponds to the worst-performing model (\(\epsilon = \epsilon_{\text{max}}\)).
This normalization yields a bounded, interpretable metric that facilitates cross-benchmark aggregation (e.g., average score over multiple tasks) while preserving rank order and emphasizing relative improvement rather than absolute error scale. All reported comparative results in Section \ref{sec:results} are presented using this standardized scoring system unless otherwise noted.

\subsection{Main Results}\label{sec:results}
We evaluate the proposed DyMixOp architecture across a comprehensive suite of PDE benchmarks spanning diverse physical regimes in dimensions from 1D to 3D. Our assessment encompasses both predictive accuracy and computational efficiency, benchmarked against eight representative neural operator baselines.
\begin{figure}[htbp]
  \centering
  \includegraphics[width=\linewidth]{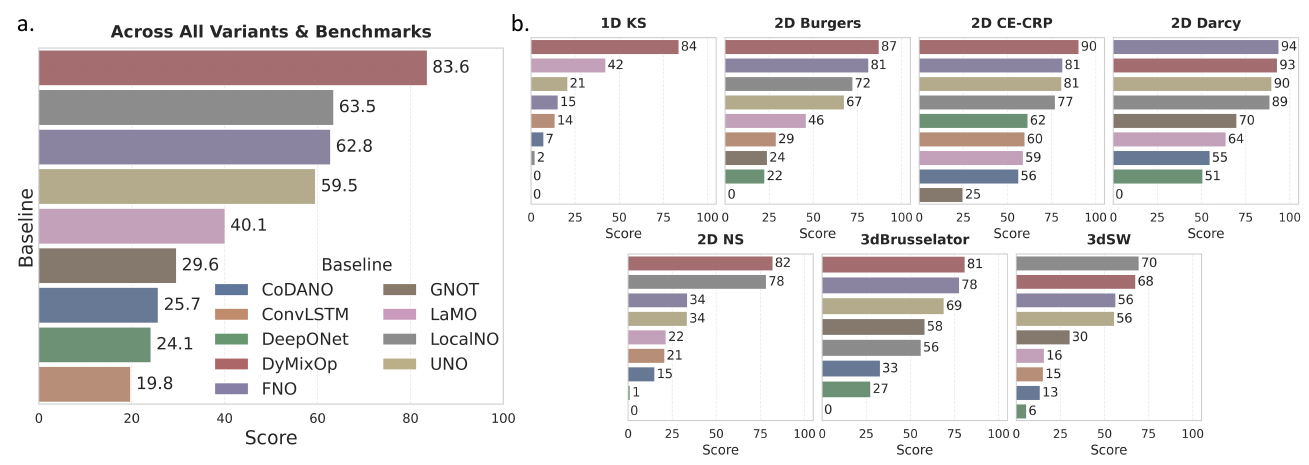}
  \caption{Performance of neural operator architectures across diverse PDE benchmarks.
    (a) Averaged Log-Min-Max normalized scores (0–100) over all benchmarks, model variants and random seeds, showing DyMixOp’s substantial lead.
  (b) Per-benchmark breakdown. DyMixOp dominates five of seven benchmarks, with its only minor deficit occurring in the 2D Darcy and 3D SW benchmarks.}
  \label{fig:performance}
\end{figure}

\subsubsection{Accuracy and Generalization Across PDE Benchmarks}
As shown in Fig. \ref{fig:performance}a, DyMixOp achieves a mean normalized score 83.6, substantially outperforming all competitors. FNO (scoring 62.8) and LocalNO (scoring 63.5)—among the strongest existing architectures—lag by approximately 20 points, while UNO (scoring 59.5), LaMO (scoring 40.1), GNOT (scoring 29.6), CoDANO (scoring 25.7), DeepONet (scoring 24.1), and ConvLSTM (scoring 19.8) exhibit progressively weaker generalization. It underscores DyMixOp’s excellent general capacity across all variants and benchmarks in a unified perspective.
A per-benchmark breakdown (Fig. \ref{fig:performance}b) reveals that DyMixOp leads on five of seven tasks. On the chaotic 1D Kuramoto–Sivashinsky system, DyMixOp scores 84, more than double FNO’s 42. All competitor baselines fail to capture the instability-driven chaos behavior where rapid multi-scale gradient transformations and stiff fourth-order dynamics overwhelm conventional architectures lacking explicit physical constraints. For 2D Burgers’ and compressible Euler systems (involving shocks and rarefactions), DyMixOp achieves 87 and 90, respectively—outpacing FNO and UNO by 6–9 points. On 2D Navier–Stokes and 3D Brusselator systems, DyMixOp (scoring 82 and 81) again leads, reflecting robustness in turbulent-like and pattern-forming dynamics.
In 2D Darcy flow (elliptic), all top models perform well (FNO: scoring 94, DyMixOp: scoring 93, UNO: scoring 90, LocalNO: scoring 89), and DyMixOp performs slightly worse than FNO.
Notably, FNO (scoring 94) slightly outperforms DyMixOp (scoring 93) on the 2D Darcy case, and LocalNO (scoring 70) marginally exceeds DyMixOp (scoring 68) on the 3D SW benchmark, likely due to the benchmarks’ predominantly linear dynamic character, where the mixing interaction offers less advantage.
These results confirm that DyMixOp excels precisely in regimes dominated by strong nonlinearity, multi-scale coupling, and rapid structural evolution. These are exactly the where other operators degrade.

\begin{table}[htbp]
  \centering
  \small
  \setlength{\tabcolsep}{4pt}
  \renewcommand{\arraystretch}{1.2}

  \caption{Comparison of Model Loss Statistics. \textbf{DyMixOp (Ours)} is compared against the best performance of each baseline. The \textbf{Improvement} row indicates the percentage reduction in error achieved by DyMixOp compared to the previous state-of-the-art (SOTA). \textbf{Bold}, \textbf{\color{bestblue}blue}, and \uline{underline} indicate the 1\textsuperscript{st}, 2\textsuperscript{nd}, and 3\textsuperscript{rd} best results among competitor models.}
  \label{tab:results_comparison}
  \sisetup{
    output-exponent-marker = \ensuremath{\mathrm{e}},
    retain-explicit-plus = false,
    mode = match,
    reset-text-series = false,
    reset-text-family = false,
    reset-text-shape = false,
    text-series-to-math = true,
    exponent-mode = scientific,
    table-format = 1.2e-1
  }
  \begin{tabular}{l *{7}{S}}
    \toprule
    & {\textbf{1D}} & \multicolumn{4}{c}{\textbf{2D}} & \multicolumn{2}{c}{\textbf{3D}} \\
    \cmidrule(lr){2-2} \cmidrule(lr){3-6} \cmidrule(lr){7-8}
    \textbf{Model} & {\textbf{KS}} & {\textbf{Burgers}} & {\textbf{CE-CRP}} & {\textbf{Darcy}} & {\textbf{NS}} & {\textbf{Brussel.}} & {\textbf{SW}} \\
    \midrule

    FNO \cite{li2020fourier} & {\third{3.75e-1}} & {\best{1.82e-3}} & {\third{3.39e-2}} & {\best{7.40e-5}} & {\third{5.58e-2}} & {\best{5.40e-5}} & {\second{1.27e-3}} \\

    DeepONet \cite{lu2021learning} & 1.00e0 & 1.16e-1 & 8.78e-2 & 2.59e-3 & 8.56e-1 & 1.11e-2 & 4.13e-1 \\

    GNOT \cite{hao2023gnot} & 1.00e0 & 1.07e-1 & 1.73e-1 & 3.80e-4 & 9.93e-1 & \third{4.67e-4} & 2.06e-2 \\

    ConvLSTM \cite{shi2015convolutional} & 4.18e-1 & 6.84e-2 & 9.61e-2 & 2.91e-1 & 1.79e-1 & 3.20e-1 & 1.03e-1 \\

    LaMO \cite{tiwari2025latent} & {\best{5.21e-2}} & 1.53e-2 & 9.72e-2 & 1.01e-3 & 1.48e-1 & \na & 1.70e-1 \\

    LocalNO \cite{liu2024neural} & 7.61e-1 & {\second{3.05e-3}} & {\best{3.07e-2}} & {\third{1.03e-4}} & {\best{7.46e-4}} & 2.58e-4 & {\best{4.37e-4}} \\

    U-NO \cite{rahman2022uno} & {\second{1.44e-1}} & {\third{3.43e-3}} & {\second{3.33e-2}} & {\second{9.60e-5}} & {\second{4.93e-2}} & {\second{8.20e-5}} & {\third{3.80e-3}} \\

    CoDA-NO \cite{rahman2024pretraining} & 6.48e-1 & 5.02e-1 & 1.12e-1 & 2.00e-3 & 3.03e-1 & 1.27e-2 & 2.08e-1 \\

    \midrule
    \rowcolor{rowgray}
    \textbf{DyMixOp (Ours)} & \textbf{2.96e-3} & \textbf{9.18e-4} & \textbf{1.51e-2} & \textbf{5.50e-5} & \textbf{4.07e-4} & \textbf{3.60e-5} & 5.64e-4 \\

    \textbf{Improvement} & \imp{94.3} & \imp{49.5} & \imp{50.8} & \imp{25.7} & \imp{45.4} & \imp{33.3} & \noimp \\

    \bottomrule
  \end{tabular}
\end{table}

\begin{figure}[htbp]
  \centering
  \includegraphics[width=1\linewidth]{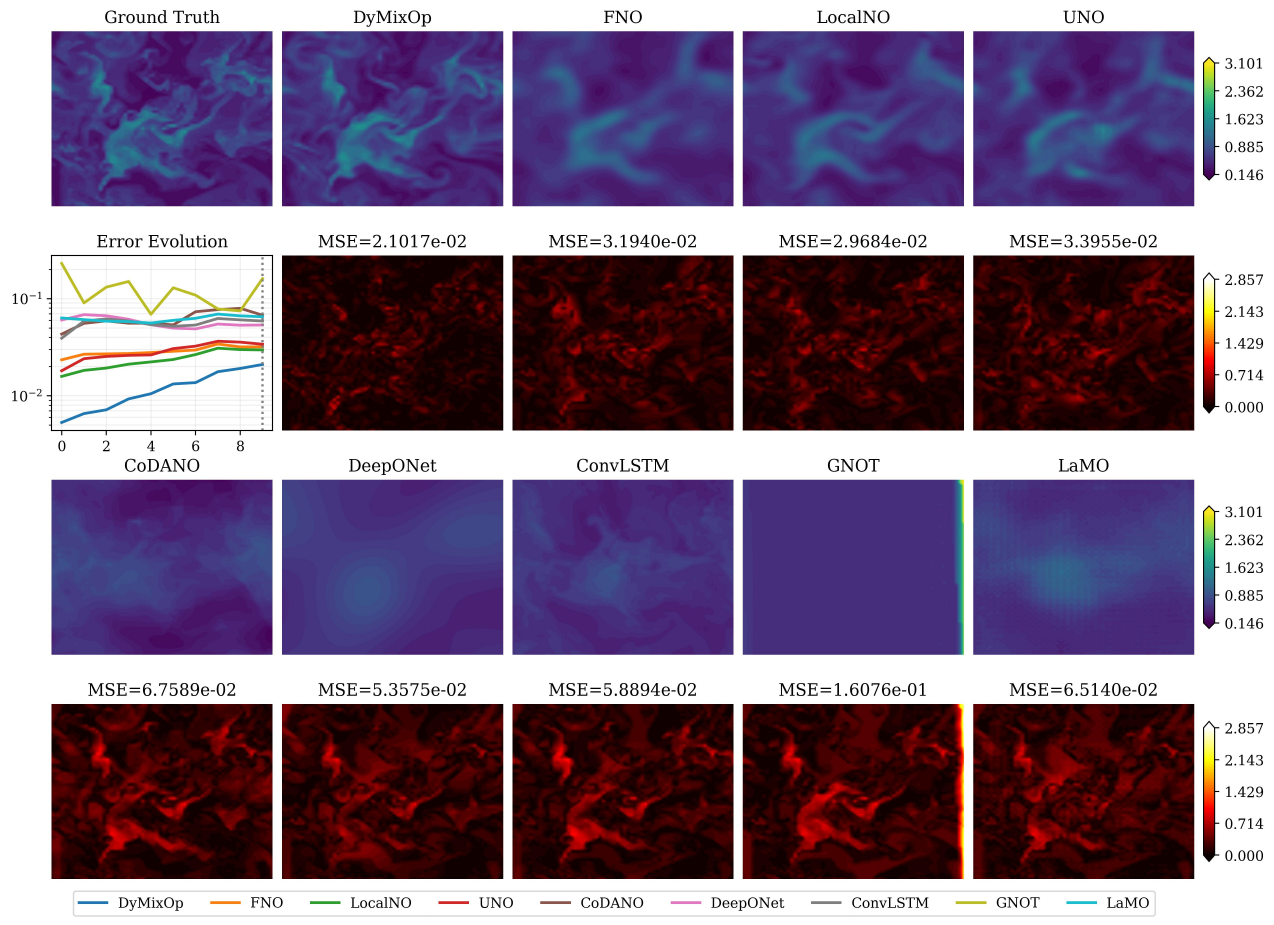}
  \caption{Visualization of baselines on 2D CE-CRP benchmark. Each panel displays a velocity scalar field over a square domain}
  \label{fig:visualization}
\end{figure}

While the Log-Min-Max scores facilitate aggregate comparison, we explicitly report the relative MSE errors to detail the precise magnitude of improvement. Table \ref{tab:results_comparison} compares the lowest errors achieved by DyMixOp against the best performing baselines. A complete comparison between all variants refers to Appendix A.5. 
The 'Improvement' row highlights the significant gains made by our approach. Most notably, on the chaotic 1D KS system, DyMixOp reduces the error by 94.3 compared to the previous state-of-the-art. For 2D tasks involving shocks and turbulence (Burgers, CE-CRP, and NS), we observe consistent error reductions ranging from 45$\%$ to 51$\%$. Even in the smoother Darcy flow and 3D Brusselator cases, DyMixOp improves accuracy by 25.7 and 33.3, respectively. While LocalNO retains a slight advantage on the 3D SW task, DyMixOp achieves the lowest error in 6 out of 7 benchmarks.
These results confirm that DyMixOp provides a robust and unified solution for diverse physical systems. By mixing local and global transformations, it excels at capturing high-frequent components and modeling the strong nonlinearities and rapid changes that limit other operators. As illustrated in Fig. \ref{fig:visualization}, DyMixOp achieves superior reconstruction fidelity on the 2D CE-CRP benchmark, particularly in resolving fine-grained dynamical structures—thereby substantiating its enhanced representational capacity. Unlike other models that specialize in specific types of flow, DyMixOp generalizes effectively across chaotic, turbulent, and steady-state dynamics without requiring problem-specific tuning. Additional visual comparisons and analysis are provided in Appendix A.8. Zero-shot super-resolution results are resported in Appendix A.9.

\begin{figure}[htbp]
  \centering
  \includegraphics[width=\linewidth]{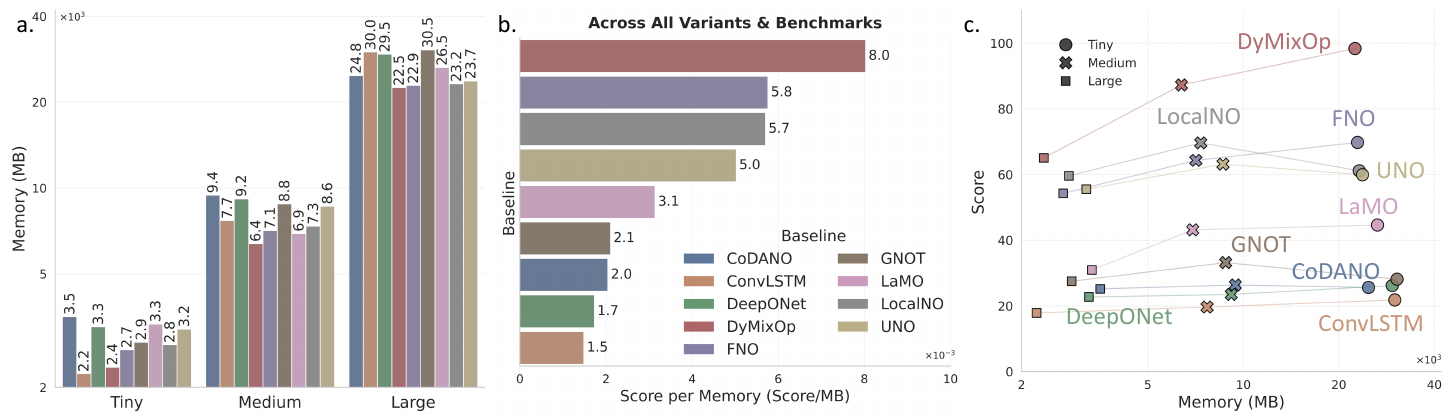}
  \caption{Computational efficiency and memory–performance trade-offs across model scales.
    (a) Peak GPU memory consumption (MB) for tiny, medium, and large variants; DyMixOp almost maintains smallest usage across all variants with a minor deficit on the tiny variant.
    (b) Performance per unit memory (Score/MB), where DyMixOp achieves the highest efficiency (reaching $8.0 \times 10^{-3}$), outperforming all baselines by $\geq$ 38$\%$.
  (c) Scatter plot of normalized score versus memory, with markers indicating model size (Tiny: $\bullet$, Medium: $\times$, Large: $\blacksquare$). DyMixOp lies on the Pareto frontier, delivering superior accuracy at comparable or lower memory cost, with monotonic scaling behavior.}
  \label{fig:efficiency}
\end{figure}

\subsubsection{Computational Efficiency and Scaling Performance}
Accuracy alone is insufficient for real-world deployment, and efficiency in memory and scalability are equally critical. Fig. \ref{fig:efficiency} evaluates these aspects across three model scales: Tiny, Medium, and Large.
Fig. \ref{fig:efficiency}a shows that DyMixOp almost maintains a lowest GPU memory usage across all scales: 2351 MB (Tiny), 6384 MB (Medium), and 22491 MB (Large). It is comparable to ConvLSTM at the tiny scale and below other models on other scales, ensuring the fair comparison between our method and baselines in a practical deployment, instead of the count of parameters.
More revealing is the performance-per-memory ratio as shown in Fig. \ref{fig:efficiency}b. DyMixOp achieves $8.0e-3$ (score/MB), surpassing FNO (reaching $5.8e-3$) and LocalNO (reaching $5.7e-3$) by over 35$\%$, and outperforming LaMO, GNOT, ConvLSTM, and DeepONet by factors of 2–5. This metric highlights DyMixOp’s extraordinary computational efficiency.
The scaling performance is visualized in Fig. \ref{fig:efficiency}c, which plots score versus memory on a log scale. DyMixOp dominates the Pareto frontier: its Large variant reaches 98 score at 22 GB memory while its Tiny variant (scoring 65 at 2.7 GB) exceeds most baselines’ Large configurations in performance. Crucially, DyMixOp exhibits monotonic, near-linear scaling that increasing capacity yields consistent gains within the same amount of data, unlike several baselines that suffer from diminishing performance likely due to the insufficient amount data.

Together, these results establish DyMixOp as a simultaneously accurate, efficient, and scalable neural operator. It not only sets a new state-of-the-art in predictive performance across a broad spectrum of PDEs but also achieves the highest utility per unit of computational resource. This dual advantage arises directly from its design intention stemming from the dynamical perspective. By embedding inertial manifold-inspired reduction, convection-informed LGM-based nonlinear mixing, and dynamics-informed architecture into a unified framework, DyMixOp realizes a dynamics-consistent approximation that is both expressive and efficient.

\subsection{Ablation Experiments}
In this section, we conduct a series of ablation experiments on several benchmarks to isolate and quantify the contributions of key architectural components: the interaction mechanisms in LGM transformations, the dimension transformation strategy, the structure of the LGM layers, and the dynamics-informed architecture (DyIA).

To establish a controlled baseline, we define a standard neural operator architecture consisting of a two-layer encoder (responsible for variable lifting and intrinsic reduction respectively), a latent processing block, and a two-layer decoder (handling the inverse transformations). In this baseline configuration, the latent block is composed of multiple sequentially connected layers, where each layer is parameterized solely by a local convolutional transformation. All components are connected in a hierarchical method. This local-convolution-only model serves as the reference backbone. Subsequent ablation experiments are performed by incrementally modifying this baseline, progressively introducing or substituting components to rigorously evaluate the efficacy of the proposed DyMixOp architecture. All statistics are obtained by setting two different random seeds.

\subsubsection{Impact of Interaction Mechanisms}
In this ablation study, we explore the significance of the nonlinear LGM transformation. To isolate its impact, we replace the core interaction module with four alternative mechanisms: purely local convolution, purely global spectral transformation, LGA, and a composite "LGM + Local" variant. All comparisons are conducted under comparable GPU memory budgets, as detailed in Table \ref{tab:ablation_lgm}.
The local-only architecture yields the weakest performance across the 2D NS and 3D SW benchmarks, consistent with the theoretical expectation that finite receptive fields are inadequate for capturing global flow evolution. Conversely, the global-only approach significantly reduces errors by resolving low-frequency modes but reaches a performance plateau (reaching $0.0455$ on 2D NS). This saturation likely arises from the spectral bias of global methods, which struggle to represent local discontinuities and high-frequency oscillations.
Introducing local features via an adding interaction improves upon the global baseline (reaching $1.63e-4$ on 2D Darcy, $0.0293$ on 2D NS), confirming that multi-scale information is valuable. Even their advantage, they still fail to promote at some specific regimes (e.g. global-only on 2D Darcy, LGA on 3D SW).
In contrast, the proposed Mixing (LGM) mechanism demonstrates superior efficiency and generalization.
On the 2D Darcy and 3D SW benchmarks, LGM achieves the lowest test losses (reaching $7.20e-5$ and $4.56e-4$, respectively), outperforming LGA. This confirms that the mixing interaction mimicking convective nonlinearity is physically more expressive than simple linear addition for coupling scales.
It also achieves the fastest training time per epoch across the autoregressive tasks ($15.45$s on 2D NS, $9.40$s on 3D SW).
We further investigated a composite "LGM + Local" variant, which explicitly adds a local convolutional branch within the nonlinear block. On the 2D NS benchmark, this variant yields the lowest error $0.0182$, suggesting that strong turbulence requires intensive local refinement. However, on 2D Darcy and 3D SW, this addition degrades performance. This indicates that incorporating explicit local linearity within the nonlinear transformation can inadvertently interfere with the model's ability to learn nonlinear dynamics in certain regimes.
Consequently, we adopt the LGM transformation alone as the standard module for approximating nonlinear dynamics ($\mathcal{N}_c$). This choice maximizes computational efficiency and robustness across diverse physics. Importantly, the benefits of local processing seen in the "LGM + Local" variant are not lost. They are structurally retained in the full DyMixOp architecture through the separate, dedicated linear dynamics branch ($\mathcal{L}_c$), ensuring that local-linear features are modeled without interfering with the nonlinear mixing process.
\begin{table}[htbp]
  \centering
  \caption{\textbf{Ablation study of interaction mechanisms.} Comparison of test relative MSE errors, GPU memory consumption, and time per epoch on 2D Darcy Flow, 2D Navier-Stokes (2D NS), and 3D Shallow Water (3D SW) benchmarks. Best results are highlighted in \textbf{bold}.}
  \label{tab:ablation_lgm}
  \vspace{0.2cm}
  \renewcommand{\arraystretch}{1.2}
  \setlength{\tabcolsep}{3pt}
  \resizebox{\textwidth}{!}{
    \begin{tabular}{l c c c c c c c c c}
      \toprule
      & \multicolumn{3}{c}{\textbf{2D Darcy}} & \multicolumn{3}{c}{\textbf{{2D NS}}} & \multicolumn{3}{c}{\textbf{3D SW}} \\
      \cmidrule(lr){2-4} \cmidrule(lr){5-7} \cmidrule(lr){8-10}
      \textbf{Interaction} & \textbf{Mem} & \textbf{Test} & \textbf{Time} & \textbf{Mem} & \textbf{Test} & \textbf{Time} & \textbf{Mem} & \textbf{Test} & \textbf{Time} \\
      \textbf{Mechanism}   & \textbf{(MB)} & \textbf{Loss} & \textbf{(s)} & \textbf{(MB)} & \textbf{Loss} & \textbf{(s)} & \textbf{(MB)} & \textbf{Loss} & \textbf{(s)} \\
      \midrule
      Local Only     & 2,854 & 2.58e-2 $\pm$ 0.00e-2 & 1.70 & 32,338 & 0.5805 $\pm$ 0.2835 & 18.75 & 17,198 & 0.4112 $\pm$ 0.1479 & 9.90 \\
      Global Only    & 2,926 & 2.88e-4 $\pm$ 0.24e-4 & 1.10 & 30,022 & 0.0455 $\pm$ 0.0011 & 28.35 & 17,174 & 2.68e-3 $\pm$ 2.28e-3 & 17.25 \\
      Adding (LGA)   & 2,930 & 1.63e-4 $\pm$ 0.02e-4 & 1.25 & 32,894 & 0.0293 $\pm$ 0.0079 & 21.55 & 17,228 & 1.20e-2 $\pm$ 1.65e-2 & 11.55 \\
      Mixing (LGM)   & 3,028 & \textbf{7.20e-5 $\pm$ 0.28e-5} & 1.30 & 31,766 & 0.0233 $\pm$ 0.0022 & 15.45 & 17,724 & \textbf{4.56e-4 $\pm$ 3.32e-4} & 9.40 \\
      LGM + Local    & 2,897 & 9.45e-5 $\pm$ 0.07e-5 & 1.50 & 32,132 & \textbf{0.0182 $\pm$ 0.0027} & 17.85 & 17,523 & 4.92e-4 $\pm$ 3.26e-4 & 10.80 \\
      \bottomrule
    \end{tabular}
  }
\end{table}

\subsubsection{Impact of Dimension Transformation Strategy}
Constructing a mapping from the infinite-dimensional physical state space to a finite-dimensional inertial manifold is a cornerstone of the DyMixOp framework. This process is structurally realized through a two-layer lifting-reduction strategy: an encoder first lifts the inputs to a high-dimensional latent space to disentangle nonlinear dynamics, followed by a projection onto the system's intrinsic dimension coordinates.
To evaluate the sensitivity of the model to the dimension transformation, we conducted an ablation study varying the depth of the encoder/decoder modules. We compare the standard two-layer lifting-reduction architecture against a simplified one-layer lifting strategy and a deeper four-layer lifting-reduction network, with results detailed in Table \ref{tab:ablation_dim}.
Notably, the empirical results indicate that the model's performance is remarkably robust to the depth of the dimension transformation. Unlike the order-of-magnitude variations observed when altering interaction mechanisms, the differences in test loss between single-layer, two-layer, and four-layer configurations are generally marginal. This suggests that while the existence of a high-dimensional lifting map is critical, the specific topological complexity of this map is less distinct.
Although task-specific tuning could yield marginal gains (e.g., using One-Layer for Darcy), the two-layer lifting-reduction strategy emerges as the most consistent performer across multiple benchmarks. Consequently, we adopt the two-layer strategy as the standard configuration, striking an optimal balance between implementation simplicity, computational efficiency, and representational sufficiency.
\begin{table}[htbp]
  \centering
  \caption{\textbf{Ablation study of dimension transformation.} Comparison of test relative MSE errors and GPU memory consumption across 2D Darcy, 2D Navier-Stokes (2D NS), and 3D Shallow Water (3D SW) benchmarks. Best results are highlighted in \textbf{bold}.}
  \label{tab:ablation_dim}
  \vspace{0.2cm}
  \renewcommand{\arraystretch}{1.2}
  \setlength{\tabcolsep}{3pt}
  \resizebox{\textwidth}{!}{
    \begin{tabular}{l c c c c c c c c c}
      \toprule
      & \multicolumn{3}{c}{\textbf{2D Darcy}} & \multicolumn{3}{c}{\textbf{2D NS}} & \multicolumn{3}{c}{\textbf{3D SW}} \\
      \cmidrule(lr){2-4} \cmidrule(lr){5-7} \cmidrule(lr){8-10}
      \textbf{Transformation} & \textbf{Mem} & \textbf{Test} & \textbf{Time} & \textbf{Mem} & \textbf{Test} & \textbf{Time} & \textbf{Mem} & \textbf{Test} & \textbf{Time} \\
      \textbf{Strategy}       & \textbf{(MB)} & \textbf{Loss} & \textbf{(s)} & \textbf{(MB)} & \textbf{Loss} & \textbf{(s)} & \textbf{(MB)} & \textbf{Loss} & \textbf{(s)} \\
      \midrule
      Two-Layer  & 3,028 & 7.20e-5 $\pm$ 0.28e-5 & 1.30 & 31,766 & \textbf{0.0233 $\pm$ 0.0022} & 15.45 & 17,724 & \textbf{4.56e-4 $\pm$ 3.32e-4} & 9.40 \\
      One-Layer  & 2,954 & \textbf{5.90e-5 $\pm$ 0.28e-5} & 1.25 & 31,598 & 0.0236 $\pm$ 0.0061 & 15.30 & 17,960 & 4.81e-4 $\pm$ 3.50e-4 & 12.90 \\
      Four-Layer & 2,886 & 8.75e-5 $\pm$ 0.49e-5 & 1.40 & 32,252 & 0.0281 $\pm$ 0.0006 & 16.05 & 17,766 & 4.73e-4 $\pm$ 2.79e-4 & 9.90 \\
      \bottomrule
    \end{tabular}
  }
\end{table}

\subsubsection{Impact of LGM layers}
Grounded in inertial manifold theory, DyMixOp approximates the reduced dynamics by decomposing the evolution into distinct linear ($\mathcal{L}_c$) and nonlinear ($\mathcal{A}(\mathcal{N}_c)$) dynamics. To validate this architectural prior, we conducted an ablation study isolating these terms while maintaining comparable GPU memory budgets, as detailed in Table \ref{tab:dynamics_ablation}.
Specifically, we establish the backbone equipped with only the nonlinear LGM transformation ($\mathcal{N}$) as the baseline, progressively introducing residual operator $\mathcal{A}$ and the linear dynamics branch $\mathcal{L}$ to evaluate the necessity of a composite representation.
The introduction of the residual operator $\mathcal{A}$ yields significant error reductions in the 2D Navier-Stokes (reducing loss from $0.0243$ to $0.0087$) and 2D Darcy (from $7.40e-5$ to $4.40e-5$) benchmarks. This indicates that the explicit modeling of residual operator is crucial for capturing the finer-grained nonlinear interactions unresolved by the LGM transformation.
On the 2D NS and Darcy benchmarks, the optimal performance is achieved only when both components are explicitly modeled. Integrating the linear dynamics branch $\mathcal{L}$ (comprising local convolution and differential kernels) reduces the error by an order of magnitude on 2dNS (reaching $3.68e-4$) and further improves 2D Darcy (reaching $4.05 e-5$). This empirically validates that for benchmarks with strong viscous dissipation or local heterogeneity, accurately resolving the interplay between nonlinear and linear dynamics is critical.
On the 3D SW benchmark, the performance gain is saturated in contrast. The baseline $\mathcal{N}$ alone achieves the best numerical performance (reaching $4.56 e-4$), while adding $\mathcal{A}$ and $\mathcal{L}$ yields slightly degraded results. This suggests that the nonlinearity of the LGM transformation is sufficient to represent the dynamics of this specific regime, and $\mathcal{A}$ and $\mathcal{L}$ enable the network parameters redundant. This is likely caused by the dominance of large-scale advective transport combined with the dataset's use of hyperdiffusion and coarse resolution. This suppresses the fine-scale viscous interactions that the residual operator and explicit linear branch are designed to resolve.
Despite the saturation observed in the 3D SW benchmark, the full decomposition is necessary for retaining universality to solve unknownn dynamics, ensuring the operator is physically complete and robust by explicitly modeling the linear branch.
\begin{table}[htbp]
  \centering
  \caption{\textbf{Ablation study of dynamics configuration.} Comparison of test relative MSE errors and GPU memory consumption across 2D Darcy, 2D Navier-Stokes (2D NS), and 3D Shallow Water (3D SW) benchmarks. Best results are highlighted in \textbf{bold}.}
  \label{tab:dynamics_ablation}
  \vspace{0.2cm}
  \renewcommand{\arraystretch}{1.2}
  \setlength{\tabcolsep}{3pt}
  \resizebox{\textwidth}{!}{
    \begin{tabular}{l c c c c c c c c c}
      \toprule
      & \multicolumn{3}{c}{\textbf{2D Darcy}} & \multicolumn{3}{c}{\textbf{2D NS}} & \multicolumn{3}{c}{\textbf{3D SW}} \\
      \cmidrule(lr){2-4} \cmidrule(lr){5-7} \cmidrule(lr){8-10}
      \textbf{Dynamics} & \textbf{Mem} & \textbf{Test} & \textbf{Time} & \textbf{Mem} & \textbf{Test} & \textbf{Time} & \textbf{Mem} & \textbf{Test} & \textbf{Time} \\
      \textbf{Configuration} & \textbf{(MB)} & \textbf{Loss} & \textbf{(s)} & \textbf{(MB)} & \textbf{Loss} & \textbf{(s)} & \textbf{(MB)} & \textbf{Loss} & \textbf{(s)} \\
      \midrule
      LGM $\mathcal{N}$ & 3,028 & 7.20e-5 $\pm$ 0.28e-5 & 1.30 & 31,766 & 0.0233 $\pm$ 0.0022 & 15.45 & 17,724 & \textbf{4.56e-4 $\pm$ 3.32e-4} & 9.40 \\
      $\mathcal{A}(\mathcal{N})$ & 3,061 & 4.40e-5 $\pm$ 0.00e-5 & 1.70 & 31,576 & 0.0087 $\pm$ 0.0010 & 15.10 & 17,010 & 5.75e-4 $\pm$ 2.36e-4 & 11.70 \\
      $\mathcal{A}(\mathcal{N})$ + $\mathcal{L}$ & 3,030 & \textbf{4.05e-5 $\pm$ 0.21e-5} & 2.25 & 31,344 & \textbf{3.68e-4 $\pm$ 0.38e-4} & 25.20 & 17,370 & 6.04e-4 $\pm$ 4.11e-4 & 33.55 \\
      \bottomrule
    \end{tabular}
  }
\end{table}


\subsubsection{Impact of Dynamics-informed Architecture}
The depth of a neural operator can be interpreted as a discrete realization of the continuous time-evolution operator. To investigate the optimal discretization strategy, we compared four architectural variants: a baseline (standard fixed-step ResNet), adaptive evolution (gated by temporal variation $\mathcal{K}$), parallel architecture (aggregating intermediate dynamics), and a hybrid architecture (combining both adaptive and parallel mechanisms). The results across diverse benchmarks are detailed in Table \ref{tab:arch_ablation}.
Unlike the interaction mechanisms where LGM was universally superior, the optimal architectural topology proves to be highly regime-dependent. There is no single strategy that dominates across all physical systems.
The baseline architecture achieves the lowest errors on the 2D NS (reaching $3.68e-4$) and 2D Darcy (reaching $4.05 e-5$) datasets. This suggests that for systems with stable dissipative structures or steady states, a uniform discretization step is sufficient and the optimization landscape benefits from the simplicity of fixed weights.
For the 1D KS benchmark, the adaptive evolution (reaching $2.62e-3$ ) outperforms the baseline. This aligns with numerical analysis theory, where adaptive step-sizes are required to resolve stiff, multiscale temporal fluctuations.
On the 3D Brusselator benchmark, the parallel architecture achieves the best performance (reaching $3.55e-5$). The reaction-diffusion dynamics appear to benefit significantly from aggregating intermediate dynamical stages, analogous to multi-stage time-stepping schemes.
On the 3D SW benchmark, the hybrid architecture achieves the best performance (reaching $5.36e-4$). The combination of adaptive gating and parallel aggregation appears necessary to capture the complex, wave-dominated atmospheric dynamics effectively.
To maintain a robust dynamics prior suitable for general benchmarks, we select the hybrid architecture as the standard configuration for DyMixOp. Although it may slightly underperform specialized architectures in simpler regimes, it offers the most consistent reliability and avoids the stability failure modes. This "generalist" capability ensures the neural operator remains effective across a broad and unforeseen spectrum of PDEs.

\begin{table}[htbp]
  \centering
  \caption{\textbf{Ablation study of dynamics-informed architecture.} Comparison of test relative MSE errors and GPU memory consumption across 1D Kuramoto-Sivashinsky (1D KS), 2D Darcy, 2D Navier-Stokes (2D NS), 3D Shallow Water (3D SW), and 3D Brusselator (3D Brussel.) benchmarks. Best results are highlighted in \textbf{bold}.}
  \label{tab:arch_ablation}
  \vspace{0.2cm}
  \renewcommand{\arraystretch}{1.2}
  \setlength{\tabcolsep}{2pt}
  \resizebox{\textwidth}{!}{
    \begin{tabular}{l c c c c c c c c c c}
      \toprule
      & \multicolumn{2}{c}{\textbf{1D KS}} & \multicolumn{2}{c}{\textbf{2D Darcy}} & \multicolumn{2}{c}{\textbf{2D NS}} & \multicolumn{2}{c}{\textbf{3D SW}} & \multicolumn{2}{c}{\textbf{3D Brussel.}} \\
      \cmidrule(lr){2-3} \cmidrule(lr){4-5} \cmidrule(lr){6-7} \cmidrule(lr){8-9} \cmidrule(lr){10-11}
      \textbf{Architecture} & \textbf{Mem} & \textbf{Test} & \textbf{Mem} & \textbf{Test} & \textbf{Mem} & \textbf{Test} & \textbf{Mem} & \textbf{Test} & \textbf{Mem} & \textbf{Test} \\
      \textbf{Type}         & \textbf{(MB)} & \textbf{Loss} & \textbf{(MB)} & \textbf{Loss} & \textbf{(MB)} & \textbf{Loss} & \textbf{(MB)} & \textbf{Loss} & \textbf{(MB)} & \textbf{Loss} \\
      \midrule
      Baseline (Fixed Step)   & 2,678 & 2.68e-3 $\pm$ 0.04e-3 & 3,030 & \textbf{4.05e-5 $\pm$ 0.21e-5} & 31,344 & \textbf{3.68e-4 $\pm$ 0.38e-4} & 17,370 & 6.04e-4 $\pm$ 4.11e-4 & 33,174 & 9.70e-5 $\pm$ 3.39e-5 \\
      Adaptive Evolution      & 2,843 & \textbf{2.62e-3 $\pm$ 0.17e-3} & 2,860 & 4.45e-5 $\pm$ 0.35e-5 & 31,742 & 4.30e-4 $\pm$ 0.49e-4 & 16,987 & 7.40e-4 $\pm$ 2.94e-4 & 33,136 & 7.35e-5 $\pm$ 0.35e-5 \\
      Parallel Arch.          & 2,633 & 3.24e-3 $\pm$ 1.11e-3 & 3,150 & 5.50e-5 $\pm$ 0.42e-5 & 31,219 & 1.13e-3 $\pm$ 0.95e-3 & 17,087 & 5.53e-4 $\pm$ 3.63e-4 & 35,936 & \textbf{3.55e-5 $\pm$ 0.21e-5} \\
      Hybrid Arch.            & 2,776 & 3.16e-3 $\pm$ 0.90e-3 & 3,011 & 5.50e-5 $\pm$ 0.42e-5 & 32,393 & 4.07e-4 $\pm$ 0.09e-4 & 17,234 & \textbf{5.36e-4 $\pm$ 2.34e-4} & 35,990 & 3.60e-5 $\pm$ 0.28e-5 \\
      \bottomrule
    \end{tabular}
  }
\end{table}

\section{Conclusions}
This work presents DyMixOp, a dynamics-consistent neural operator framework for solving partial differential equations (PDEs) that unifies theoretical insights from complex dynamical systems with practical deep learning design. At its core, DyMixOp is grounded in inertial manifold theory, which posits that the long-term behavior of dissipative PDEs can be captured by a finite-dimensional reduced system. Leveraging this principle, we explicitly construct a latent representation where infinite-dimensional physical dynamics are projected onto a low-dimensional coordinate space, enabling efficient and physically meaningful approximation without sacrificing essential nonlinear interactions.

A key architectural innovation is the local-global mixing (LGM) layer, which is inspired by the convective nonlinearity inherent in turbulent flows, specifically the multiplicative coupling between local velocity and global gradient fields. By fusing local fine-scale features with global spectral information through an element-wise product, LGM transcends the limitations of additive fusion schemes (e.g., LGA) and mitigates spectral bias. This mechanism not only recovers high-frequency components typically lost in spectral truncation but also endows the model with enhanced expressivity for capturing multiscale, chaotic, and shock-driven phenomena.
The overall architecture integrates these components within a dynamics-informed framework, where network depth corresponds to discrete temporal evolution. Through timescale-adaptive gating and parallel aggregation of intermediate dynamics, DyMixOp achieves stable and accurate integration across diverse physical regimes—from steady elliptic systems to chaotic parabolic and hyperbolic PDEs.

Comprehensive experiments across seven benchmarks (1D–3D) demonstrate that DyMixOp consistently achieves state-of-the-art accuracy, with error reductions up to 94.3$\%$ over existing methods on highly nonlinear systems such as the Kuramoto–Sivashinsky equation. Moreover, it exhibits superior computational efficiency and monotonic scaling, establishing a new Pareto frontier in the accuracy–efficiency trade-off.

In summary, DyMixOp demonstrates that embedding inertial manifold inspired dimension reduction and physics-motivated LGM layers into neural operator design yields a robust, general-purpose solver that bridges the gap between mathematical theory and data-driven learning. This approach paves the way for next-generation scientific machine learning models that are not only predictive but also interpretable, scalable, and rooted in the fundamental mechanics of complex systems.

\section*{Supplementary Materials}
\textbf{Technical Appendices and Supplementary Material.} This document includes the following sections:
\begin{itemize}
  \item \textbf{A.1 Limitation of Spectral Global Transformations} — Provides a theoretical proof and numerical validation showing that spectral truncation inherently prevents global operators from reconstructing high-frequency components.
  \item \textbf{A.2 Advantage of LGM Transformations} — Establishes the spectral expansion property of Local-Global Multiplicative (LGM) mixing, proving it can recover frequencies beyond the global truncation limit, supported by spectral and error analysis.
  \item \textbf{A.3 Datasets} — Details seven multi-dimensional PDE benchmarks (1D to 3D), including Kuramoto–Sivashinsky, Burgers, Darcy, Compressible Euler, Navier–Stokes, Shallow Water, and Brusselator, with generation protocols and tensor shapes.
  \item \textbf{A.4 Baselines} — Describes nine neural operator baselines (DeepONet, ConvLSTM, FNO, GNOT, LaMO, LocalNO, U-NO, CoDA-NO, DyMixOp), their architectures, hyperparameters, and categorization into transformation types (Global, LLM, LGA, LGM).
  \item \textbf{A.5 Detailed Results} — Presents comprehensive loss statistics across all model sizes and datasets, demonstrating consistent superiority of DyMixOp over prior methods.
  \item \textbf{A.6 Heuristic Derivation Inspired by Inertial Manifold Theory} — Offers an intuitive reduction of infinite-dimensional PDE dynamics to finite latent-space ODEs, motivating the design of dynamics-informed operators.
  \item \textbf{A.7 Dynamics-Informed Architecture Design} — Compares parallel vs. hierarchical formulations of latent dynamics in network construction, linking time-scale modeling to layer stacking.
  \item \textbf{A.8 Visualization of baselines across benchmarks} — Presents qualitative visual comparisons of all baseline models and DyMixOp on representative test samples from each PDE benchmark.
  \item \textbf{A.9 Zero-shot super-resolution results} — Reports zero-shot super-resolution results under multiple refinement factors, and analyzes the effect of differential kernels as well as the robustness brought by LGM transformations, LGM layers and dynamics-informed architecture designs.
  \item \textbf{A.10 Limitations and Discussion} — Discusses the absence of a formal Universal Approximation Theorem guarantee and challenges on irregular grids, along with potential remedies.
\end{itemize}

\begin{sloppypar}
  \textbf{Animations.} The supplementary animations provide qualitative, time-resolved comparisons between DyMixOp predictions and the reference solutions on representative test samples. For each benchmark, we visualize the temporal evolution of the solution fields (or selected channels) over the rollout horizon.
  \begin{itemize}
    \item \textbf{Animation S1 (1D Kuramoto--Sivashinsky).} One-channel rollout example: \texttt{\path{1dKS_Animation_Batch_0_Channel_0.gif}}.
    \item \textbf{Animation S2 (2D Burgers).} Two-channel example (e.g., two velocity components): \texttt{\path{2dBurgers_Animation_Batch_-1_Channel_0.gif}}, \texttt{\path{2dBurgers_Animation_Batch_-1_Channel_1.gif}}.
    \item \textbf{Animation S3 (2D Compressible Euler, CRP).} Five-channel visualization (e.g., $\rho$, $u$, $v$, $p$, $E$): \texttt{\path{2dCE-CRP_Animation_Batch_0_Channel_0.gif}} $\sim$ \texttt{\path{2dCE-CRP_Animation_Batch_0_Channel_4.gif}}.
    \item \textbf{Animation S4 (2D Navier--Stokes).} Single-channel example (e.g., vorticity): \texttt{\path{2dNS_Animation_Batch_-1_Channel_0.gif}}.
    \item \textbf{Animation S5 (3D Shallow Water).} Two-channel example (e.g., free-surface height and vorticity): \texttt{\path{3dSW_Animation_Batch_-1_Channel_0.gif}}, \texttt{\path{3dSW_Animation_Batch_-1_Channel_1.gif}}.
  \end{itemize}
\end{sloppypar}

\section*{Acknowledgments}
This work is supported by the Natural Science Foundation of China (No. 92270109). The authors would like to thank Dr. Jing Wang (School of Aeronautics and Astronautics, Shanghai Jiao Tong University) for helpful discussions.

\section*{Declarations}

The authors declare no conflict of interest.


\section*{Data Availability Statement}
The code has been open-sourced in \href{https://github.com/Lain-PY/DyMixOp}{https://github.com/Lain-PY/DyMixOp}.

\section*{Declaration of generative AI and AI-assisted technologies in the manuscript preparation process}
During the preparation of this work the authors used Gemini (Google LLC) and Qwen (Alibaba Cloud) in order to assist with language polishing and organization of textual content. After using these tools, the authors thoroughly reviewed and edited the generated output, ensuring its scientific accuracy, coherence, and alignment with the original research. The authors take full responsibility for the content of the published article.

\bibliographystyle{elsarticle-num}
\bibliography{reference}

@article{breakspear2017dynamic,
  title={Dynamic models of large-scale brain activity},
  author={Breakspear, Michael},
  journal={Nature neuroscience},
  volume={20},
  number={3},
  pages={340--352},
  year={2017},
  publisher={Nature Publishing Group}
}

@article{bi2023accurate,
  title={Accurate medium-range global weather forecasting with 3D neural networks},
  author={Bi, Kaifeng and Xie, Lingxi and Zhang, Hengheng and Chen, Xin and Gu, Xiaotao and Tian, Qi},
  journal={Nature},
  volume={619},
  number={7970},
  pages={533--538},
  year={2023},
  publisher={Nature Publishing Group UK London}
}

@article{blasius1999complex,
  title={Complex dynamics and phase synchronization in spatially extended ecological systems},
  author={Blasius, Bernd and Huppert, Amit and Stone, Lewi},
  journal={Nature},
  volume={399},
  number={6734},
  pages={354--359},
  year={1999},
  publisher={Nature Publishing Group UK London}
}

@article{mukherjee2023intermittency,
  title={Intermittency, fluctuations and maximal chaos in an emergent universal state of active turbulence},
  author={Mukherjee, Siddhartha and Singh, Rahul K and James, Martin and Ray, Samriddhi Sankar},
  journal={Nature Physics},
  pages={1--7},
  year={2023},
  publisher={Nature Publishing Group UK London}
}

@article{lam2023learning,
  title={Learning skillful medium-range global weather forecasting},
  author={Lam, Remi and Sanchez-Gonzalez, Alvaro and Willson, Matthew and Wirnsberger, Peter and Fortunato, Meire and Alet, Ferran and Ravuri, Suman and Ewalds, Timo and Eaton-Rosen, Zach and Hu, Weihua and others},
  journal={Science},
  pages={eadi2336},
  year={2023},
  publisher={American Association for the Advancement of Science}
}

@book{jensen2017introduction,
  title={Introduction to computational chemistry},
  author={Jensen, Frank},
  year={2017},
  publisher={John wiley \& sons}
}

@article{lagaris1998artificial,
  title={Artificial neural networks for solving ordinary and partial differential equations},
  author={Lagaris, Isaac E and Likas, Aristidis and Fotiadis, Dimitrios I},
  journal={IEEE transactions on neural networks},
  volume={9},
  number={5},
  pages={987--1000},
  year={1998},
  publisher={IEEE}
}

@article{kovachki2023neural,
  title={Neural operator: Learning maps between function spaces with applications to pdes},
  author={Kovachki, Nikola and Li, Zongyi and Liu, Burigede and Azizzadenesheli, Kamyar and Bhattacharya, Kaushik and Stuart, Andrew and Anandkumar, Anima},
  journal={Journal of Machine Learning Research},
  volume={24},
  number={89},
  pages={1--97},
  year={2023}
}

@article{lu2021learning,
  title={Learning nonlinear operators via DeepONet based on the universal approximation theorem of operators},
  author={Lu, Lu and Jin, Pengzhan and Pang, Guofei and Zhang, Zhongqiang and Karniadakis, George Em},
  journal={Nature machine intelligence},
  volume={3},
  number={3},
  pages={218--229},
  year={2021},
  publisher={Nature Publishing Group UK London}
}

@article{psichogios1992hybrid,
  title={A hybrid neural network-first principles approach to process modeling},
  author={Psichogios, Dimitris C and Ungar, Lyle H},
  journal={AIChE Journal},
  volume={38},
  number={10},
  pages={1499--1511},
  year={1992},
  publisher={Wiley Online Library}
}

@article{lecun2015deep,
  title={Deep learning},
  author={LeCun, Yann and Bengio, Yoshua and Hinton, Geoffrey},
  journal={nature},
  volume={521},
  number={7553},
  pages={436--444},
  year={2015},
  publisher={Nature Publishing Group UK London}
}

@article{raissi2019physics,
  title={Physics-informed neural networks: A deep learning framework for solving forward and inverse problems involving nonlinear partial differential equations},
  author={Raissi, Maziar and Perdikaris, Paris and Karniadakis, George E},
  journal={Journal of Computational physics},
  volume={378},
  pages={686--707},
  year={2019},
  publisher={Elsevier}
}

@article{karniadakis2021physics,
  title={Physics-informed machine learning},
  author={Karniadakis, George Em and Kevrekidis, Ioannis G and Lu, Lu and Perdikaris, Paris and Wang, Sifan and Yang, Liu},
  journal={Nature Reviews Physics},
  volume={3},
  number={6},
  pages={422--440},
  year={2021},
  publisher={Nature Publishing Group}
}

@article{buaria2023forecasting,
  title={Forecasting small-scale dynamics of fluid turbulence using deep neural networks},
  author={Buaria, Dhawal and Sreenivasan, Katepalli R},
  journal={Proceedings of the National Academy of Sciences},
  volume={120},
  number={30},
  pages={e2305765120},
  year={2023},
  publisher={National Acad Sciences}
}

@article{ling2016reynolds,
  title={Reynolds averaged turbulence modelling using deep neural networks with embedded invariance},
  author={Ling, Julia and Kurzawski, Andrew and Templeton, Jeremy},
  journal={Journal of Fluid Mechanics},
  volume={807},
  pages={155--166},
  year={2016},
  publisher={Cambridge University Press}
}

@article{yu2018deep,
  title={The deep Ritz method: a deep learning-based numerical algorithm for solving variational problems},
  author={Yu, Bing and others},
  journal={Communications in Mathematics and Statistics},
  volume={6},
  number={1},
  pages={1--12},
  year={2018},
  publisher={Springer}
}

@book{trefethen2000spectral,
  title={Spectral methods in MATLAB},
  author={Trefethen, Lloyd N},
  year={2000},
  publisher={SIAM}
}

@article{li2020fourier,
  title={Fourier neural operator for parametric partial differential equations},
  author={Li, Zongyi and Kovachki, Nikola and Azizzadenesheli, Kamyar and Liu, Burigede and Bhattacharya, Kaushik and Stuart, Andrew and Anandkumar, Anima},
  journal={arXiv preprint arXiv:2010.08895},
  year={2020}
}

@article{tripura2022wavelet,
  title={Wavelet neural operator: a neural operator for parametric partial differential equations},
  author={Tripura, Tapas and Chakraborty, Souvik},
  journal={arXiv preprint arXiv:2205.02191},
  year={2022}
}

@article{cao2024laplace,
  title={Laplace neural operator for solving differential equations},
  author={Cao, Qianying and Goswami, Somdatta and Karniadakis, George Em},
  journal={Nature Machine Intelligence},
  volume={6},
  number={6},
  pages={631--640},
  year={2024},
  publisher={Nature Publishing Group UK London}
}

@article{kim2024prediction,
  title={Prediction and control of two-dimensional decaying turbulence using generative adversarial networks},
  author={Kim, Jiyeon and Kim, Junhyuk and Lee, Changhoon},
  journal={Journal of Fluid Mechanics},
  volume={981},
  pages={A19},
  year={2024},
  publisher={Cambridge University Press}
}

@article{kim2021unsupervised,
  title={Unsupervised deep learning for super-resolution reconstruction of turbulence},
  author={Kim, Hyojin and Kim, Junhyuk and Won, Sungjin and Lee, Changhoon},
  journal={Journal of Fluid Mechanics},
  volume={910},
  pages={A29},
  year={2021},
  publisher={Cambridge University Press}
}

@article{qu2022learning,
  title={Learning time-dependent PDEs with a linear and nonlinear separate convolutional neural network},
  author={Qu, Jiagang and Cai, Weihua and Zhao, Yijun},
  journal={Journal of Computational Physics},
  volume={453},
  pages={110928},
  year={2022},
  publisher={Elsevier}
}

@article{gao2021phygeonet,
  title={PhyGeoNet: Physics-informed geometry-adaptive convolutional neural networks for solving parameterized steady-state PDEs on irregular domain},
  author={Gao, Han and Sun, Luning and Wang, Jian-Xun},
  journal={Journal of Computational Physics},
  volume={428},
  pages={110079},
  year={2021},
  publisher={Elsevier}
}

@article{xuan2023reconstruction,
  title={Reconstruction of three-dimensional turbulent flow structures using surface measurements for free-surface flows based on a convolutional neural network},
  author={Xuan, Anqing and Shen, Lian},
  journal={Journal of Fluid Mechanics},
  volume={959},
  pages={A34},
  year={2023},
  publisher={Cambridge University Press}
}

@article{ren2022phycrnet,
  title={PhyCRNet: Physics-informed convolutional-recurrent network for solving spatiotemporal PDEs},
  author={Ren, Pu and Rao, Chengping and Liu, Yang and Wang, Jian-Xun and Sun, Hao},
  journal={Computer Methods in Applied Mechanics and Engineering},
  volume={389},
  pages={114399},
  year={2022},
  publisher={Elsevier}
}

@article{list2022learned,
  title={Learned turbulence modelling with differentiable fluid solvers: physics-based loss functions and optimisation horizons},
  author={List, Bj{\"o}rn and Chen, Li-Wei and Thuerey, Nils},
  journal={Journal of Fluid Mechanics},
  volume={949},
  pages={A25},
  year={2022},
  publisher={Cambridge University Press}
}

@inproceedings{long2018pde,
  title={Pde-net: Learning pdes from data},
  author={Long, Zichao and Lu, Yiping and Ma, Xianzhong and Dong, Bin},
  booktitle={International conference on machine learning},
  pages={3208--3216},
  year={2018},
  organization={PMLR}
}

@article{rao2023encoding,
  title={Encoding physics to learn reaction--diffusion processes},
  author={Rao, Chengping and Ren, Pu and Wang, Qi and Buyukozturk, Oral and Sun, Hao and Liu, Yang},
  journal={Nature Machine Intelligence},
  volume={5},
  number={7},
  pages={765--779},
  year={2023},
  publisher={Nature Publishing Group UK London}
}

@article{shi2015convolutional,
  title={Convolutional LSTM network: A machine learning approach for precipitation nowcasting},
  author={Shi, Xingjian and Chen, Zhourong and Wang, Hao and Yeung, Dit-Yan and Wong, Wai-Kin and Woo, Wang-chun},
  journal={Advances in neural information processing systems},
  volume={28},
  year={2015}
}

@article{li2020neural,
  title={Neural operator: Graph kernel network for partial differential equations},
  author={Li, Zongyi and Kovachki, Nikola and Azizzadenesheli, Kamyar and Liu, Burigede and Bhattacharya, Kaushik and Stuart, Andrew and Anandkumar, Anima},
  journal={arXiv preprint arXiv:2003.03485},
  year={2020}
}

@article{li2020multipole,
  title={Multipole graph neural operator for parametric partial differential equations},
  author={Li, Zongyi and Kovachki, Nikola and Azizzadenesheli, Kamyar and Liu, Burigede and Stuart, Andrew and Bhattacharya, Kaushik and Anandkumar, Anima},
  journal={Advances in Neural Information Processing Systems},
  volume={33},
  pages={6755--6766},
  year={2020}
}

@article{vaswani2017attention,
  title={Attention is all you need},
  author={Vaswani, A},
  journal={Advances in Neural Information Processing Systems},
  year={2017}
}

@inproceedings{sohl2015deep,
  title={Deep unsupervised learning using nonequilibrium thermodynamics},
  author={Sohl-Dickstein, Jascha and Weiss, Eric and Maheswaranathan, Niru and Ganguli, Surya},
  booktitle={International conference on machine learning},
  pages={2256--2265},
  year={2015},
  organization={PMLR}
}

@article{ho2020denoising,
  title={Denoising diffusion probabilistic models},
  author={Ho, Jonathan and Jain, Ajay and Abbeel, Pieter},
  journal={Advances in neural information processing systems},
  volume={33},
  pages={6840--6851},
  year={2020}
}

@article{hopfield1982neural,
  title={Neural networks and physical systems with emergent collective computational abilities.},
  author={Hopfield, John J},
  journal={Proceedings of the national academy of sciences},
  volume={79},
  number={8},
  pages={2554--2558},
  year={1982},
  publisher={National Acad Sciences}
}

@book{feynman2015feynman,
  title={The Feynman lectures on physics, Vol. I: The new millennium edition: mainly mechanics, radiation, and heat},
  author={Feynman, Richard P and Leighton, Robert B and Sands, Matthew},
  volume={1},
  year={2015},
  publisher={Basic books}
}

@book{dennis1996numerical,
  title={Numerical methods for unconstrained optimization and nonlinear equations},
  author={Dennis Jr, John E and Schnabel, Robert B},
  year={1996},
  publisher={SIAM}
}

@book{rapaport2004art,
  title={The art of molecular dynamics simulation},
  author={Rapaport, Dennis C},
  year={2004},
  publisher={Cambridge university press}
}

@book{succi2001lattice,
  title={The Lattice Boltzmann Equation for Fluid Dynamics and Beyond},
  author={Succi, S},
  year={2001},
  publisher={Oxford University Press}
}

@book{howell2020thermal,
  title={Thermal radiation heat transfer},
  author={Howell, John R and Meng{\"u}{\c{c}}, M Pinar and Daun, Kyle and Siegel, Robert},
  year={2020},
  publisher={CRC press}
}

@article{moin1998direct,
  title={Direct numerical simulation: a tool in turbulence research},
  author={Moin, Parviz and Mahesh, Krishnan},
  journal={Annual review of fluid mechanics},
  volume={30},
  number={1},
  pages={539--578},
  year={1998},
  publisher={Annual Reviews 4139 El Camino Way, PO Box 10139, Palo Alto, CA 94303-0139, USA}
}

@book{smith1985numerical,
  title={Numerical solution of partial differential equations: finite difference methods},
  author={Smith, Gordon D},
  year={1985},
  publisher={Oxford university press}
}

@book{versteeg2007introduction,
  title={An introduction to computational fluid dynamics the finite volume method, 2/E},
  author={Versteeg, Henk Kaarle},
  year={2007},
  publisher={Pearson Education India}
}

@misc{zienkiewicz1971finite,
  title={The finite element method in engineering science},
  author={Zienkiewicz, OC},
  year={1971},
  publisher={McGraw-Hill}
}

@article{jordan2015machine,
  title={Machine learning: Trends, perspectives, and prospects},
  author={Jordan, Michael I and Mitchell, Tom M},
  journal={Science},
  volume={349},
  number={6245},
  pages={255--260},
  year={2015},
  publisher={American Association for the Advancement of Science}
}

@article{lu2021physics,
  title={Physics-informed neural networks with hard constraints for inverse design},
  author={Lu, Lu and Pestourie, Raphael and Yao, Wenjie and Wang, Zhicheng and Verdugo, Francesc and Johnson, Steven G},
  journal={SIAM Journal on Scientific Computing},
  volume={43},
  number={6},
  pages={B1105--B1132},
  year={2021},
  publisher={SIAM}
}

@article{azizzadenesheli2024neural,
  title={Neural operators for accelerating scientific simulations and design},
  author={Azizzadenesheli, Kamyar and Kovachki, Nikola and Li, Zongyi and Liu-Schiaffini, Miguel and Kossaifi, Jean and Anandkumar, Anima},
  journal={Nature Reviews Physics},
  pages={1--9},
  year={2024},
  publisher={Nature Publishing Group UK London}
}

@article{pang2019fpinns,
  title={fPINNs: Fractional physics-informed neural networks},
  author={Pang, Guofei and Lu, Lu and Karniadakis, George Em},
  journal={SIAM Journal on Scientific Computing},
  volume={41},
  number={4},
  pages={A2603--A2626},
  year={2019},
  publisher={SIAM}
}

@article{hornik1989multilayer,
  title={Multilayer feedforward networks are universal approximators},
  author={Hornik, Kurt and Stinchcombe, Maxwell and White, Halbert},
  journal={Neural networks},
  volume={2},
  number={5},
  pages={359--366},
  year={1989},
  publisher={Elsevier}
}

@article{hendrycks2016gaussian,
  title={Gaussian error linear units (gelus)},
  author={Hendrycks, Dan and Gimpel, Kevin},
  journal={arXiv preprint arXiv:1606.08415},
  year={2016}
}

@inproceedings{he2016deep,
  title={Deep residual learning for image recognition},
  author={He, Kaiming and Zhang, Xiangyu and Ren, Shaoqing and Sun, Jian},
  booktitle={Proceedings of the IEEE conference on computer vision and pattern recognition},
  pages={770--778},
  year={2016}
}

@article{loshchilov2017decoupled,
  title={Decoupled weight decay regularization},
  author={Loshchilov, Ilya and Hutter, Frank},
  journal={arXiv preprint arXiv:1711.05101},
  year={2017}
}

@book{guckenheimer2013nonlinear,
  title={Nonlinear oscillations, dynamical systems, and bifurcations of vector fields},
  author={Guckenheimer, John and Holmes, Philip},
  volume={42},
  year={2013},
  publisher={Springer Science \& Business Media}
}

@article{berkooz1993proper,
  title={The proper orthogonal decomposition in the analysis of turbulent flows},
  author={Berkooz, Gal and Holmes, Philip and Lumley, John L},
  journal={Annual review of fluid mechanics},
  volume={25},
  number={1},
  pages={539--575},
  year={1993},
  publisher={Annual Reviews 4139 El Camino Way, PO Box 10139, Palo Alto, CA 94303-0139, USA}
}

@article{schmid2010dynamic,
  title={Dynamic mode decomposition of numerical and experimental data},
  author={Schmid, Peter J},
  journal={Journal of fluid mechanics},
  volume={656},
  pages={5--28},
  year={2010},
  publisher={Cambridge University Press}
}

@book{isidori1985nonlinear,
  title={Nonlinear control systems: an introduction},
  author={Isidori, Alberto},
  year={1985},
  publisher={Springer}
}

@book{poincare1893methodes,
  title={Les m{\'e}thodes nouvelles de la m{\'e}canique c{\'e}leste: M{\'e}thodes de MM. Newcomb, Gyld{\'e}n, Lindstedt et Bohlin},
  author={Poincar{\'e}, Henri},
  volume={2},
  year={1893},
  publisher={Gauthier-Villars}
}

@book{murdock2003normal,
  title={Normal forms and unfoldings for local dynamical systems},
  author={Murdock, James A},
  year={2003},
  publisher={Springer}
}

@book{lasota2013chaos,
  title={Chaos, fractals, and noise: stochastic aspects of dynamics},
  author={Lasota, Andrzej and Mackey, Michael C},
  volume={97},
  year={2013},
  publisher={Springer Science \& Business Media}
}

@article{foias1988inertial,
  title={Inertial manifolds for nonlinear evolutionary equations},
  author={Foias, Ciprian and Sell, George R and Temam, Roger},
  journal={Journal of differential equations},
  volume={73},
  number={2},
  pages={309--353},
  year={1988},
  publisher={Elsevier}
}

@inproceedings{sanchez2020learning,
  title={Learning to simulate complex physics with graph networks},
  author={Sanchez-Gonzalez, Alvaro and Godwin, Jonathan and Pfaff, Tobias and Ying, Rex and Leskovec, Jure and Battaglia, Peter},
  booktitle={International conference on machine learning},
  pages={8459--8468},
  year={2020},
  organization={PMLR}
}

@article{kochkov2021machine,
  title={Machine learning--accelerated computational fluid dynamics},
  author={Kochkov, Dmitrii and Smith, Jamie A and Alieva, Ayya and Wang, Qing and Brenner, Michael P and Hoyer, Stephan},
  journal={Proceedings of the National Academy of Sciences},
  volume={118},
  number={21},
  pages={e2101784118},
  year={2021},
  publisher={National Academy of Sciences}
}

@article{duraisamy2019turbulence,
  title={Turbulence modeling in the age of data},
  author={Duraisamy, Karthik and Iaccarino, Gianluca and Xiao, Heng},
  journal={Annual review of fluid mechanics},
  volume={51},
  number={1},
  pages={357--377},
  year={2019},
  publisher={Annual Reviews}
}

@book{temam2012infinite,
  title={Infinite-dimensional dynamical systems in mechanics and physics},
  author={Temam, Roger},
  volume={68},
  year={2012},
  publisher={Springer Science \& Business Media}
}

@article{liu2024neural,
  title={Neural operators with localized integral and differential kernels},
  author={Liu-Schiaffini, Miguel and Berner, Julius and Bonev, Boris and Kurth, Thorsten and Azizzadenesheli, Kamyar and Anandkumar, Anima},
  journal={arXiv preprint arXiv:2402.16845},
  year={2024}
}

@article{ocampo2022scalable,
  title={Scalable and equivariant spherical CNNs by discrete-continuous (DISCO) convolutions},
  author={Ocampo, Jeremy and Price, Matthew A and McEwen, Jason D},
  journal={arXiv preprint arXiv:2209.13603},
  year={2022}
}

@article{raonic2023convolutional,
  title={Convolutional neural operators for robust and accurate learning of PDEs},
  author={Raonic, Bogdan and Molinaro, Roberto and De Ryck, Tim and Rohner, Tobias and Bartolucci, Francesca and Alaifari, Rima and Mishra, Siddhartha and de B{\'e}zenac, Emmanuel},
  journal={Advances in Neural Information Processing Systems},
  volume={36},
  pages={77187--77200},
  year={2023}
}

@article{bartolucci2023representation,
  title={Representation equivalent neural operators: a framework for alias-free operator learning},
  author={Bartolucci, Francesca and de Bezenac, Emmanuel and Raonic, Bogdan and Molinaro, Roberto and Mishra, Siddhartha and Alaifari, Rima},
  journal={Advances in Neural Information Processing Systems},
  volume={36},
  pages={69661--69672},
  year={2023}
}

@inproceedings{gao2025discretizationinvariance,
	title={Discretization-invariance? On the Discretization Mismatch Errors in Neural Operators},
	author={Wenhan Gao and Ruichen Xu and Yuefan Deng and Yi Liu},
	booktitle={The Thirteenth International Conference on Learning Representations},
	year={2025},
	url={https://openreview.net/forum?id=J9FgrqOOni}
}

@article{gupta2021multiwavelet,
  title={Multiwavelet-based operator learning for differential equations},
  author={Gupta, Gaurav and Xiao, Xiongye and Bogdan, Paul},
  journal={Advances in neural information processing systems},
  volume={34},
  pages={24048--24062},
  year={2021}
}

@article{cao2021choose,
  title={Choose a transformer: Fourier or galerkin},
  author={Cao, Shuhao},
  journal={Advances in neural information processing systems},
  volume={34},
  pages={24924--24940},
  year={2021}
}

@article{li2022transformer,
  title={Transformer for partial differential equations' operator learning},
  author={Li, Zijie and Meidani, Kazem and Farimani, Amir Barati},
  journal={arXiv preprint arXiv:2205.13671},
  year={2022}
}

@inproceedings{hao2023gnot,
  title={Gnot: A general neural operator transformer for operator learning},
  author={Hao, Zhongkai and Wang, Zhengyi and Su, Hang and Ying, Chengyang and Dong, Yinpeng and Liu, Songming and Cheng, Ze and Song, Jian and Zhu, Jun},
  booktitle={International Conference on Machine Learning},
  pages={12556--12569},
  year={2023},
  organization={PMLR}
}

@article{wu2024transolver,
  title={Transolver: A fast transformer solver for pdes on general geometries},
  author={Wu, Haixu and Luo, Huakun and Wang, Haowen and Wang, Jianmin and Long, Mingsheng},
  journal={arXiv preprint arXiv:2402.02366},
  year={2024}
}

@inproceedings{rahaman2019spectral,
  title={On the spectral bias of neural networks},
  author={Rahaman, Nasim and Baratin, Aristide and Arpit, Devansh and Draxler, Felix and Lin, Min and Hamprecht, Fred and Bengio, Yoshua and Courville, Aaron},
  booktitle={International conference on machine learning},
  pages={5301--5310},
  year={2019},
  organization={PMLR}
}

@article{han2018solving,
  title={Solving high-dimensional partial differential equations using deep learning},
  author={Han, Jiequn and Jentzen, Arnulf and E, Weinan},
  journal={Proceedings of the National Academy of Sciences},
  volume={115},
  number={34},
  pages={8505--8510},
  year={2018},
  publisher={National Academy of Sciences}
}

@inproceedings{gu2024mamba,
  title={Mamba: Linear-time sequence modeling with selective state spaces},
  author={Gu, Albert and Dao, Tri},
  booktitle={First conference on language modeling},
  year={2024}
}

@article{dao2024transformers,
  title={Transformers are ssms: Generalized models and efficient algorithms through structured state space duality},
  author={Dao, Tri and Gu, Albert},
  journal={arXiv preprint arXiv:2405.21060},
  year={2024}
}

@article{hu2025deepomamba,
  title={Deepomamba: State-space model for spatio-temporal pde neural operator learning},
  author={Hu, Zheyuan and Cao, Qianying and Kawaguchi, Kenji and Karniadakis, George Em},
  journal={Journal of Computational Physics},
  pages={114272},
  year={2025},
  publisher={Elsevier}
}

@article{tiwari2025latent,
  title={Latent Mamba Operator for Partial Differential Equations},
  author={Tiwari, Karn and Dutta, Niladri and Krishnan, NM and others},
  journal={arXiv preprint arXiv:2505.19105},
  year={2025}
}

@article{han2025geomano,
  title={GeoMaNO: Geometric Mamba Neural Operator for Partial Differential Equations},
  author={Han, Xi and Zhang, Jingwei and Samaras, Dimitris and Hou, Fei and Qin, Hong},
  journal={arXiv preprint arXiv:2505.12020},
  year={2025}
}

@article{lai2024neural,
  title={Neural downscaling for complex systems: from large-scale to small-scale by neural operator},
  author={Lai, Pengyu and Wang, Jing and Wang, Rui and Yang, Dewu and Fei, Haoqi and Chen, Yihe and Xu, Hui},
  journal={Engineering Applications of Computational Fluid Mechanics},
  volume={18},
  number={1},
  pages={2399672},
  year={2024},
  publisher={Taylor \& Francis}
}

@article{rahman2022uno,
  title     = {{U-NO: U-shaped Neural Operators}},
  author    = {Rahman, Md Ashiqur and Ross, Zachary E. and Azizzadenesheli, Kamyar},
  journal   = {arXiv preprint arXiv:2204.11127},
  year      = {2022}
}

@article{rahman2024pretraining,
  title={Pretraining codomain attention neural operators for solving multiphysics pdes},
  author={Rahman, Md Ashiqur and George, Robert Joseph and Elleithy, Mogab and Leibovici, Daniel and Li, Zongyi and Bonev, Boris and White, Colin and Berner, Julius and Yeh, Raymond A and Kossaifi, Jean and others},
  journal={Advances in Neural Information Processing Systems},
  volume={37},
  pages={104035--104064},
  year={2024}
}

@article{xiong2024koopman,
  title={Koopman neural operator as a mesh-free solver of non-linear partial differential equations},
  author={Xiong, Wei and Huang, Xiaomeng and Zhang, Ziyang and Deng, Ruixuan and Sun, Pei and Tian, Yang},
  journal={Journal of Computational Physics},
  volume={513},
  pages={113194},
  year={2024},
  publisher={Elsevier}
}

@article{champion2019data,
  title={Data-driven discovery of coordinates and governing equations},
  author={Champion, Kathleen and Lusch, Bethany and Kutz, J Nathan and Brunton, Steven L},
  journal={Proceedings of the National Academy of Sciences},
  volume={116},
  number={45},
  pages={22445--22451},
  year={2019},
  publisher={National Academy of Sciences}
}

@article{floryan2022data,
  title={Data-driven discovery of intrinsic dynamics},
  author={Floryan, Daniel and Graham, Michael D},
  journal={Nature Machine Intelligence},
  volume={4},
  number={12},
  pages={1113--1120},
  year={2022},
  publisher={Nature Publishing Group UK London}
}

@inproceedings{wu2024neural,
  title={Neural manifold operators for learning the evolution of physical dynamics},
  author={Wu, Hao and Weng, Kangyu and Zhou, Shuyi and Huang, Xiaomeng and Xiong, Wei},
  booktitle={Proceedings of the 30th ACM SIGKDD Conference on Knowledge Discovery and Data Mining},
  pages={3356--3366},
  year={2024}
}

@article{weinan2017proposal,
  title={A proposal on machine learning via dynamical systems},
  author={Weinan, Ee},
  journal={Communications in Mathematics and Statistics},
  volume={5},
  number={1},
  pages={1--11},
  year={2017},
  publisher={Springer Science and Business Media LLC}
}

@book{hairer1993solving,
  title={Solving ordinary differential equations I: Nonstiff problems},
  author={Hairer, Ernst and Wanner, Gerhard and N{\o}rsett, Syvert P},
  year={1993},
  publisher={Springer}
}

@article{li2023geometry,
  title={Geometry-informed neural operator for large-scale 3d pdes},
  author={Li, Zongyi and Kovachki, Nikola and Choy, Chris and Li, Boyi and Kossaifi, Jean and Otta, Shourya and Nabian, Mohammad Amin and Stadler, Maximilian and Hundt, Christian and Azizzadenesheli, Kamyar and others},
  journal={Advances in Neural Information Processing Systems},
  volume={36},
  pages={35836--35854},
  year={2023}
}

@article{mcgreivy2024weak,
  title={Weak baselines and reporting biases lead to overoptimism in machine learning for fluid-related partial differential equations},
  author={McGreivy, Nick and Hakim, Ammar},
  journal={Nature machine intelligence},
  volume={6},
  number={10},
  pages={1256--1269},
  year={2024},
  publisher={Nature Publishing Group UK London}
}

@article{herde2024poseidon,
  title={Poseidon: Efficient foundation models for pdes},
  author={Herde, Maximilian and Raonic, Bogdan and Rohner, Tobias and K{\"a}ppeli, Roger and Molinaro, Roberto and de B{\'e}zenac, Emmanuel and Mishra, Siddhartha},
  journal={Advances in Neural Information Processing Systems},
  volume={37},
  pages={72525--72624},
  year={2024}
}

@article{zwicker2020py,
  title={py-pde: A python package for solving partial differential equations},
  author={Zwicker, David},
  journal={Journal of Open Source Software},
  volume={5},
  number={48},
  pages={2158},
  year={2020}
}

@article{galewsky2004initial,
  title={An initial-value problem for testing numerical models of the global shallow-water equations},
  author={Galewsky, Joseph and Scott, Richard K and Polvani, Lorenzo M},
  journal={Tellus A: Dynamic Meteorology and Oceanography},
  volume={56},
  number={5},
  pages={429--440},
  year={2004},
  publisher={Taylor \& Francis}
}

@article{lu2022comprehensive,
  title={A comprehensive and fair comparison of two neural operators (with practical extensions) based on fair data},
  author={Lu, Lu and Meng, Xuhui and Cai, Shengze and Mao, Zhiping and Goswami, Somdatta and Zhang, Zhongqiang and Karniadakis, George Em},
  journal={Computer Methods in Applied Mechanics and Engineering},
  volume={393},
  pages={114778},
  year={2022},
  publisher={Elsevier}
}

@article{yu2019review,
  title={A review of recurrent neural networks: LSTM cells and network architectures},
  author={Yu, Yong and Si, Xiaosheng and Hu, Changhua and Zhang, Jianxun},
  journal={Neural computation},
  volume={31},
  number={7},
  pages={1235--1270},
  year={2019},
  publisher={MIT Press One Rogers Street, Cambridge, MA 02142-1209, USA journals-info~…}
}

@article{li2023fourier,
  title={Fourier neural operator with learned deformations for pdes on general geometries},
  author={Li, Zongyi and Huang, Daniel Zhengyu and Liu, Burigede and Anandkumar, Anima},
  journal={Journal of Machine Learning Research},
  volume={24},
  number={388},
  pages={1--26},
  year={2023}
}

@article{kovachki2021universal,
  title={On universal approximation and error bounds for Fourier neural operators},
  author={Kovachki, Nikola and Lanthaler, Samuel and Mishra, Siddhartha},
  journal={Journal of Machine Learning Research},
  volume={22},
  number={290},
  pages={1--76},
  year={2021}
}

@article{lu2019deeponet,
  title={Deeponet: Learning nonlinear operators for identifying differential equations based on the universal approximation theorem of operators},
  author={Lu, Lu and Jin, Pengzhan and Karniadakis, George Em},
  journal={arXiv preprint arXiv:1910.03193},
  year={2019}
}

@inproceedings{ronneberger2015unet,
  title     = {{U-Net: Convolutional Networks for Biomedical Image Segmentation}},
  author    = {Ronneberger, Olaf and Fischer, Philipp and Brox, Thomas},
  booktitle = {International Conference on Medical Image Computing and Computer-Assisted Intervention (MICCAI)},
  pages     = {234--241},
  year      = {2015},
  publisher = {Springer, Cham}
}


\end{document}


\begin{frontmatter}
    \title{DyMixOp: A Neural Operator Designed from a Complex Dynamics Perspective with Local-Global-Mixing for solving PDEs}

%

  \author[aff1]{Pengyu Lai}
  \author[aff1]{Yixiao Chen}
  \author[aff1]{Dewu Yang}
  \author[aff1]{Rui Wang}
  \author[aff1]{Feng Wang}
  \author[aff1]{Hui Xu\corref{cor1}}

  \cortext[cor1]{Corresponding author: dr.hxu@sjtu.edu.cn}

  \affiliation[aff1]{%
    organization={School of Aeronautics and Astronautics, Shanghai Jiao Tong University},
    city={Shanghai},
    postcode={200240},
    country={China}
  }

\end{frontmatter}

\appendix
\section{Technical Appendices and Supplementary Material}\label{APPENDIX}

\subsection{Limitation}
In this section, we analyze the inherent limitations of spectral global transformations, specifically demonstrating their inability to capture high-frequency components (HFCs) due to spectral truncation. We provide both a theoretical proof and a supporting numerical validation.

\subsubsection{Theoretical Proof}
\begin{corollary} In a spectral global transformation that involves truncation of high-frequency or high-order spectral coefficients, the resulting truncated representation will fail to capture high-frequency or high-order components of the target function.
\end{corollary}

\begin{proof} Let $f : \Omega \to \mathbb{R}$ (or $\mathbb{C}$ ) be a target function defined over a domain $\Omega \subset \mathbb{R}^d$ , assumed to belong to a suitable function space $L^2(\Omega)$ that allows a spectral decomposition. In general, we can express $f(x)$ in terms of a set of orthogonal basis functions $\{\phi_k(x)\}_{k \in \mathbb{N}}$ , which may be Fourier functions, orthogonal polynomials, or any complete orthonormal basis on $\Omega$ :
\begin{equation}
    f(x) = \sum_{k \in \mathbb{N}} \hat{f}(k) \phi_k(x),
\end{equation}
where $\hat{f}(k)$ denotes the spectral coefficients associated with each basis function $\phi_k(x) $. 

For a truncation threshold $K>0$, define a truncation operator $T_K$ that keeps only the coefficients corresponding to basis functions indexed by $k$ with $|k| \leq K$ (where $|k|$ represents a suitable measure of the “order” or “frequency” of $k$ , such as a norm). The truncated coefficients $\hat{f}_T(k)$ are given by
\begin{equation}\label{trunk_coeff}
\hat{f}_T(k)=T_K(\hat{f}(k))= \begin{cases}\hat{f}(k), & \text { if }|k| \leq K, \\ 0, & \text { if }|k|>K .\end{cases}
\end{equation}
The truncated representation $f_T(x)$ of $f(x)$ is then given by
\begin{equation}
f_T(x) = \sum_{|k| \leq K} \hat{f}(k) \phi_k(x).    
\end{equation}
Define the approximation error $E(x)$ as the difference between the original function $f(x)$ and the truncated function $f_T(x)$ :
\begin{equation}
    E(x) = f(x) - f_T(x) = \sum_{|k| > K} \hat{f}(k) \phi_k(x).
\end{equation}
By Parseval’s theorem (or the equivalent norm preservation property for general orthogonal bases), the $L^2$ norm of $E(x)$ is given by
\begin{equation}
\|E\|_{L^2(\Omega)}^2=\int_\Omega |E(x)|^2 d x=\sum_{|k|>K}|\hat{f}(k)|^2 .
\end{equation}
The magnitude of $\|E\|_{L^2(\Omega)}^2$ depends directly on the truncation threshold $K$, implying a greater loss of information from the omitted high-order components. Therefore, truncating the spectral coefficients at a finite $K$ restricts the method’s ability to approximate any component of $f(x)$ associated with $|k| > K$ , which limits its capacity to represent high-frequency or high-order features of the target function. This completes the proof.
\end{proof}

\subsubsection{Numerical Validation}
\begin{figure}[htbp]
    \centering
    \includegraphics[width=\linewidth]{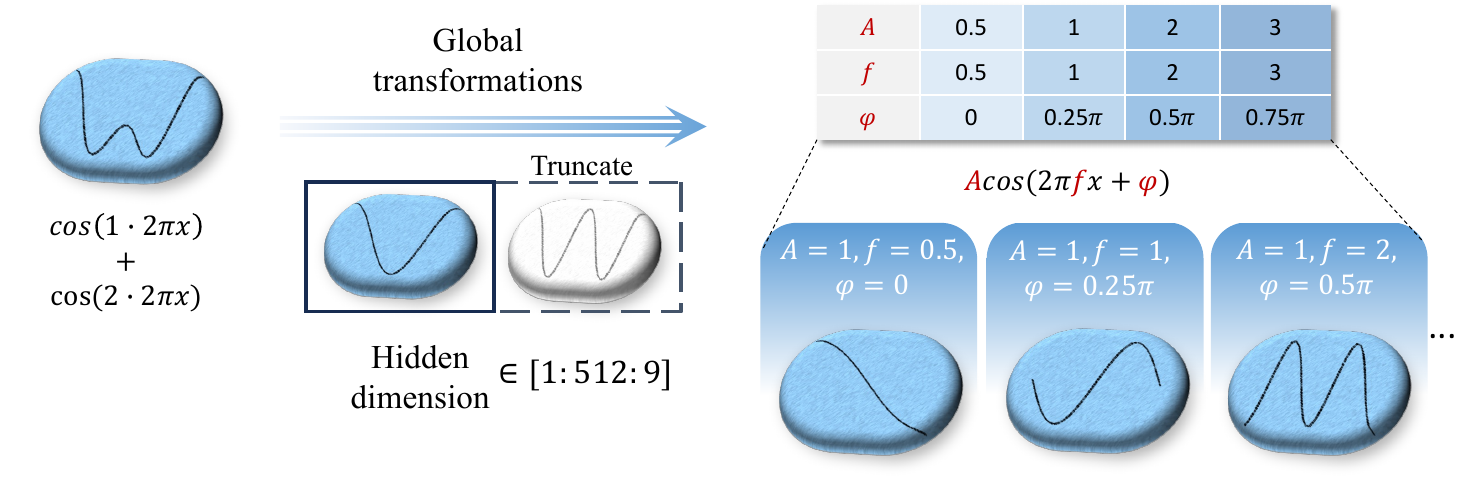}
    \caption{Comparison between the pure global transformation, the one with activation and the one with the local transformation, on the same output with 512 hidden dimensions. }
    \label{fig:limitation_experiment_settings}
\end{figure}

We designed a simple numerical experiment to investigate the performance sources in prevalent neural operators. Specifically, we compare the approximation capabilities of the global transformation versus the LGA transformation, both with and without activation functions.
For the experimental setup, we define a function mapping task between trigonometric functions. The input is fixed as a composite function $x(t) = \cos(1\cdot2\pi t) + \cos(2\cdot2\pi t)$, containing frequency components at 1 and 2. The target output is defined as $y(t) = A\cos(2\pi f t + \psi)$, where the parameters vary as follows: amplitude $A \in \{0.5, 1, 2, 3\}$, frequency $f \in \{0.5, 1, 2, 3\}$, and phase $\psi \in \{0, 0.25\pi, 0.5\pi, 0.75\pi\}$. These combinations correspond to varying degrees of difficulty, including lower-, equal-, and super-frequency mapping. 
Crucially, all spectral global transformations in this experiment are truncated at frequency $1$ ($K=1$). This implies that the input component $\cos(2\cdot2\pi t)$ is discarded, and the spectral layer preserves only the fundamental frequency $\cos(1\cdot2\pi t)$. We evaluate performance across latent dimensions ranging from $[1, 512]$ with a step size of 9, as shown in \ref{fig:limitation_experiment_settings}.

\begin{figure}
    \centering
    \includegraphics[width=\linewidth]{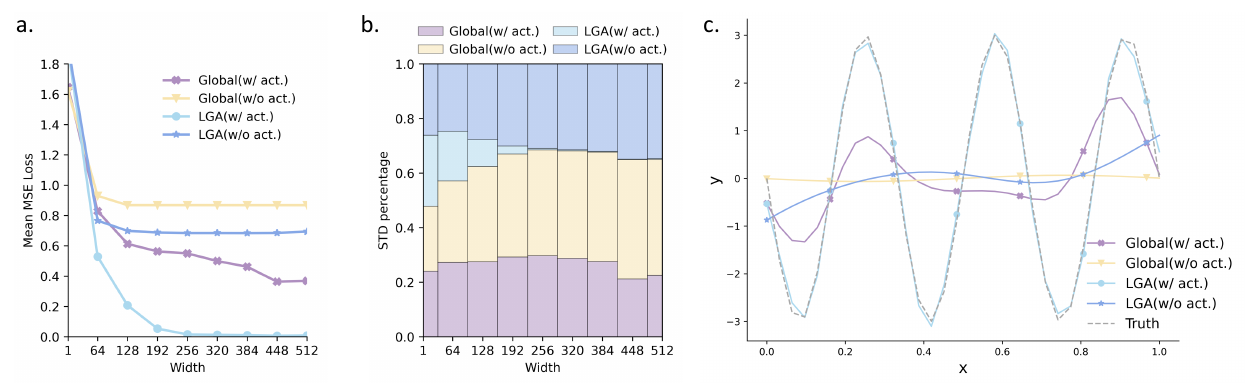}
    \caption{Comparison between the pure global transformation and the LGA transformation w/ (w/o) the activation and the local transformation. a. The mean of MSE losses across various tasks including super-, lower- and equal-mappings. b. The standard deviation of MSE losses. c. The super-mapping results from the different transformations in 512 hidden dimensions.}
    \label{fig:global_versus_lga}
\end{figure}

The means and standard deviations of the approximation error are depicted in Fig. \ref{fig:global_versus_lga}a and \ref{fig:global_versus_lga}b, respectively. Several key observations can be drawn: (1) Without activation or local transformation, the spectral global transformation exhibits the poorest performance. Due to the inherent truncation at $K=1$, it is structurally unable to represent target functions with frequencies $f=2$ or $f=3$, resulting in significant errors in super-frequency mapping tasks. (2) Introducing an activation function improves performance. This is attributed to the nonlinearity of activation functions (e.g., $\sigma(\cdot)$), which naturally generates high-frequency harmonics from low-frequency inputs, partially recovering the lost spectral information. However, the relatively high standard deviation suggests a lack of robustness across varying tasks. (3) The LGA transformation, particularly when combined with activation, yields the superior performance and notably lower standard deviations as the latent dimension increases.
Fig. \ref{fig:global_versus_lga}c visualizes the super-frequency mapping results ($f=3$) for a latent dimension of 512. The results suggest a synergistic effect: the activation function equips the network with the non-linearity required for frequency upscaling, while the local transformation (bypassing the spectral truncation) allows for a more precise preservation of input details. These numerical findings strongly support the theoretical limitation of pure spectral global transformations.

\subsection{Advantage of LGM transformations}
In this section, we provide a theoretical justification and a supporting numerical validation for the superior high-frequency reconstruction capabilities of the LGM transformation. 

\subsubsection{Theoretical Proof}
Here We explicitly show that the multiplicative interaction between a band-limited global spectral operator and a local operator extends the spectral bandwidth of the output, thereby overcoming the truncation limits derived in Corollary 1.

\begin{proposition}[Spectral Expansion Property]
Let $\mathcal{G}v$ be a global spectral transformation truncated at frequency bandwidth $K$, and let $\mathcal{L}v$ be a local transformation with effective bandwidth $M$, where typically $M \gg K$. The mixing transformation defined by $u(x) = (\mathcal{G}v)(x) \odot (\mathcal{L}v)(x)$ possesses a spectral support bounded by $K+M$. Consequently, $u(x)$ can represent frequency components strictly greater than the global truncation threshold $K$.
\end{proposition}

\begin{proof}
Let the domain $\Omega$ be the periodic torus $\mathbb{T}^d$ to facilitate Fourier analysis. We consider the output of the global path, $g(x) = (\mathcal{G}v)(x)$, and the output of the local path, $l(x) = (\mathcal{L}v)(x)$.

By the definition of the spectral truncation given in Eq. \ref{trunk_coeff}, the Fourier coefficients of the global component, denoted $\hat{g}(k)$, satisfy:
\begin{equation}
\text{supp}(\hat{g}) \subseteq {k \in \mathbb{Z}^d : |k| \leq K}.
\end{equation}
Conversely, the local transformation $l(x)$ operates on a localized spatial domain $P_\tau$. By the uncertainty principle of harmonic analysis, a function localized in the spatial domain exhibits broad support in the frequency domain. We denote the effective bandwidth of the local operator as $M$, such that:
\begin{equation}
\text{supp}(\hat{l}) \subseteq {q \in \mathbb{Z}^d : |q| \leq M}.
\end{equation}
where $M$ is typically bounded only by the Nyquist frequency of the discretization grid, implying $M \gg K$.

The LGM transformation is defined by the Hadamard product (element-wise mixing):
\begin{equation}
u(x) = g(x) \cdot l(x).
\end{equation}
According to the Convolution Theorem, multiplication in the spatial domain corresponds to convolution in the frequency domain. Thus, the spectral coefficients $\hat{u}(n)$ of the output are given by the discrete convolution of $\hat{g}$ and $\hat{l}$:
\begin{equation}
\hat{u}(n) = (\hat{g} * \hat{l})(n) = \sum_{k \in \mathbb{Z}^d} \hat{g}(k) \hat{l}(n-k).
\end{equation}
For $\hat{u}(n)$ to be non-zero, there must exist at least one $k$ such that both $\hat{g}(k)$ and $\hat{l}(n-k)$ are non-zero. This requires:
\begin{equation}
|k| \leq K \quad \text{and} \quad |n-k| \leq M.
\end{equation}
Using the triangle inequality $|n| = |n - k + k| \leq |n-k| + |k|$, we derive the bound for the support of $\hat{u}$:
\begin{equation}
|n| \leq M + K.
\end{equation}
This establishes that the spectral support of the LGM output is $\text{supp}(\hat{u}) \subseteq \{n \in \mathbb{Z}^d : |n| \leq K + M\}$.
\end{proof}

Unlike the pure global transformation, which strictly eliminates all frequencies $|n| > K$, the LGM transformation is capable of generating non-zero spectral responses for frequencies up to $K+M$. Since $M$ represents the fine-grained local resolution, the mixing interaction effectively "modulates" the global low-frequency carrier $g(x)$ with the high-frequency content of $l(x)$, thereby recovering high-frequency components that are structurally inaccessible to the global operator alone.

\subsubsection{Numerical Validation}
\begin{figure}[htbp]
    \centering
    \includegraphics[width=\linewidth]{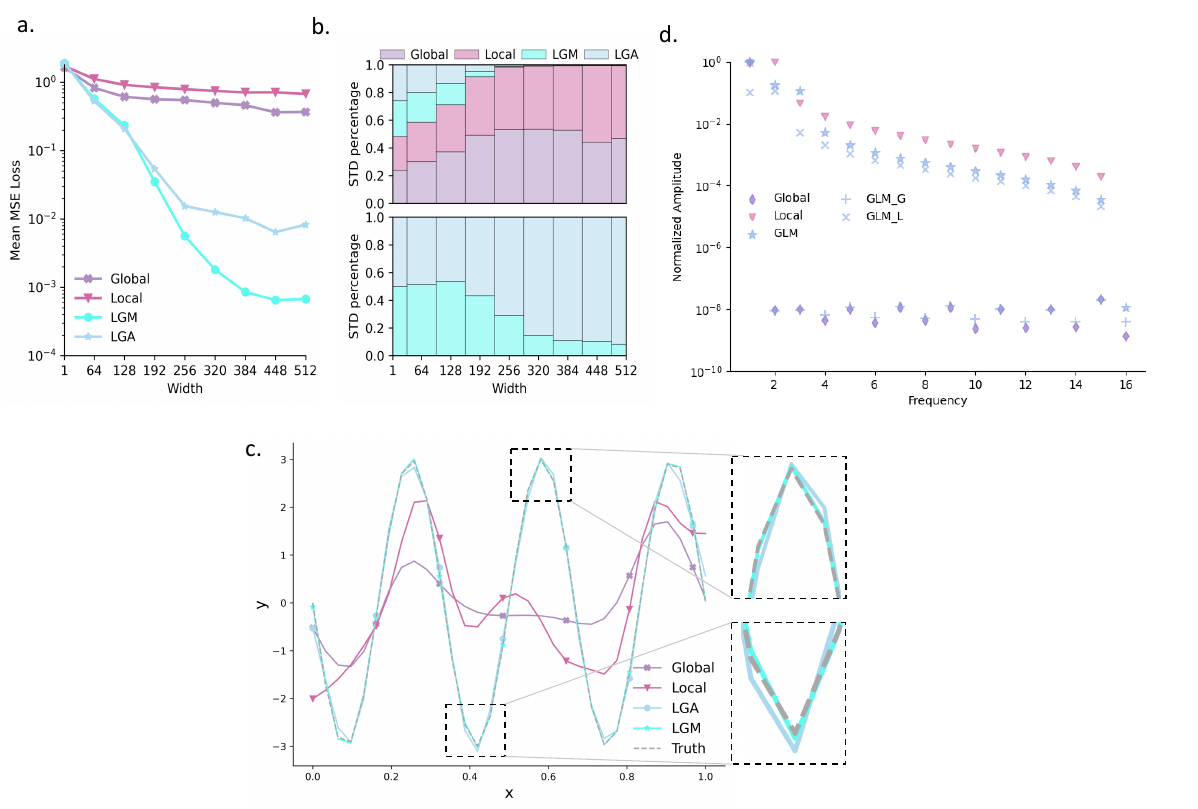}
    \caption{Performance and spectral analysis of local, global, LGA, and LGM transformations. (a–b) LGM achieves the lowest mean loss and smallest variance, especially at latent dimension 512. (c) In super-resolution tasks with sharp gradients, LGM outperforms LGA in capturing fine details. (d) Spectral decomposition shows LGM’s global component captures low frequencies (rapid decay), while its local component preserves high frequencies. Their mixing interaction yields a broad-spectrum representation that overcomes spectral bias. All variants use GELU activation.}
    \label{fig:lgm_versus_others}
\end{figure}

A thorough comparison is conducted between the local transformation, global transformation, LGA transformation, and \textbf{LGM transformation} to highlight the benefits of the LGM transformation. Importantly, each transformation is followed by the GELU activation function \cite{hendrycks2016gaussian}. The results are illustrated in Fig. \ref{fig:lgm_versus_others}, which shows that both the global and local transformations underperform and exhibit significant standard deviations across all tasks. In contrast, the LGA and LGM transformations show substantial performance improvements with increased latent dimensions. Notably, with a latent dimension of 512, the LGM transformation surpasses the LGA transformation by approximately an order of magnitude in mean loss and standard deviation, demonstrating superior performance and reliability. In Fig. \ref{fig:lgm_versus_others}c, an evaluation is conducted for super-frequency mappings, analogous to the previous comparison, to offer further insights into the advantages of the LGM transformation. The main disparities between the LGA transformation and the LGM transformation are shown in the enlarged views. Even though wave crests with sudden gradient shifts are challenging to model accurately, the LGM transformation still successfully captures this mapping. This satisfactory performance of the LGM transformation may be brought by its inherent multi-scales and nonlinearity features.
To elucidate the internal synergy of the LGM transformation, Fig. \ref{fig:lgm_versus_others}d presents a spectral decomposition of the trained model into its constituent global ($LGM\_G$) and local ($LGM\_L$) transformations. The amplitude spectrum reveals that the global component exhibits rapid spectral decay, efficiently capturing the low-frequency macro-structure, while the local component maintains high-amplitude responses across the spectrum, preserving fine-grained high-frequency details. Crucially, the final LGM output (marked by stars) demonstrates that the multiplicative mixing interaction effectively fuses these distinct behaviors. By modulating the global low-frequency base with local high-frequency details, the LGM architecture successfully reconstructs a broad-spectrum representation, overcoming the spectral bias inherent in pure global transformations.

\subsection{Datasets}\label{datasets}
To demonstrate the scalability and versatility of our method, we conduct experiments on the following datasets across multiple domains and PDE types. The input and output shapes in these datasets are conclude in Table \ref{tab:all_datasets}.

\begin{table}[htbp!]
    \centering 
    \caption{Summary of benchmark datasets and experimental configurations. The tensor shapes are denoted by $(C, T, S)$, where $C$ represents the channel dimension (vector size), $T$ is the temporal sequence length, and $S$ indicates the spatial resolution. The dataset split defines the sample count for training and testing ($N_{\text{train}} / N_{\text{test}}$).}
    \label{tab:all_datasets}
    \renewcommand{\arraystretch}{1.2} 
    \setlength{\tabcolsep}{6pt}       
    \small 
    
    \begin{tabular}{@{} lcccc @{}} 
        \toprule
        \textbf{Benchmark} & \textbf{Shape} & \textbf{Split} & \textbf{Operator} & \textbf{Mapping} \\
         & $(C, T, S)$ & ($N_{\text{train}} / N_{\text{test}}$) & \textbf{Type} & \textbf{Task} \\
        \midrule
        1D KS & $(1, 10, 256)$ & \multirow{5}{*}{$1000 / 200$} & \multirow{5}{*}{Evolutionary} & \multirow{5}{*}{Sol $\to$ Sol} \\
        2D Burgers & $(2, 10, 64^2)$ & & & \\
        2D CE-CRP & $(5, 10, 64^2)$ & & & \\
        2D Navier-Stokes & $(1, 10, 64^2)$ & & & \\
        3D Shallow Water & $(2, 10, 64 \times 32)$ & & & \\
        \midrule
        2D Darcy & $(1, 1, 49^2)$ & $1000 / 200$ & Solution & Param $\to$ Sol \\
        3D Brusselator & $(1, 1, 39 \times 28^2)$ & \phantom{0}$800 / 200$ & Response & Forcing $\to$ Sol \\
        \bottomrule
    \end{tabular}
\end{table}

\subsubsection{1D Kuramoto-Sivashinsky}
The Kuramoto-Sivashinsky equation is a nonlinear, fourth-order PDE, typically written as 
\begin{equation}
    u_t+u u_x+u_{x x}+ u_{x x x x}=0,
\end{equation}
where $u$ is a scalar field representing a physical quantity, such as the height of an interface or a velocity perturbation, evolving over space $x$ (a 1D coordinate) and time $t$.
 $u_t$ is the time derivative and $u_x$ the space derivative. The presence of the second- and fourth-order spatial derivatives, named the diffusion term and hyperdiffusion term, gives it a dissipative, diffusion-like character, classifying it as a parabolic PDE, despite its nonlinear and chaotic behavior. 

To generate the numerical solution as the dataset, the KS equation is solved by the pseudospectral method combined with the fourth-order exponential time-differencing  Runge–Kutta formula, generating an $20$-step temporal sequence $\{v_i\}_{i=1}^{20}$. The initial condition $v_0$ is sampled from $U(-1, 1)$. With $L = 64 \pi$, the time integration is implemented starting from $250s$ when chaos is fully developed until the final time $T= 121$ where $2048$ Fourier modes are employed to discretize the spatial domain. A total of $N=5000$ temporal sequences are generated. Consequently, this dataset is downsampled to the resolution 256 and involves predicting how the $u$ evolves in later 10 steps from given initial 10 steps.
 
\subsubsection{2D Darcy}
The 2d Darcy dataset models steady-state flow through porous media, such as groundwater movement or oil reservoir dynamics, using a linear elliptic PDE with spatially varying coefficients. The governing equation is:
\begin{equation}
    -\nabla \cdot(k(x, y) \nabla u)=f(x, y),
\end{equation}
where \( u(x, y) \) represents the pressure or potential field across a 2D spatial domain defined by coordinates \( x \) and \( y \), \( k(x, y) > 0 \) is the permeability or diffusion coefficient that varies with position, and \( f(x, y) \) is a source or sink term driving the flow (e.g., injection or extraction rates). The divergence \( \nabla \cdot \) measures the net flow of the vector field \( k \nabla u \). Physically, this equation balances the flux of \( u \) scaled by \( k \) with the source \( f \), describing a steady-state system without time dependence. It is classified as an elliptic PDE.

The 2D Darcy equation is solved by using a second-order finite difference scheme on a $421 \times 421$ grid. The diffusion coefficient $k$ are sampled from $\psi \mathcal{N}\left(0,(-\Delta+9 I)^{-2}\right)$ with zero Neumann boundary conditions on the Laplacian where the mapping $\psi$  takes the value $12$ on the positive part of the real line and 3 on the negative and the push-forward is defined pointwise. The forcing term $f$ is set to $1$. 
This dataset is downsampled to the resoultion $49$ and involves predicting \( u(x, y) \) given \( k(x, y) \).

\subsubsection{2D Burgers}
The 2D Burgers dataset extends the classic 1D Burgers’ equation to two dimensions, modeling viscous fluid flow with a nonlinear, time-dependent PDE. For a velocity vector \( \mathbf{u} = (u, v) \), the system is:
\begin{equation}
    \mathbf{u}_t + (\mathbf{u} \cdot \nabla) \mathbf{u} = \nu \Delta \mathbf{u},
\end{equation}
where \( \mathbf{u} = (u, v) \) is the velocity field in the \( x \) and \( y \) directions, \( \nu > 0 \) is the viscosity coefficient. The nonlinear term \( (\mathbf{u} \cdot \nabla) \mathbf{u} \) represents advection and \( \nu \Delta \mathbf{u} \) is the diffusion term that smooths out sharp gradients due to viscosity. This equation is classified as a parabolic PDE. 

To generate the numerical solution for the dataset, the 2D Burgers equation is solved numerically with a viscosity parameter \(\nu = 0.005\). The spatial discretization employs a pseudospectral method on a \(64 \times 64\) grid over the domain \([0, 1] \times [0, 1]\) with periodic boundary conditions. The spatial derivatives are computed using Fourier transforms, leveraging the efficiency of the pseudospectral approach. Time integration is performed using a fourth-order Runge-Kutta method with a time step of \(\Delta t = 0.0025\), advancing the solution from \(t = 0\) to \(t = 0.5\) over 200 time steps. The solution is recorded at intervals of 0.025 time units, resulting in a temporal sequence of 21 time points (from \(t = 0\) to \(t = 0.5\), inclusive of the initial condition).
The initial condition for each simulation is generated as a random field using a Fourier series with frequencies ranging from \(-4\) to \(4\) in each spatial direction. This field is normalized such that the maximum velocity magnitude across all batches and spatial points is scaled to approximately 1.5, and then shifted by a random constant vector with components uniformly distributed between \(-1\) and \(1\). Consequently, this dataset involves predicting how the $\mathbf{u}$ evolves in later 10 steps from given initial 10 steps.

\subsubsection{2D CE-CRP}
The 2D CE-CRP dataset extends the stochastic four-quadrant Riemann problem to include curved subdomains, modeling inviscid fluid flow with the compressible Euler equations:
\[ \frac{\partial}{\partial t} \begin{pmatrix} \rho \\ \rho \mathbf{u} \\ E \end{pmatrix} + \nabla \cdot \begin{pmatrix} \rho \mathbf{u} \\ \rho \mathbf{u} \otimes \mathbf{u} + p \mathbf{I} \\ (E + p) \mathbf{u} \end{pmatrix} = 0, \]
Therefore , it is classified as a hyperbolic PDEs. The domain is $[0,1]^2$ with periodic boundary conditions, partitioned into four curved subdomains using random sine functions. Each subdomain has constant initial conditions for density $\rho$, velocity $\mathbf{u} = (u, v)$, and pressure $p$, sampled from uniform distributions: $\rho \sim U[0.1,1]$, $u \sim U[-1,1]$, $v \sim U[-1,1]$, $p \sim U[0.1,1]$. Unlike viscous models, the Euler equations permit discontinuities like shocks. The dataset contains 10,000 trajectories capturing the time evolution of the flow field and was simulated on the unit square up to $T=1$. Details can refer to \cite{herde2024poseidon}. 

This dataset is saved in a $128\times128$ grid. In our work, it is downsampled to the resoultion $64 \times 64$ and involves predicting how the 5-dimensional solution vector (i.e., density, horizontal velocity, vertical velocity, pressure, energy) evolves in later 10 steps from given initial 10 steps.

\subsubsection{2D Navier-Stokes}
The 2D Navier-Stokes (NS) dataset models the dynamics of an incompressible, viscous fluid in two dimensions, governed by the Navier-Stokes equations. In this dataset, the equations are solved in the vorticity-stream function formulation, focusing on the evolution of the vorticity field \(\omega = \nabla \times \mathbf{u}\), which in 2D satisfies:
\begin{equation}
    \frac{\partial \omega}{\partial t} + (\mathbf{u} \cdot \nabla) \omega = \nu \Delta \omega.
\end{equation}
The velocity field is recovered from the stream function \(\psi\), where \(\Delta \psi = -\omega\) and \(\mathbf{u} = (\partial \psi / \partial y, -\partial \psi / \partial x)\). This equation is classified as a parabolic PDE.

To generate the numerical solution for the dataset, the 2D Navier-Stokes equation is solved using a pseudospectral method on a \(256 \times 256\) grid over the periodic domain \([0, 2\pi) \times [0, 2\pi)\). Spatial derivatives are computed efficiently via Fourier transforms, and dealiasing is applied using the 2/3 rule to mitigate aliasing errors. Time integration is performed with the Crank-Nicholson method, a second-order implicit-explicit scheme, using a time step of \(\Delta t = 0.001\). The simulation advances from \(t = 0\) to \(t = 30\), with solutions recorded every 1 time unit from \(t = 10\) to \(t = 30\), resulting in a temporal sequence of 21 time points per trajectory.

Each simulation in the dataset begins with a random initial vorticity field \(\omega_0\), where each grid point is independently sampled from a uniform distribution \(U(-1, 1)\). The viscosity is set to \(\nu = 10^{-5}\), promoting nearly inviscid behavior and the development of complex flow structures over time. The dataset comprises 1200 independent trajectories, each capturing the evolution of the vorticity \(\omega\) across the 21 time points. Consequently, this dataset involves predicting how the vorticity $\omega$ evolves in later 10 steps from given initial 10 steps.

\subsubsection{3D Shallow Water}
The 3D Shallow-water equations dataset models the dynamics of large-scale Rossby waves in the atmosphere, focusing on zonal flows in a viscous, rotating fluid layer over a spherical surface. For a fluid layer, the system is governed by the viscous Shallow-water equations:
\begin{align}
    \frac{\partial h}{\partial t} + \nabla \cdot (h \mathbf{V}) &= 0, \\
    \frac{\partial \mathbf{V}}{\partial t} + (\mathbf{V} \cdot \nabla) \mathbf{V} + f \mathbf{k} \times \mathbf{V} &= -g \nabla h + \nu \Delta \mathbf{V} - k \mathbf{V},
\end{align}
where \( h \) is the fluid layer thickness, \( \mathbf{V} = (u, v) \) is the velocity field in the eastward (\( u \)) and northward (\( v \)) directions, \( f = 2 \Omega \sin \phi \) is the Coriolis parameter, with \( \Omega \) as the Earth’s angular velocity, \( g \) is the acceleration due to gravity, \( \nu \) is the diffusion coefficient, \( k \) is the viscous drag coefficient, \( \mathbf{k} \) is the vertical unit vector. This system is classified as a hyperbolic PDE and is widely used to study atmospheric phenomena, such as Rossby waves, jet streams, and barotropic instabilities.

To generate the numerical solution for the dataset, the Shallow-water equations are solved using the Dedalus Project \href{https://doi.org/10.3402/tellusa.v56i5.14436}{https://doi.org/10.3402/tellusa.v56i5.14436}, a spectral method-based solver, for multiple cases with varying initial perturbation parameters, following the test case of a barotropically unstable mid-latitude jet as described in \cite{galewsky2004initial}. The simulation parameters are set as follows: Earth's angular velocity \( \Omega = 7.292 \times 10^{-5} \, \text{s}^{-1} \), gravitational acceleration \( g = 9.80616 \, \text{m/s}^2 \), hyperdiffusion coefficient \( \nu = 1.0 \times 10^5 \, \text{m}^2/\text{s} \) (matched at \(\ell = 32\)), maximum zonal velocity \( u_{\text{max}} = 80 \, \text{m/s} \), and jet boundaries \( \phi_0 = \frac{\pi}{7} \), \( \phi_1 = \frac{\pi}{2} - \phi_0 \), with the jet’s midpoint at latitude \( \frac{\pi}{4} \). The spatial domain is a spherical grid spanning longitudes \( \phi \in [0, 2\pi] \) and colatitudes \( \theta \in [0, \pi] \), discretized on a \( 256 \times 128 \) mesh, where 256 corresponds to the longitudinal direction and 128 to the latitudinal direction, reflecting the standard resolution for spherical coordinates where the colatitude range is half that of longitude. Time integration is performed from \( t = 120 \) to \( t = 360 \) hours with a time step of \( \Delta t = 600 \, \text{seconds} \) (equivalent to \( \frac{1}{6} \) hours), and solutions are recorded every 12 hours, resulting in 20 time points per simulation.

For each case, the initial conditions consist of a zonal jet velocity profile defined over latitudes between \( \phi_0 \) and \( \phi_1 \), and a balanced height field computed via a linear boundary value problem (LBVP) solver to ensure geostrophic balance with the imposed jet. A localized Gaussian perturbation is then added to the height field to induce barotropic instability, parameterized by shape parameters \( \alpha \) and \( \beta \), which control the longitudinal and latitudinal extent of the perturbation, respectively. The perturbation is expressed as:
\begin{equation}
    h'(\phi, \theta, t=0) = h_{\text{pert}} \cos(\text{lat}) \exp\left[ -\left( \frac{\phi}{\alpha} \right)^2 \right] \exp\left[ -\left( \frac{\text{lat} - \text{lat}_2}{\beta} \right)^2 \right],
\end{equation}
where latitude \( \text{lat} = \frac{\pi}{2} - \theta \), the perturbation center is at \( \text{lat}_2 = \frac{\pi}{4} \), the perturbation amplitude is \( h_{\text{pert}} = 120 \, \text{m} \), \( \phi \) is the longitude, and \( \theta \) is the colatitude. The parameters \( \alpha \) and \( \beta \) are systematically varied across a grid of values: \( \alpha \) ranges from \( \frac{1}{120} \) to 10 over 40 evenly spaced points, and \( \beta \) ranges from \( \frac{1}{300} \) to 2 over 30 evenly spaced points, yielding a total of \( 40 \times 30 = 1200 \) unique simulations. These ranges allow the perturbation to vary from highly localized (small \( \alpha \) and \( \beta \)) to broadly spread (large \( \alpha \) and \( \beta \)) in both longitudinal and latitudinal directions.
The dataset comprises 1200 simulations, and the spatial fields, originally on a \( 256 \times 128 \) grid, are downsampled to the resolution \( 64 \times 32 \). Each simulation records snapshots of the height field $h$ and vorticity $\omega$ every 12 hours, saving up to 30 snapshots per case. Consequently, this dataset involves predicting how the height and velocity fields evolve in later 10 steps from given initial 10 steps.

\subsubsection{3D Brusselator}
The 3D Brusselator dataset composed of 2D spatial dimension and 1D temporal dimension models autocatalytic chemical reactions, capturing the spatiotemporal evolution of reactant concentrations through a system of nonlinear reaction-diffusion equations. For concentration fields \( u \) and \( v \), the system is defined as:
\begin{align}
    \frac{\partial u}{\partial t} &= D_0 \Delta u + a + f(t) - (b + 1)u + u^2 v, \\
    \frac{\partial v}{\partial t} &= D_1 \Delta v + b u - u^2 v,
\end{align}
where \( u \) and \( v \) represent the concentrations of two chemical species at location \( \mathbf{x} = (x, y) \) and time \( t \), \( D_0 \) and \( D_1 \) are the diffusion coefficients for \( u \) and \( v \), respectively, \( a \) and \( b \) are constant parameters representing fixed concentrations, \( f(t) \) is a time-dependent signal introducing fluctuations to the system, and the nonlinear term \( u^2 v \) accounts for the autocatalytic reactions. This system is classified as a parabolic PDEs.

To generate the numerical solution for the dataset, the 3D Brusselator reaction-diffusion equations are solved numerically with the following parameters: \( a = 1 \), \( b = 3 \), \( D_0 = 1 \), and \( D_1 = 0.5 \). The spatial domain is a periodic square \([0, 1) \times [0, 1)\), discretized on a \( 28 \times 28 \) grid. The time evolution is computed using the finite difference method implemented in the py-pde solver \cite{zwicker2020py}, with a computational time step of \( \Delta t = 0.02 \) seconds, spanning the time range from \( t = 0 \) to \( t = 20 \). Snapshots of the solution are recorded at intervals of \( 0.5 \) seconds, resulting in a temporal sequence of 38 time points per trajectory. Each simulation produces a spatiotemporal dataset of size \( 28 \times 28 \times 38 \) for the concentration fields.

The initial conditions are defined as \( u(\mathbf{x}, 0) = 1 \) and \( v(\mathbf{x}, 0) = 1 + \epsilon(\mathbf{x}) \), where \( \epsilon(\mathbf{x}) \) is a randomly generated spatially varying field across the \( 28 \times 28 \) grid. A time-dependent signal \( f(t) \) introduces fluctuations and differs between the training and testing datasets. For the training set, the signal is \( f_{\text{train}}(t) = A_{\text{train}} e^{-0.01t} \sin(t) \), while for the testing set, it is \( f_{\text{test}}(t) = A_{\text{test}} e^{-0.05t} \sin(t) \). The amplitude \( A \) is randomly sampled from the interval \([0.01, 10]\), with 800 distinct values used for \( A_{\text{train}} \) in the training set and 200 values for \( A_{\text{test}} \) in the testing set. This dataset is designed to predict the evolved concentration field \( u(\mathbf{x}, t) \) in three dimensions (2D space + 1D time) given the signal \( f(t) \). To adapt to the network architecture, the one-dimensional signal \( f(t) \) is extended to a three-dimensional field \( f(\mathbf{x}, t) \), where the value of \( f \) remains constant across the spatial dimensions for each corresponding time point.

\subsection{Baselines}\label{baselines}
In the following models, they only take the previous temporal solutions and spatial coordinates as the generic input for a fair comparison. Additional prior informations applied in some models, such as POD-modulated inputs in DeepONet \cite{lu2022comprehensive} and edges \cite{hao2023gnot} in GNOT, are eliminated. All models predict temporal solutions for future multiple steps in an auto-regressive way unless specially specified. The applied convolutional transformations are kept with the kernel size of 1, unless specially specified, for retaining the mesh independency which is the important property of the neural operator. 

Based on the property of key blocks in baselines, we briefly classified them as different transformations in Table \ref{tab:grouped_baselines}, where some baselines distribute at different category in terms of different perspective.  

\begin{table}[htbp]
    \centering
    \caption{Baselines grouped by their Transformation Category. Baselines marked with an asterisk (*) appear in multiple categories due to their hybrid nature or specific structural properties (e.g., residual blocks combining global and local ops). \textbf{Global}: Global Transformation; \textbf{LLM}: Local-Local-Mixing Transformation; \textbf{LGA}: Local-Global-Adding Transformation.}
    \label{tab:grouped_baselines}
    \renewcommand{\arraystretch}{1.3}
    \begin{tabular}{l l p{8.5cm}}
        \toprule
        \textbf{Category} & \textbf{Baseline} & \textbf{Key Mechanism} \\
        \midrule
        \multirow{3}{*}{\textbf{Global}} & DeepONet & Dot product of globally integrated coefficients (Branch) and coordinate embeddings (Trunk). \\
         & GNOT* & Linear self-attention mechanism where queries aggregate information from the entire domain. \\
         & LaMO* & Latent State-Space Model (SSM) with multi-directional scanning for global receptive fields. \\
        \midrule
        \multirow{2}{*}{\textbf{LLM}} & ConvLSTM & Local $1 \times 1$ convolutions (pointwise Nemytskii operator) mixed via recurrent gating. \\
         & CoDA-NO* & Codomain Attention mechanism mixes distinct physical variables (tokens) at corresponding spatial coordinates. \\
        \midrule
        \multirow{6}{*}{\textbf{LGA}} & FNO & Additive composition of a global spectral convolution branch and a local linear branch. \\
         & LocalNO & Explicit sum of a global Fourier operator and local differential/integral operator branches. \\
         & U-NO & U-shaped architecture applying spectral (global) and linear (local) operators at varying scales. \\
         & CoDA-NO* & Utilizes FNO backbones for embeddings, integrating global spectral and local linear transformations via addition. \\
         & GNOT* & Integrates global Attention outputs with local features (Pointwise MLP) via residual addition. \\
         & LaMO* & Combines global SSM updates with local pointwise MLP channel mixing via sequential residual connections. \\
        \bottomrule
    \end{tabular}
\end{table}

\subsubsection{DeepONet}
We employ the DeepONet \cite{lu2019deeponet, lu2021learning}, a framework designed to approximate continuous operators by operating on the Cartesian product of the input domains, as a representative baseline. The architecture is grounded in the universal approximation theorem for operators, approximating the solution operator $G$ acting on an input function $u$ at a specific coordinate $y$. 
Mathematically, for an input function $u$ and a spatial coordinate $y$, the approximated solution component $G_\theta^{(j)}(u)(y)$ for the $j$-th physical quantity is formulated as:
$$G_\theta^{(j)}(u)(y) = \frac{1}{d_h^2} \sum_{k=1}^{d_h} b_k^{(j)}(u) \cdot t_k(y) + \beta^{(j)}$$
where $\mathbf{b}^{(j)}(u) \in \mathbb{R}^{d_h}$ is the output of the $j$-th Branch network, $\mathbf{t}(y) \in \mathbb{R}^{d_h}$ is the output of the shared Trunk network, $d_h$ is the hidden dimension (number of basis functions), and $\beta^{(j)}$ is a learnable scalar bias. Note that the dot product is explicitly normalized by $d_h^2$, a specific scaling factor adopted in this implementation to stabilize training dynamics.

The illustration of DeepONet architecture on a 2D case is shown in Table \ref{tab:deeponet_architecture}. The branch encoder processes high-dimensional spatiotemporal inputs. It consists of a stack of convolutional layers (each followed by BatchNorm and GELU), an adaptive average pooling layer, and a flattening operation. This is followed by a projection head composed of two linear layers with GELU activations, which compresses the flattened feature map into the final $d_h$ basis coefficients. Besides, the trunk encoder is a fully connected network processing spatial coordinates. It is composed of stacked linear layers with GELU activations. The width of the layers increases geometrically (doubling at each layer) to facilitate the extraction of high-level spatial features, culminating in an output dimension of $d_h$.

In this work, the DeepONet architecture provides flexibility through several adjustable hyperparameters, including the number of basis functions $d_h$ (width of the final hidden layer), the depth of the MLPs in both sub-networks, the kernel size $k$, the kernel stride $s$ and the target pooling dimension for the feature map. However, to align the model with the high-dimensional PDE benchmarks and computational constraints of this study, we impose specific structural configurations:
\begin{itemize}
\item[1.] Aggressive Downsampling Strategy ($k=2, s=5$): Within the Branch network's convolutional block, we adopt an aggressive downsampling scheme using a kernel size of 2 and a stride of 5. This constraint is critical to mitigate excessive GPU memory consumption and ensure sufficient feature compression of the input function.
\item[2.] Fixed Depth ($L=4$): We fix the number of layers in both the Branch and Trunk networks to 4. This depth provides a balance between expressivity and computational cost; shallow networks with smaller strides were found to yield feature maps with unnecessarily high resolutions, creating bottlenecks without proportional performance gains.
\end{itemize}
The applied experiment settings in all cases are listed in Table \ref{tab:deeponet-settings-wide}  and their corresponding model statistics are shown in Table \ref{tab:deeponet-model-stats}.

We categorize DeepONet as a global transformation. While the Branch network utilizes local convolutional layers, the architecture culminates in global pooling operations that compress the entire input domain $E_{\tau}$ into a unified latent vector. Thus, the final output relies on the interaction between coordinate embeddings and these globally integrated coefficients, fitting the definition of a global integral operator. However, it is acknowledged that if the convolutional component of the Branch network is viewed in isolation, it operates as a local transformation; yet, within the full DeepONet framework, this locality is subordinated to the global aggregation mechanism.

\begin{table}[htbp!]
    \centering
    \caption{DeepONet Architecture Breakdown on a 2D case. Key parameters: $d_{i}$: Input feature channels; $d_{out}$: Output feature channels; $d_{c}$: Hidden Dimension; $L$: Total layers (fixed at $L=4$); $R_f$: Resolution after Adaptive Average Pooling (res\_feat); $k$: Kernel size; $s$: Stride; $P$: Padding; $d_{s}$: Spatial coordinate dimension (fixed at $d_{s}=2$ for 2D inputs).}
    \label{tab:deeponet_architecture}
    \resizebox{\textwidth}{!}{%
    \begin{tabular}{llcc}
        \toprule
        \textbf{Stage} & \textbf{Operation} & \textbf{Branch Network (Input $\mathbf{u}$)} & \textbf{Trunk Network (Input $\mathbf{x}$)} \\
        \midrule
        \multirow{4}{*}{\textbf{I. Conv. Encoding}} & \multirow{2}{*}{Conv Layer 1} & $\text{Conv}(d_i \cdot (H+1) \to d_{c}/L \cdot 1, k, s, P)$ & \multirow{2}{*}{$\text{FCNL}(d_{s} \to d_{c}/(2^{L-1})), \text{GELU}$} \\
        & & $\text{BN}, \text{GELU}$ & \\
        \cmidrule{2-4}
        \multirow{2}{*}{($i=1$ to $L-1$)} & Conv Layer $i$ & $\text{Conv}(\text{In}_{i} \to \text{Out}_{i}, k, s, P)$ & $\text{FCNL}(\text{In}_{i} \to \text{Out}_{i}), \text{GELU}$ \\
        & & $\text{BN}, \text{GELU}$ & \\
        \cmidrule{3-4}
        \multicolumn{2}{l}{\textbf{Channel/Dim. Change $(\text{In} \to \text{Out})$}} & $d_{c} \cdot \frac{i}{L} \to d_{c} \cdot \frac{i+1}{L}$ & $d_{c} \cdot \frac{2^{i-1}}{2^{L-1}} \to d_{c} \cdot \frac{2^{i}}{2^{L-1}}$ \\
        \midrule
        \multicolumn{2}{l}{\textbf{II. Feature Compression}} & $\text{AdaptiveAvgPool2d}(R_{f})$ & --- \\
        \midrule
        \multirow{4}{*}{\textbf{III. MLP Projection}} & \multirow{2}{*}{Linear Layer 1} & $\text{Flatten} \to \text{FCNL}(d_{fs} \to d_{fs}/2), \text{GELU}$ & $\text{FCNL}(\dots \to d_{c}), \text{GELU}$ \\
        & & $(d_{fs} = R_{f}[0] \cdot R_{f}[1] \cdot d_{c})$ & \\
        \cmidrule{2-4}
        & Linear Layer 2 & $\text{FCNL}(d_{fs}/2 \to d_{c})$ & --- \\
        \midrule
        \multicolumn{2}{l}{\textbf{Final Output Dim.}} & $d_{c}$ & $d_{c}$ \\
        \midrule
        \multicolumn{2}{l}{\textbf{IV. Fusion}} & \multicolumn{2}{c}{$ (\mathbf{u}_{\text{branch}} \cdot \mathbf{x}_{\text{trunk}}) / d^2_c + b_{j}$} \\
        \bottomrule
    \end{tabular}
    }
\end{table}

\begin{table}[htbp!]
    \centering
    \caption{Experiment settings for the DeepONet baseline. Tiny (T), Medium (M), and Large (L) are defined by systematically scaling the hidden dimension. "Feat.~Dim.", "Depth" denote the target feature dimension after pooling and the depth in both the Branch and Trunk networks.}
    \label{tab:deeponet-settings-wide}
    \resizebox{\textwidth}{!}{
    \begin{tabular}{lccccccc}
        \toprule
        & \multicolumn{3}{c}{\textbf{Hidden Dimension $d_h$}} & & & \\
        \cmidrule(lr){2-4}
        \textbf{Dataset} & \textbf{T} & \textbf{M} & \textbf{L} & \textbf{Feat.~Dim.} & \textbf{Kernel} & \textbf{Stride} & \textbf{Depth} \\
        \midrule
        \multicolumn{8}{l}{\textbf{\textit{1D Experiments}}} \\
        Kuramoto-Sivashinsky (KS) & 128 & 224 & 448 & [64] & [5] & [2] & 4 \\
        \addlinespace
        \multicolumn{8}{l}{\textbf{\textit{2D Experiments}}} \\
        Burgers & 400 & 800 & 1600 & [16, 16] & [5, 5] & [2, 2] & 4 \\
        CE-CRP & 78 & 156 & 320 & [16, 16] & [5, 5] & [2, 2] & 4 \\
        Darcy & 78 & 156 & 320 & [16, 16] & [5, 5] & [2, 2] & 4 \\
        Navier--Stokes (NS) & 200 & 380 & 720 & [16, 16] & [5, 5] & [2, 2] & 4 \\
        Shallow Water (SW) & 200 & 380 & 720 & [10, 10] & [5, 5] & [2, 2] & 4 \\
        \addlinespace
        \multicolumn{8}{l}{\textbf{\textit{3D Experiments}}} \\
        Brusselator & 164 & 320 & 540 & [8, 6, 6] & [5, 5, 5] & [2, 2, 2] & 4\\
        \bottomrule
    \end{tabular}
    }
\end{table}

\begin{table}[H]
    \centering
    \caption{Model statistics of the DeepONet baseline. The parameter count (Param), peak GPU memory usage (Mem), and FLOPs for the Tiny, Medium, and Large variants are reported.}
    \label{tab:deeponet-model-stats}
    \resizebox{\textwidth}{!}{
    \begin{tabular}{l ccc ccc ccc}
        \toprule
        & \multicolumn{3}{c}{\textbf{Tiny (T)}} & \multicolumn{3}{c}{\textbf{Medium (M)}} & \multicolumn{3}{c}{\textbf{Large (L)}} \\
        \cmidrule(lr){2-4} \cmidrule(lr){5-7} \cmidrule(lr){8-10}
        \textbf{Benchmark} & \textbf{Param} & \textbf{Mem} & \textbf{FLOPs} & \textbf{Param} & \textbf{Mem} & \textbf{FLOPs} & \textbf{Param} & \textbf{Mem} & \textbf{FLOPs} \\
        & (M) & (MiB) & (M) & (M) & (MiB) & (M) & (M) & (MiB) & (M) \\
        \midrule
        1D KS          & 9  & 456   & 142    & 27  & 1,176  & 434    & 107   & 4,342  & 1,727 \\
        2D Burgers     & 54 & 4,566 & 15,524 & 215 & 12,636 & 60,001 & 858   & 39,836 & 235,816 \\
        2D CE-CRP      & 64 & 6,498 & 3,178  & 257 & 14,068 & 10,385 & 1,082 & 48,246 & 38,282 \\
        2D Darcy       & 7  & 342   & 25     & 30  & 1,254  & 100    & 126   & 4,880  & 418 \\
        2D NS          & 84 & 4,170 & 2,721  & 305 & 13,730 & 9,592  & 1,095 & 47,134 & 34,012 \\
        3D Brusselator & 91 & 3,842 & 986    & 348 & 13,816 & 3,714  & 991   & 38,458 & 10,523 \\
        3D SW          & 45 & 2,622 & 2,439  & 162 & 7,446  & 8,357  & 580   & 22,942 & 29,164 \\
        \bottomrule
    \end{tabular}
    }
\end{table}

\subsubsection{ConvLSTM}
We adopt the Convolutional LSTM (ConvLSTM), originally proposed by \cite{shi2015convolutional}, as a representative baseline for recurrent-convolutional architectures. The ConvLSTM extends the classical LSTM \cite{yu2019review} by replacing fully connected transitions with convolutional operations, thereby preserving the spatial topology of dynamical systems. At each time step $t$, the ConvLSTM cell processes the current spatial input $\mathcal{X}_t$ and the previous hidden state $\mathcal{H}_{t-1}$. The internal dynamics are governed by the input ($i_t$), forget ($f_t$), output ($o_t$) gates, and the cell input ($g_t$), formulated as follows:
\begin{align}
i_t &= \sigma(W_{xi} * \mathcal{X}_t + W_{hi} * \mathcal{H}_{t-1} + b_i) \\
f_t &= \sigma(W_{xf} * \mathcal{X}_t + W_{hf} * \mathcal{H}_{t-1} + b_f) \\
o_t &= \sigma(W_{xo} * \mathcal{X}_t + W_{ho} * \mathcal{H}_{t-1} + b_o) \\
g_t &= \tanh(W_{xc} * \mathcal{X}_t + W_{hc} * \mathcal{H}_{t-1} + b_c)
\end{align}
where $*$ denotes the convolution operator, $\sigma$ is the sigmoid activation, and $W$ represents the learnable convolutional kernels. The cell state $\mathcal{C}_t$ and the updated hidden state $\mathcal{H}_t$ are computed via element-wise interactions:
$$
\mathcal{C}_t = f_t \odot \mathcal{C}_{t-1} + i_t \odot g_t
$$
$$
\mathcal{H}_t = o_t \odot \tanh(\mathcal{C}_t)
$$
where $\odot$ denotes the Hadamard product.

In this work, the ConvLSTM architecture allows for adjustable hyperparameters, including the hidden dimension $d_h$, the kernel size $k$, and the number of stacked layers $L$. However, to align the model with the requirements of operator learning and computational feasibility in this study, we impose two critical constraints:
\begin{itemize}
    \item[1.] Mesh-Invariance Constraint ($k=1$): Unlike standard convolutional networks that utilize spatial neighborhoods (e.g., $3 \times 3$), we fix the kernel size to $1 \times 1$. A $1 \times 1$ convolution functions as a pointwise Nemytskii operator, processing features at each coordinate independently of the grid resolution. This constraint is essential to endow the ConvLSTM with the mesh-invariance property, formally characterizing it as a neural operator rather than a resolution-dependent solver.
    \item[2.] Computational Efficiency Constraint ($L=2$): While deeper recurrence can theoretically capture longer-term dependencies, it incurs significant memory and time overheads during backpropagation through time. Consequently, we fix the number of ConvLSTM layers to 2, balancing model expressivity with an affordable computational budget.
\end{itemize}
The applied experiment settings in all cases are listed in Table \ref{tab:convlstm-settings-wide}  and their corresponding model statistics are shown in Table \ref{tab:convlstm-model-stats}.

We classify the ConvLSTM as a LLM transformation.
The $1 \times 1$ convolution acts on a partial domain (specifically, a single spatial point), qualifying it as a local transformation. The subsequent gating mechanisms utilize the Hadamard product to combine these locally transformed historical states ($\mathcal{C}_{t-1}$) and current inputs ($g_t$). Because the architecture relies on the element-wise mixing of local transformations without an explicit global integration step (such as a spectral transform), it strictly adheres to the definition of the Local-Local-Mixing transformation.

\begin{table}[htbp!]
    \centering
    \caption{Experiment settings for the ConvLSTM baseline. Tiny (T), Medium (M), and Large (L) are defined by systematically scaling the hidden dimension. }
    \label{tab:convlstm-settings-wide}
    \begin{tabular}{lccccc}
        \toprule
        & \multicolumn{3}{c}{\textbf{Hidden Dimension}} \\
        \cmidrule(lr){2-4}
        \textbf{Dataset} & \textbf{T} & \textbf{M} & \textbf{L} & \textbf{Kernel Size} & \textbf{Depth} \\
        \midrule
        \textbf{\textit{1D Experiments}} & 32 & 128 & 512 & [1] & 2 \\
        \textbf{\textit{2D Experiments}} & 32 & 128 & 512 & [1, 1] & 2 \\
        \textbf{\textit{3D Experiments}} & 32 & 128 & 512 & [1, 1, 1] & 2 \\
        \bottomrule
    \end{tabular}
\end{table}

\begin{table}[htbp!]
    \centering
    \caption{Model statistics of the ConvLSTM baseline. The parameter count (Param), peak GPU memory usage (Mem), and FLOPs for the Tiny, Medium, and Large variants are reported.}
    \label{tab:convlstm-model-stats}
    \resizebox{\textwidth}{!}{
    \begin{tabular}{l *{3}{ccc}}
        \toprule
        & \multicolumn{3}{c}{\textbf{Tiny (T)}} & \multicolumn{3}{c}{\textbf{Medium (M)}} & \multicolumn{3}{c}{\textbf{Large (L)}} \\
        \cmidrule(lr){2-4} \cmidrule(lr){5-7} \cmidrule(lr){8-10}
        \textbf{Benchmark} & \textbf{Param} & \textbf{Mem} & \textbf{FLOPs} & \textbf{Param} & \textbf{Mem} & \textbf{FLOPs} & \textbf{Param} & \textbf{Mem} & \textbf{FLOPs} \\
        & (K) & (MiB) & (M) & (K) & (MiB) & (M) & (M) & (MiB) & (M) \\
        \midrule
        1D KS           & 13 & 198 & 32 & 198 & 924 & 505 & 3 & 3,096 & 8,060 \\
        2D Burgers      & 13 & 3,070 & 516 & 199 & 11,568 & 8,105 & 3 & 45,618 & 129,059 \\
        2D CE-CRP       & 13 & 4,500 & 520 & 199 & 11,836 & 8,121 & 3 & 46,168 & 129,122 \\
        2D Darcy        & 13 & 244 & 30 & 199 & 868 & 475 & 3 & 3522 & 7,564 \\
        2D NS           & 13 & 2,952 & 515 & 199 & 11,500 & 8,100 & 3 & 45,098 & 129,038 \\
        3D Brusselator  & 13 & 2,816 & 388 & 199 & 10,830 & 6,062 & 3 & 42,368 & 96,387 \\
        3D SW           & 13 & 1,696 & 258 & 199 & 5,788 & 4,053 & 3 & 22,828 & 64,529 \\
        \bottomrule
    \end{tabular}
    }
\end{table}

\subsubsection{FNO}
We adopt the Fourier Neural Operator (FNO) \cite{li2020fourier}, a resolution-invariant architecture that parameterizes the integral kernel in Fourier space, as a classical baseline. The FNO approximates the solution operator by transforming the input function $u(x)$ into a high-dimensional latent representation, evolving it through a sequence of spectral-convolutional layers, and projecting it back to the target domain. Mathematically, the operator is defined as a composition of a lifting mapping $\mathcal{P}$, a sequence of $L$ iterative Fourier layers, and a projection mapping $\mathcal{Q}$:
$$
G_\theta(u) = \mathcal{Q} \circ \mathcal{L}_L \circ \dots \circ \mathcal{L}_1 \circ \mathcal{P}(u)
$$
The lifting layer $\mathcal{P}$ acts as a pointwise MLP that maps the input $u(x)$ and coordinate information to a higher-dimensional latent field $v_0(x) \in \mathbb{R}^{d_h}$.
Inside each Fourier layer $\mathcal{L}_l$, the update rule combines a global spectral transformation with a local linear transformation. In this specific implementation, the update is formulated as:
$$
v_{l+1}(x) = \text{Norm}\left( \sigma \left( \mathcal{M}_l(\mathcal{K}_l(v_l)(x)) + \mathcal{W}_l(v_l)(x) \right) \right)
$$
where $\mathcal{W}_l$ is the Local Transformation, implemented as a $1 \times 1$ convolution (pointwise linear pass-through), capturing local dependencies and channel mixing. $\mathcal{K}_l$ is the Spectral Global Transformation, defined by the multiplication of frequency modes in the Fourier domain:
$$
\mathcal{K}_l(v)(x) = \mathcal{F}^{-1} \left( R_l \cdot \mathcal{F}(v)(k) \right)(x)
$$
Here, $\mathcal{F}$ and $\mathcal{F}^{-1}$ denote the Fast Fourier Transform and its inverse, respectively. $R_l$ is a learnable complex-valued tensor that weights the truncated frequency modes $k \le k_{max}$. $\mathcal{M}_l$ is a supplementary MLP applied specifically to the output of the spectral branch to refine the global features before aggregation. $\sigma$ is the GELU activation function, and $\text{Norm}$ denotes Batch Normalization.

In this work, the FNO architecture allows for adjustable hyperparameters, including the hidden dimension $d_h$, global modes $k_{max}$, depth $L$. The applied experiment settings in all cases are listed in Table \ref{tab:fno-settings-wide}  and their corresponding model statistics are shown in Table \ref{tab:fno-model-stats}.

We classify the FNO as a LGA transformation.
The architecture explicitly decomposes the dynamics into two parallel branches: a global branch governed by the spectral convolution $\mathcal{K}_l$ (which mixes information across the entire domain $E_{\tau}$ via Fourier bases) and a local branch governed by the pointwise convolution $\mathcal{W}_l$ (which operates on the partial domain $P_{\tau}$ consisting of a single point). The outputs of these two branches are combined via element-wise addition, fitting the precise definition of the LGA transformation.

\begin{table}[htbp!]
    \centering
    \caption{Experiment settings for the FNO baseline. Tiny (T), Medium (M), and Large (L) are defined by systematically scaling the hidden dimension and depth. "Modes" denote the truncated global modes in the spectrum.}
    \label{tab:fno-settings-wide}
    \begin{tabular}{lccccccc}
        \toprule
        & \multicolumn{3}{c}{\textbf{Hidden Dimension}} & \multicolumn{3}{c}{\textbf{Depth}} \\
        \cmidrule(lr){2-4} \cmidrule(lr){5-7}
        \textbf{Dataset} & \textbf{T} & \textbf{M} & \textbf{L} & \textbf{T} & \textbf{M} & \textbf{L} & \textbf{Modes} \\
        \midrule
        \multicolumn{5}{l}{\textbf{\textit{1D Experiments}}} \\
        Kuramoto-Sivashinsky (KS) & 32 & 64 & 128 & 2 & 4 & 8 & [12] \\
        \addlinespace
        \multicolumn{5}{l}{\textbf{\textit{2D Experiments}}} \\
        Burgers & 32 & 64 & 128 & 2 & 4 & 8 & [12, 12] \\
        CE-CRP & 32 & 64 & 128 & 2 & 4 & 8 & [12, 12] \\
        Darcy & 32 & 64 & 128 & 2 & 4 & 8 & [12, 12] \\
        Navier--Stokes (NS) & 32 & 64 & 128 & 2 & 4 & 8 & [12, 12] \\
        Shallow Water (SW) & 32 & 64 & 128 & 2 & 4 & 8 & [10, 10] \\
        \addlinespace
        \multicolumn{5}{l}{\textbf{\textit{3D Experiments}}} \\
        Brusselator & 32 & 64 & 128 & 2 & 4 & 8 & [6, 6, 6]\\
        \bottomrule
    \end{tabular}
\end{table}


\begin{table}[htbp!]
    \centering
    \caption{Model statistics of the FNO baseline. The parameter count (Param), peak GPU memory usage (Mem), and FLOPs for the Tiny, Medium, and Large variants are reported.}
    \label{tab:FNO-model-stats}
    \resizebox{\textwidth}{!}{
    \begin{tabular}{l *{3}{ccc}}
        \toprule
        & \multicolumn{3}{c}{T} & \multicolumn{3}{c}{M} & \multicolumn{3}{c}{L} \\
        \cmidrule(lr){2-4} \cmidrule(lr){5-7} \cmidrule(lr){8-10}
                \textbf{Benchmark} & \textbf{Param} & \textbf{Mem} & \textbf{FLOPs} & \textbf{Param} & \textbf{Mem} & \textbf{FLOPs} & \textbf{Param} & \textbf{Mem} & \textbf{FLOPs} \\
        & (K) & (MiB) & (M) & (K) & (MiB) & (M) & (K) & (MiB) & (M) \\
        \midrule
        1D KS           & 9 & 246 & 23 & 51 & 814 & 130 & 333 & 2,916 & 847 \\
        2D Burgers      & 10 & 3,930 & 398 & 53 & 9,938 & 2,139 & 336 & 32,652 & 13,673 \\
        2D CE-CRP       & 12 & 5,092 & 485 & 57 & 11,372 & 2,312 & 344 & 33,792 & 14,019 \\
        2D Darcy        & 9 & 282 & 20 & 50 & 778 & 119 & 331 & 2,858 & 789 \\
        2D NS           & 9 & 3,238 & 370 & 52 & 9,598 & 2,081 & 333 & 32,458 & 13,558 \\
        3D Brusselator  & 9 & 3,886 & 260 & 50 & 11,460 & 1,522 & 331 & 38,612 & 10,058 \\
        3D SW           & 10 & 2,020 & 199 & 53 & 5,096 & 1,070 & 336 & 16,636 & 6,837 \\
        \bottomrule
    \end{tabular}
    }
\end{table}

\subsubsection{GNOT}
We adopt the Generalized Neural Operator Transformer (GNOT) \cite{hao2023gnot}, a transformer-based architecture designed to learn resolution-invariant operators, as a representative baseline. The GNOT employs a heterogeneous embedding strategy and a sequence of attention-based interaction blocks to approximate the solution operator. Unlike standard autoregressive approaches that iterate step-by-step, this implementation utilizes a sequence-to-sequence (Seq2Seq) non-autoregressive strategy due to a high requirement of GPU memory in the attention operation. The model takes the current state and spatial coordinates as input and simultaneously predicts the full trajectory of future temporal solutions. This design choice significantly enhances memory efficiency for multi-step predictions, allowing the model to allocate more parameters to representation learning rather than storing recurrent computation graphs.

The architecture proceeds in three stages. Firstly, the Trunk network encodes the concatenated input features $u_{in}$ (comprising the current solution field and spatial coordinates) into a latent embedding $z_0 \in \mathbb{R}^{N \times d_h}$. Multiple Branch networks independently embed distinct input components $c^{(i)}$ (e.g., spatial coordinates alone, mesh parameters) into context embeddings $e^{(i)} \in \mathbb{R}^{M \times d_h}$. In the following, the latent embedding $z_l$ is refined through $L$ stacked blocks, each consisting of a linear cross-attention block that fuses global context from the branch embeddings $e^{(i)}$ into the trunk embedding $z_l$, a linear self-attention block that captures long-range spatial dependencies within the trunk embedding itself and a pointwise MLP that enhances feature representation. Noteworthy, the attention mechanism employed is linear attention, which reduces the quadratic complexity $O(N^2)$ to linear complexity $O(N)$ by avoiding the explicit materialization of the attention matrix. The update rule for a query $Q$, key $K$, and value $V$ is given by:
\begin{equation}
    \text{LinearAttn}(Q, K, V) = \frac{\sum_{j} \phi(Q_i) \phi(K_j)^T V_j}{\sum_{j} \phi(Q_i) \phi(K_j)^T}
\end{equation}
where $\phi(\cdot) = \text{softmax}(\cdot)$ serves as the kernel feature map.
Finally, the final latent representation $z_L$ is projected by an output MLP to the target domain dimension $d_{out} \times T_{pred}$, generating the entire future sequence at once.

In this work, the GNOT architecture allows for several adjustable hyperparameters, including the hidden dimension $d_h$, the number of interaction blocks $L$, the number of attention heads $H$, the inner MLP expansion factor $n_{inner}$, and the depth of the MLPs within the sub-layers.
We impose the following fixed parameters:
\begin{itemize}
    \item[1.] Attention Heads ($H=2$): We fix the number of attention heads to 2.
    \item[2.] MLP Depth ($L_{mlp}=2$): The Trunk, Branch, and output MLPs are all fixed to a depth of 2 layers.
    \item[3.] Inner Expansion Factor ($n_{inner}=4$): The expansion ratio for the inner dimension of the FFN within the attention blocks is fixed to 4.
\end{itemize}
These values are the default settings of the original GNOT implementation \cite{hao2023gnot}. Adhering to these defaults ensures that the baseline represents the canonical performance of the architecture. The experiment settings for different model sizes are listed in Table \ref{tab:gnot-settings-wide} and their corresponding model statistics are shown in Table \ref{tab:gnot-model-stats}.

We classify the GNOT as a global/LGA transformation.
The core mechanism of the GNOT is the linear self-attention layer. In this mechanism, the update of a feature at any specific spatial location $x_i$ (Query) is computed by aggregating information from all other locations $x_j$ (Keys and Values) across the entire domain $E_{\tau}$. This dense, all-to-all communication enables the model to capture global dependencies and approximate global integral operators directly, fitting the definition of a Global Transformation. Simultaneously, structurally, the GNOT block integrates this global attention output with local features processed by a pointwise MLP (FeedForward Network) via residual connections. This formulation, $x + \text{Attention}(x) + \text{MLP}(x)$, aligns with the LGA transformation, where the outputs of global and local transformations are composed additively.

\begin{table}[htbp!]
    \centering
    \caption{Experiment settings for the GNOT baseline. Tiny (T), Medium (M), and Large (L) are defined by systematically scaling the hidden dimension and depth. The attention block expansion ratio is fixed to 4 and the attention heads are fixed to 2 across all experiments."MLP Layers" specifies the depth of the Trunk, Branch, and output MLPs.}
    \label{tab:gnot-settings-wide}
    \resizebox{\textwidth}{!}{
    \begin{tabular}{lccccccccc}
        \toprule
        & \multicolumn{3}{c}{\textbf{Hidden Dimension}} & \multicolumn{3}{c}{\textbf{Depth}} & \multicolumn{2}{c}{\textbf{Attention Block}} \\
        \cmidrule(lr){2-4} \cmidrule(lr){5-7} \cmidrule(lr){8-9}
        \textbf{Dataset} & \textbf{T} & \textbf{M} & \textbf{L} & \textbf{T} & \textbf{M} & \textbf{L} & \textbf{Heads} & \textbf{Ratio} & \textbf{MLP Layers} \\
        \midrule
        \multicolumn{7}{l}{\textbf{\textit{1D Experiments}}} \\
        Kuramoto-Sivashinsky (KS) & 64 & 128 & 256 & 2 & 4 & 8 & 2 & 4 & 2 \\
        \multicolumn{7}{l}{\textbf{\textit{2D Experiments}}} \\
        Burgers & 64 & 128 & 256 & 2 & 4 & 8 & 2 & 4 & 2 \\
        CE-CRP & 64 & 128 & 256 & 2 & 4 & 8 & 2 & 4 & 2 \\
        Darcy & 16 & 36 & 84 & 1 & 2 & 4 & 2 & 4 & 2 \\
        Navier--Stokes (NS) & 64 & 128 & 256 & 2 & 4 & 8 & 2 & 4 & 2 \\
        Shallow Water (SW) & 64 & 128 & 256 & 2 & 4 & 8 & 2 & 4 & 2 \\
        \multicolumn{7}{l}{\textbf{\textit{3D Experiments}}} \\
        Brusselator & 16 & 32 & 64 & 1 & 2 & 4 & 2 & 4 & 2 \\
        \bottomrule
    \end{tabular}
    }
\end{table}

\begin{table}[H]
    \centering
    \caption{Model statistics of the GNOT baseline. The parameter count (Param), peak GPU memory usage (Mem), and FLOPs for the Tiny, Medium, and Large variants are reported.}
    \label{tab:gnot-model-stats}
    \resizebox{\textwidth}{!}{
    \begin{tabular}{l *{3}{ccc}}
        \toprule
        & \multicolumn{3}{c}{\textbf{Tiny (T)}} & \multicolumn{3}{c}{\textbf{Medium (M)}} & \multicolumn{3}{c}{\textbf{Large (L)}} \\
        \cmidrule(lr){2-4} \cmidrule(lr){5-7} \cmidrule(lr){8-10}
        \textbf{Benchmark} & \textbf{Param} & \textbf{Mem} & \textbf{FLOPs} & \textbf{Param} & \textbf{Mem} & \textbf{FLOPs} & \textbf{Param} & \textbf{Mem} & \textbf{FLOPs} \\
        & (K) & (MiB) & (M) & (K) & (MiB) & (M) & (K) & (MiB) & (G) \\
        \midrule
        1D KS           & 269 & 304 & 68 & 1,923 & 994 & 491 & 14,496 & 3,832 & 4 \\
        2D Burgers      & 271 & 3,882 & 1,104 & 1,927 & 12,848 & 7,870 & 14,504 & 44,886 & 59 \\
        2D CE-CRP       & 277 & 5,398 & 1,127 & 1,939 & 13,708 & 7,917 & 14,527 & 45,622 & 59 \\
        2D Darcy        & 11 & 396 & 25 & 86 & 1,402 & 204 & 831 & 4,924 & 2 \\
        2D NS           & 270 & 3,646 & 1,096 & 1,923 & 12,836 & 7,854 & 14,497 & 44,876 & 59 \\
        3D Brusselator  & 11 & 4,544 & 315 & 68 & 13,552 & 2,063 & 485 & 45,698 & 14,739 \\
        3D SW           & 271 & 1,968 & 552 & 1,927 & 6,238 & 3,935 & 14,504 & 23,794 & 30 \\
        \bottomrule
    \end{tabular}
    }
\end{table}


\subsubsection{LaMO}
We adopt the Latent Mamba Operator (LaMO) \cite{tiwari2025latent}, a recently proposed neural operator that leverages Structured State-Space Models (SSMs) to efficiently model long-range dependencies in PDE solutions, as one of baselines. LaMO mitigates the quadratic computational complexity associated with Transformer-based operators by utilizing the linear complexity of the Mamba architecture, while operating within a compact latent space.

The architecture processes the input physical field $u(x)$ through the following stages: Firstly, the input domain is partitioned into non-overlapping local patches. These patches are linearly projected to form a sequence of latent tokens $Z^0 \in \mathbb{R}^{M \times d_h}$, where $M$ is the number of patches and $d_h$ is the embedding dimension. This operation, defined as $\mathcal{E}$, effectively reduces the sequence length while preserving local structures and acts as an encoder for latent patches:
$$Z^0 = \mathcal{E}(u(x)) = \text{Patchify}(u),$$
but its mesh-dependence violates the mesh-invariance property.
After latent patchifying, the core processing consists of $L$ stacked blocks. Each block integrates a Latent-SSM layer for global information propagation and an MLP for channel mixing. The update rule for the $l$-th layer is given by:
$$
\begin{aligned}
\hat{Z}^l &= \text{Norm}(Z^{l-1}) \\
Z^{l'} &= Z^{l-1} + \text{SSM}(\hat{Z}^l) \\
Z^l &= Z^{l'} + \text{MLP}(\text{Norm}(Z^{l'}))
\end{aligned}
$$
The SSM layer employs a multi-directional scanning mechanism (e.g., cross-scan) to traverse the 2D grid of latent tokens. The underlying continuous dynamics are governed by 
$$h'(t) = A h(t) + B x(t), \quad y(t) = C h(t)$$
where $h'(t)$ represents the instantaneous rate of change of the latent state, determined by the system's internal dynamics matrix $A$ and the influence of the external input $x(t)$ via matrix $B$. This continuous ODE is discretized to approximate a global kernel integral operator $\mathcal{K}$ in the latent space, allowing the model to efficiently capture long-range dependencies.
Finally, the processed latent tokens $Z^L$ are projected back to the spatial domain and reshaped to reconstruct the solution field at the original resolution.

In this work, the LaMO architecture allows for adjustable hyperparameters, including the hidden dimension $d_h$, the number of blocks $L$, the number of heads, MLP expansion ratio and the patch size. To ensure a convenient and relatively fair comparison within our experimental framework, we impose the following constraints:

\begin{itemize}
    \item[1.] Fixed MLP Expansion Ratio ($r=4$): The expansion ratio for the Multi-Layer Perceptron (MLP) within the Mamba blocks is fixed to 4. This adheres to the default configuration of the original LaMO architecture.
\end{itemize}
It is noteworthy that, due to the inherent design of the patchification and 2D cross-scanning modules, the LaMO baseline is restricted to 2D spatial problems. For 1D problems, we adapt the input by expanding it into a pseudo-2D format with dummy padding to match the patch size requirements. This baseline is not applied to 3D volumetric cases due to the limitations of the current scanning implementation. The experiment settings for different model sizes are listed in Table \ref{tab:lamo-settings-wide} and their corresponding model statistics are shown in Table \ref{tab:LaMO-model-stats}. 
To avoid the bias of task-specific hyperparameter engineering, we evaluate the LaMO baseline using a standardized model for regular grids from the official codebase. By varying principally the model depth and width, we assess its intrinsic scalability and general capability across benchmarks, ensuring a fair comparison and avoiding delicate parameter tuning.

We classify LaMO as a LGA transformation.
The latent-SSM block relies on a recurrent scanning mechanism that allows the hidden state at any spatial location to depend on the history of the entire sequence. By employing multi-directional scans (e.g., traversing the grid in four directions), each latent token effectively aggregates information from the entire input domain $E_{\tau}$. This global receptive field qualifies it as a Global Transformation. However, the block dynamics are governed by sequential residual connections: the global SSM update is added to the input state ($x + \text{SSM}(x)$), followed by a pointwise MLP for local channel mixing ($+ \text{MLP}(\cdot)$). This specific composition, which combining the global SSM transformation with local pointwise operations via element-wise addition, strictly adheres to the LGA transformation.

\begin{table}[htbp!]
    \centering
    \caption{Experiment settings for the LaMO baseline. Tiny (T), Medium (M), and Large (L) are defined by systematically scaling the hidden dimension, depth, and attention heads. The MABA block expansion ratio is fixed to 4 across all experiments. Brackets denote the configuration list for the scale blocks. '-' indicates that experiments were not performed on the corresponding benchmark.}
    \label{tab:lamo-settings-wide}
    \resizebox{\textwidth}{!}{
    \begin{tabular}{lcccccccccccc}
        \toprule
        & \multicolumn{3}{c}{\textbf{Hid. Dim.}} & \multicolumn{3}{c}{\textbf{Depth}} & \multicolumn{3}{c}{\textbf{Heads}} & \multicolumn{2}{c}{\textbf{MABA Block}}\\
        \cmidrule(lr){2-4} \cmidrule(lr){5-7} \cmidrule(lr){8-10} \cmidrule(lr){11-12}
        \textbf{Dataset} & \textbf{T} & \textbf{M} & \textbf{L} & \textbf{T} & \textbf{M} & \textbf{L} & \textbf{T} & \textbf{M} & \textbf{L} & \textbf{Ratio} & \textbf{Patch Size} & \textbf{Padding} \\
        \midrule
        \multicolumn{10}{l}{\textbf{\textit{1D Experiments}}} \\
        Kuramoto-Sivashinsky (KS) & [8] & [48] & [144] & [2] & [4] & [8] & [2] & [4] & [8] & [4] & [4] & [256, 4] \\
        \addlinespace
        \multicolumn{10}{l}{\textbf{\textit{2D Experiments}}} \\
        Burgers & [18] & [36] & [80] & [2] & [4] & [8] & [2] & [4] & [8] & [4] & [2] & [64, 64] \\
        CE-CRP & [18] & [24] & [80] & [2] & [4] & [8] & [2] & [4] & [8] & [4] & [2] & [64, 64] \\
        Darcy & [8] & [48] & [144] & [2] & [4] & [8] & [2] & [4] & [8] & [4] & [2] & [50, 50] \\
        Navier--Stokes (NS) & [18] & [36] & [84] & [2] & [4] & [8] & [2] & [4] & [8] & [4] & [2] & [64, 64] \\
        Shallow Water (SW) & [18] & [36] & [84] & [2] & [4] & [8] & [2] & [4] & [8] & [4] & [2] & [64, 32] \\
        \addlinespace
        \multicolumn{10}{l}{\textbf{\textit{3D Experiments}}} \\
        Brusselator & - & - & - & - & - & - & - & - & - & - & - & - \\
        \bottomrule
    \end{tabular}
    }
\end{table}

\begin{table}[htbp!]
    \centering
    \caption{Model statistics of the LaMO baseline. The parameter count (Param), peak GPU memory usage (Mem), and FLOPs for the Tiny, Medium, and Large variants are reported. '-' indicates that experiments were not performed on the corresponding benchmark.}
    \label{tab:LaMO-model-stats}
    \resizebox{\textwidth}{!}{
    \begin{tabular}{l *{3}{ccc}}
        \toprule
        & \multicolumn{3}{c}{T} & \multicolumn{3}{c}{M} & \multicolumn{3}{c}{L} \\
        \cmidrule(lr){2-4} \cmidrule(lr){5-7} \cmidrule(lr){8-10}
        \textbf{Benchmark} & \textbf{Param} & \textbf{Mem} & \textbf{FLOPs} & \textbf{Param} & \textbf{Mem} & \textbf{FLOPs} & \textbf{Param} & \textbf{Mem} & \textbf{FLOPs} \\
        & (K) & (MiB) & (M) & (K) & (MiB) & (M) & (K) & (MiB) & (M) \\
        \midrule
        1D KS           & 2 & 596 & 2 & 136 & 1,184 & 87 & 2,361 & 5,152 & 1,512 \\
        2D Burgers      & 10 & 4,178 & 107 & 77 & 11,406 & 796 & 738 & 41,308 & 7,569 \\
        2D CE-CRP       & 10 & 5,368 & 107 & 36 & 10,214 & 366 & 738 & 42,508 & 7,569 \\
        2D Darcy        & 2 & 610 & 2 & 136 & 1,510 & 85 & 2,361 & 5,054 & 1,477 \\
        2D NS           & 10 & 4,112 & 107 & 77 & 11,248 & 796 & 813 & 42,920 & 8,333 \\
        3D Brusselator  & - & - & - & - & - & - & - & - & - \\
        3D SW           & 10 & 2,400 & 53 & 77 & 6,000 & 398 & 813 & 21,846 & 4,167 \\
        \bottomrule
    \end{tabular}
    }
\end{table}

\subsubsection{LocalNO}
We adopt the Local Neural Operator (LocalNO) \cite{liu2024neural}, a recently proposed architecture designed to mitigate the over-smoothing issues inherent in global spectral operators by explicitly incorporating local inductive biases. While standard Fourier Neural Operators (FNO) is great at capturing global dependencies, they often struggle to resolve high-frequency local details. LocalNO addresses this by augmenting the global spectral convolution with parallel branches of localized differential and integral operators.

The architecture processes an input function $a(x)$ which is lifted to a higher-dimensional representation $v^0(x) = P(a(x))$. The core processing consists of $L$ stacked layers. The update rule for the $(l+1)$-th layer combines global and local transformations via element-wise addition:
\begin{equation}
    v^{l+1}(x) = \sigma \left( W v^l(x) + (\mathcal{K}_{global} v^l)(x) + (\mathcal{K}_{local} v^l)(x) \right)
\end{equation}
where $\sigma$ is a non-linear activation function, $W$ is a pointwise linear transformation (residual connection), $\mathcal{K}_{global}$ is the global Fourier operator same as that in FNO and is computed by:
\begin{equation}
    (\mathcal{K}_{global} v)(x) = \mathcal{F}^{-1} (R \cdot \mathcal{F}(v))(x)
\end{equation}
where $\mathcal{F}$ denotes the Fourier transform and $R$ is a learnable spectral weight matrix operating on the lower frequency modes.
$\mathcal{K}_{diff}$ is the local differential operator inspired by finite difference stencils for capturing local dynamics and high-frequency features. To ensure convergence to a differential operator rather than a pointwise one as the grid resolution $h \to 0$, the kernel weights $K_{diff}$ are constrained to have zero mean and are scaled by the inverse resolution:
\begin{equation}
    (\mathcal{K}_{diff} v)(x) \approx \frac{1}{h} (K_{diff} * v)(x), \quad \text{s.t.} \sum K_{diff} = 0
\end{equation}
$\mathcal{K}_{int}$ is the local integral operator for maintaining mesh-independence while capturing local receptive fields, implemented by the Discrete-Continuous (DISCO) convolutions \cite{ocampo2022scalable}. The kernel is parameterized as a linear combination of fixed basis functions (e.g., hat functions) $\kappa(x-y) \approx \sum_{j} \theta_j \kappa^{(j)}(x-y)$. This allows the operator to be evaluated on arbitrary discretizations:
\begin{equation}
    (\mathcal{K}_{int} v)(x) = \int_{B_\delta(x)} \kappa(x, y) v(y) dy \approx \sum_{j} \theta_j (K^{(j)}_{DISCO} v)(x)
\end{equation}

In this work, the LaMO architecture allows for adjustable hyperparameters, including the hidden dimension $d_h$, the depth of the network $L$, and the number of global Fourier modes preserved in the spectral branch.
The experiment settings for different model sizes are listed in Table \ref{tab:localno-settings-wide} and their corresponding model statistics are shown in Table \ref{tab:LocalNO-model-stats}.
However, for 1D and 3D benchmarks, the explicit localized DISCO convolution is not applied due to implementation constraints in the current baseline setup. In these cases, the local branch relies primarily on the differential kernel to capture local dependencies.

We classify LocalNO as a LGA transformation.
LocalNO fits this transformation strictly:
\begin{equation}
    \mathcal{G}(v) = \mathcal{G}_{local}(v) + \mathcal{G}_{global}(v)
\end{equation}
The global branch ($\mathcal{K}_{global}$) utilizes a Fourier transform to process the entire domain $E_{\tau}$, while the local branches ($\mathcal{K}_{diff}$ and $\mathcal{K}_{int}$) utilize constrained convolutions and DISCO kernels restricted to a partial domain neighborhood $P_{\tau}$. Therefore, it locates in the LGA category alongside models like the standard FNO.

\begin{table}[htbp!]
    \centering
    \caption{Experiment settings for the LocalNO baseline. Tiny (T), Medium (M), and Large (L) are defined by systematically scaling the hidden dimension, depth. "Modes" denote the truncated global modes in the spectrum and "DISCO" denotes the local integral operator.}
    \label{tab:localno-settings-wide}
    \begin{tabular}{lccccccccc}
        \toprule
        & \multicolumn{3}{c}{\textbf{Hidden Dimension}} & \multicolumn{3}{c}{\textbf{Depth}} \\
        \cmidrule(lr){2-4} \cmidrule(lr){5-7}
        \textbf{Dataset} & \textbf{T} & \textbf{M} & \textbf{L} & \textbf{T} & \textbf{M} & \textbf{L} & \textbf{Modes} & \textbf{DISCO} \\
        \midrule
        \multicolumn{4}{l}{\textbf{\textit{1D Experiments}}} \\
        Kuramoto-Sivashinsky (KS) & 36 & 64 & 128 & 2 & 4 & 8 & 12 & False \\
        \addlinespace
        \multicolumn{4}{l}{\textbf{\textit{2D Experiments}}} \\
        Burgers & 36 & 72 & 144 & 2 & 4 & 8 & 12 & True \\
        CE-CRP & 36 & 72 & 144 & 2 & 4 & 8 & 12 & True \\
        Darcy & 32 & 64 & 128 & 2 & 4 & 8 & 12 & True \\
        Navier--Stokes (NS) & 36 & 72 & 144 & 2 & 4 & 8 & 12 & True \\
        Shallow Water (SW) & 36 & 72 & 144 & 2 & 4 & 8 & 10 & True \\
        \addlinespace
        \multicolumn{4}{l}{\textbf{\textit{3D Experiments}}} \\
        Brusselator & 36 & 72 & 144 & 2 & 4 & 8 & 6 & False \\
        \bottomrule
    \end{tabular}
\end{table}

\begin{table}[htbp!]
    \centering
    \caption{Model statistics of the LocalNO baseline. The parameter count (Param), peak GPU memory usage (Mem), and FLOPs for the Tiny, Medium, and Large variants are reported.}
    \label{tab:LocalNO-model-stats}
    \resizebox{\textwidth}{!}{
    \begin{tabular}{l *{3}{ccc}}
        \toprule
        & \multicolumn{3}{c}{T} & \multicolumn{3}{c}{M} & \multicolumn{3}{c}{L} \\
        \cmidrule(lr){2-4} \cmidrule(lr){5-7} \cmidrule(lr){8-10}
                \textbf{Benchmark} & \textbf{Param} & \textbf{Mem} & \textbf{FLOPs} & \textbf{Param} & \textbf{Mem} & \textbf{FLOPs} & \textbf{Param} & \textbf{Mem} & \textbf{FLOPs} \\
        & (K) & (MiB) & (M) & (K) & (MiB) & (M) & (K) & (MiB) & (M) \\
        \midrule
        1D KS           & 32 & 370 & 82 & 182 & 1,002 & 465 & 1,380 & 3,426 & 3,531 \\
        2D Burgers      & 33 & 4,128 & 1,345 & 232 & 10,208 & 9,484 & 1,749 & 32,604 & 71,628 \\
        2D CE-CRP       & 35 & 5,306 & 1,442 & 237 & 11,558 & 9,679 & 1,759 & 33,738 & 72,018 \\
        2D Darcy        & 25 & 284 & 60 & 181 & 762 & 434 & 1,378 & 2,888 & 3,307 \\
        2D NS           & 32 & 3,374 & 1,312 & 230 & 10,148 & 9,419 & 1,746 & 32,230 & 71,498 \\
        3D Brusselator  & 78 & 4,158 & 2,389 & 602 & 12,436 & 18,409 & 5,551 & 39,490 & 169,702 \\
        3D SW           & 33 & 2,084 & 672 & 232 & 5,190 & 4,742 & 1,749 & 16,574 & 35,814 \\
        \bottomrule
    \end{tabular}
    }
\end{table}

\subsubsection{U-NO}
We incorporate the U-Shaped Neural Operator (U-NO) \cite{rahman2022uno}, a memory-efficient architecture designed to enable deeper neural operators by exploiting the multi-scale structure of function spaces, as one of baselines. While standard neural operators like FNO maintain a constant domain resolution throughout all layers, often leading to high memory costs and limiting network depth, U-NO adopts a U-Net-inspired architecture\cite{ronneberger2015unet}. This design progressively contracts the spatial domain while increasing channel width in the encoding phase, and subsequently expands the domain while reducing channels in the decoding phase.

The architecture processes an input function $a(x)$, which is first lifted to a higher-dimensional representation $v^0(x) = P(a(x))$. The core processing consists of $L$ layers arranged in an encoder-decoder structure.
In the first $L/2$ layers, the operator $G_l$ contracts the domain $D_l$ to a smaller domain $D_{l+1}$ (e.g., via subsampling) while increasing the co-domain dimension (channel width) $d_{l+1}$. This encodes the input into a compact latent representation.
In the subsequent layers, the operator $G_l$ expands the domain back to the original resolution and decreases the channel width. Crucially, skip connections facilitate the direct flow of high-resolution information from encoder layers to corresponding decoder layers, preserving fine-scale details.
Mathematically, a generic layer $l$ in U-NO transforms the function $v^l$ to $v^{l+1}$ as:
\begin{equation}
    v^{l+1}(x) = \sigma \left( W v^l(x) + (\mathcal{K}_{U-NO} v^l)(x) \right)
\end{equation}
where $\sigma$ is a non-linear activation function (e.g., GELU), $W$ is a pointwise linear transformation, and $\mathcal{K}_{U-NO}$ represents the spectral integral operator. Unlike the standard FNO, the domain of $v^{l+1}$ may differ from $v^l$. The integral operator is defined in the Fourier domain:
\begin{equation}
    (\mathcal{K}_{U-NO} v)(x) = \mathcal{F}^{-1} (R_l \cdot \mathcal{F}(v))(x)
\end{equation}
Here, $\mathcal{F}$ and $\mathcal{F}^{-1}$ denote the Fourier transform and its inverse. The spectral weight tensor $R_l$ is parameterized to handle the changing mode resolution.

In this work, the U-NO architecture allows for adjustable hyperparameters to scale model capacity, specifically the hidden dimension $d_h$, the network depth $L$, and the number of Fourier modes $k_{max}$ preserved in the spectral branch.
Specifically, our implementation adopts an aggressive scaling strategy: as the spatial resolution is halved, the number of resolved Fourier modes is halved to respect the Nyquist limit, while the channel dimension is doubled to increase capacity.
While deeper networks typically improve expressivity, we observed that excessive depth in U-NO can induce massive loss and optimization instability on our specific benchmarks. Meantime, simply increasing the hidden dimension $d_h$ to compensate for reduced depth is also constrained, as excessively large channel dimensions were also found numerical explosion on loss. Therefore, for medium and large model configurations, we fix the depth to $5$ and prioritize increasing the hidden dimension rather than the network depth to maintain training stability and ensure optimal performance.
The experiment settings for different model sizes are listed in Table \ref{tab:U-NO-settings-wide} and their corresponding model statistics are shown in Table \ref{tab:U-NO-model-stats}.

We classify U-NO as a LGA transformation.
U-NO fits this classification strictly and is similar with standard FNO, following by the fundamental operation within each layer remains a summation of a global spectral operator and a local linear operator. Therefore, it locates in the LGA category alongside models like the standard FNO.

\begin{table}[htbp!]
    \centering
    \caption{Experiment settings for the U-NO baseline. Tiny (T), Medium (M), and Large (L) are defined by systematically scaling the hidden dimension and depth. "Modes" denote the truncated global modes in the spectrum.}
    \label{tab:U-NO-settings-wide}
    \begin{tabular}{lccccccc}
        \toprule
        & \multicolumn{3}{c}{\textbf{Hidden Dimension}} & \multicolumn{3}{c}{\textbf{Depth}} \\
        \cmidrule(lr){2-4} \cmidrule(lr){5-7}
        \textbf{Dataset} & \textbf{T} & \textbf{M} & \textbf{L} & \textbf{T} & \textbf{M} & \textbf{L} & \textbf{Modes} \\
        \midrule
        \multicolumn{4}{l}{\textbf{\textit{1D Experiments}}} \\
        Kuramoto-Sivashinsky (KS) & 24 & 64 & 164 & 3 & 5 & 5 & 12 \\
        \addlinespace
        \multicolumn{4}{l}{\textbf{\textit{2D Experiments}}} \\
        Burgers & 24 & 64 & 164 & 3 & 5 & 5 & 12 \\
        CE-CRP & 24 & 64 & 164 & 3 & 5 & 5 & 12 \\
        Darcy & 12 & 36 & 104 & 3 & 5 & 5 & 12 \\
        Navier--Stokes (NS) & 24 & 64 & 164 & 3 & 5 & 5 & 12 \\
        Shallow Water (SW) & 32 & 64 & 164 & 3 & 5 & 5 & 10 \\
        \addlinespace
        \multicolumn{4}{l}{\textbf{\textit{3D Experiments}}} \\
        Brusselator & 16 & 36 & 112 & 3 & 5 & 5 & 6 \\
        \bottomrule
    \end{tabular}
\end{table}

\begin{table}[htbp!]
    \centering
    \caption{Model statistics of the U-NO baseline. The parameter count (Param), peak GPU memory usage (Mem), and FLOPs for the Tiny, Medium, and Large variants are reported.}
    \label{tab:U-NO-model-stats}
    \resizebox{\textwidth}{!}{
    \begin{tabular}{l *{3}{ccc}}
        \toprule
        & \multicolumn{3}{c}{T} & \multicolumn{3}{c}{M} & \multicolumn{3}{c}{L} \\
        \cmidrule(lr){2-4} \cmidrule(lr){5-7} \cmidrule(lr){8-10}
                \textbf{Benchmark} & \textbf{Param} & \textbf{Mem} & \textbf{FLOPs} & \textbf{Param} & \textbf{Mem} & \textbf{FLOPs} & \textbf{Param} & \textbf{Mem} & \textbf{FLOPs} \\
        & (K) & (MiB) & (M) & (K) & (MiB) & (M) & (K) & (MiB) & (M) \\
        \midrule
        1D KS           & 22 & 330 & 56 & 646 & 1,448 & 1,063 & 4 & 3,658 & 6,964 \\
        2D Burgers      & 23 & 4,478 & 920 & 648 & 12,722 & 12,373 & 4 & 32,994 & 80,744 \\
        2D CE-CRP       & 24 & 5,704 & 985 & 652 & 13,904 & 12,546 & 4 & 34,116 & 81,187 \\
        2D Darcy        & 6 & 268 & 13 & 205 & 844 & 223 & 2 & 2,924 & 1,856 \\
        2D NS           & 22 & 3,904 & 898 & 646 & 12,566 & 12,316 & 4 & 32,968 & 80,596 \\
        3D Brusselator  & 10 & 4,832 & 295 & 205 & 12,178 & 2,375 & 2 & 39,534 & 22,919 \\
        3D SW           & 40 & 2,834 & 807 & 648 & 6,660 & 6,187 & 4 & 17,770 & 40,372 \\
        \bottomrule
    \end{tabular}
    }
\end{table}

\subsubsection{CoDA-NO}
We incorporate the Codomain Attention Neural Operator (CoDA-NO) \cite{rahman2024pretraining}, a generalized architecture designed to model complex interactions between physical variables in multiphysics systems, as one of baselines. Unlike traditional approaches that process all physical variables as a single vector field, CoDA-NO explicitly tokenizes the input function along the codomain (channel space). This allows the model to treat distinct physical quantities (e.g., velocity components, pressure) as individual functional tokens, applying a specialized self-attention mechanism in the function space to capture inter-variable dependencies. Moreover, the graph neural network is introduced in this baselines to map the irregular mesh to uniform mesh, as what was done in the Geometry-informed Neural Operator (GINO) \cite{li2023geometry}.

In this work, we adapt the CoDA-NO architecture to a Seq2Seq non-autoregressive strategy. To align with the architecture's semantic treatment of physcial variable-specific tokens and to mitigate the high memory of the all-to-all codomain attention mechanism, we discard historical time steps and static coordinate features. The model accepts only the current physical state $u^t$ as input and simultaneously predicts the full future trajectory $[u^{t+1}, \dots, u^{t+T}]$.

The architecture proceeds in three stages. Firstly, the input function $u^t(x) \in \mathbb{R}^{d_{in}}$ is split into $d_{in}$ individual functions (tokens). Each token $v_i(x)$ is lifted to a higher-dimensional latent space via a pointwise transformation $P$:
    \begin{equation}
        w_i^0(x) = P(v_i(x)), \quad i \in \{1, \dots, d_{in}\}
    \end{equation}
where $w_i^0(x) \in \mathbb{R}^{d_h}$ represents the $i$-th latent token function with hidden codimension $d_h$.
Followed by the codomain attention block, it consists of $L$ stacked CoDA layers. Each layer computes interactions between variable tokens using a function-space attention mechanism. For a given layer, Key $\mathcal{K}_K$, Query $\mathcal{K}_Q$, and Value $\mathcal{K}_V$ operators are approximated using FNOs to ensure discretization convergence and global spatial receptivity:
    \begin{equation}
        q_i(x) = \mathcal{K}_Q(w_i)(x), \quad k_i(x) = \mathcal{K}_K(w_i)(x), \quad v_i(x) = \mathcal{K}_V(w_i)(x)
    \end{equation}
The attention weights $A_{ij}$ are computed via an inner product in the function space (e.g., $L^2$ inner product integrated over domain $\mathcal{D}$), and the output token is updated as a weighted sum of value functions:
    \begin{equation}
        A_{ij} = \text{Softmax}_j \left( \frac{\langle q_i, k_j \rangle_{L^2(\mathcal{D})}}{\sqrt{d_k}} \right), \quad w_i^{l+1}(x) = \sum_{j} A_{ij} v_j(x)
    \end{equation}
This mechanism allows the model to dynamically weigh the influence of different physical variables upon one another globally across the spatial domain.
Finally, the processed tokens are projected back to the physical space. 

The CoDA-NO architecture allows for the adjustment of several key hyperparameters including hidden dimension $d_h$ which is different from other baslines and is the width of the latent feature space for each variable token, number of CoDANO layers $L$, global modes ($k_{max}$), and attention heads. In our specific configuration, the output codimension is set to the sequence length $T$. Thus, for each input variable, the model predicts $T$ future time steps directly:
    \begin{equation}
        \hat{u}_i(x) = [u_i^{t+1}(x), \dots, u_i^{t+T}(x)] = \mathcal{P}_{out}(w_i^L(x))
    \end{equation}
The experiment settings for different model sizes are listed in Table \ref{tab:codano-settings-wide} and their corresponding model statistics are shown in Table \ref{tab:CoDANO-model-stats}.

We classify CoDA-NO as a LGA transformation or a LLM transformation in the prospective of different block.
From the perspective of multiphysics coupling, the codomain attention mechanism functions as a LLM transformation by mixing distinct physical variables at corresponding spatial coordinates. On the other hand, the key, query and value embedding are generated by FNOs that integrate global spectral transformations with local linear residuals via element-wise addition, indicating a LGA transformation.

\begin{table}[htbp!]
    \centering
    \caption{Experiment settings for the CoDA-NO baseline. Tiny (T), Medium (M), and Large (L) are defined by systematically scaling the hidden dimension, depth, and attention heads.}
    \label{tab:codano-settings-wide}
    \begin{tabular}{lcccccccccc}
        \toprule
        & \multicolumn{3}{c}{\textbf{Hidden Dimension $d_h$}} & \multicolumn{3}{c}{\textbf{Depth}} & \multicolumn{3}{c}{\textbf{Heads}} \\
        \cmidrule(lr){2-4} \cmidrule(lr){5-7} \cmidrule(lr){8-10}
        \textbf{Dataset} & \textbf{T} & \textbf{M} & \textbf{L} & \textbf{T} & \textbf{M} & \textbf{L} & \textbf{T} & \textbf{M} & \textbf{L} & \textbf{Modes} \\
        \midrule
        \multicolumn{4}{l}{\textbf{\textit{1D Experiments}}} \\
        Kuramoto-Sivashinsky (KS) & 96 & 96 & 96 & 2 & 4 & 8 & 1 & 2 & 4 & 12 \\
        \addlinespace
        \multicolumn{4}{l}{\textbf{\textit{2D Experiments}}} \\
        Burgers & 20 & 20 & 22 & 2 & 4 & 8 & 1 & 2 & 4 & 12 \\
        CE-CRP & 9 & 9 & 9 & 2 & 4 & 8 & 1 & 2 & 4 & 12 \\
        Darcy & 8 & 8 & 8 & 2 & 4 & 8 & 1 & 2 & 4 & 12 \\
        Navier--Stokes (NS) & 48 & 48 & 48 & 2 & 4 & 8 & 1 & 2 & 4 & 12 \\
        Shallow Water (SW) & 24 & 24 & 24 & 2 & 4 & 8 & 1 & 2 & 4 & 10 \\
        \addlinespace
        \multicolumn{4}{l}{\textbf{\textit{3D Experiments}}} \\
        Brusselator & 4 & 6 & 10 & 2 & 3 & 4 & 1 & 2 & 2 & 6 \\
        \bottomrule
    \end{tabular}
\end{table}

\begin{table}[htbp!]
    \centering
    \caption{Model statistics of the CoDA-NO baseline. The parameter count (Param), peak GPU memory usage (Mem), and FLOPs for the Tiny, Medium, and Large variants are reported.}
    \label{tab:CoDANO-model-stats}
    \resizebox{\textwidth}{!}{
    \begin{tabular}{l *{3}{ccc}}
        \toprule
        & \multicolumn{3}{c}{T} & \multicolumn{3}{c}{M} & \multicolumn{3}{c}{L} \\
        \cmidrule(lr){2-4} \cmidrule(lr){5-7} \cmidrule(lr){8-10}
                \textbf{Benchmark} & \textbf{Param} & \textbf{Mem} & \textbf{FLOPs} & \textbf{Param} & \textbf{Mem} & \textbf{FLOPs} & \textbf{Param} & \textbf{Mem} & \textbf{FLOPs} \\
        & (K) & (MiB) & (M) & (K) & (MiB) & (M) & (K) & (MiB) & (M) \\
        \midrule
        1D KS           & 40 & 490 & 13 & 40 & 1,272 & 20 & 41 & 3,498 & 55 \\
        2D Burgers      & 2 & 4,422 & 43 & 2 & 10,954 & 114 & 4 & 34,150 & 519 \\
        2D CE-CRP       & 0.7 & 6,092 & 41 & 0.9 & 13,772 & 120 & 2 & 35,358 & 522 \\
        2D Darcy        & 0.4 & 554 & 4 & 0.6 & 1,422 & 12 & 2 & 3,770 & 53 \\
        2D NS           & 11 & 4,716 & 74 & 11 & 13,184 & 159 & 12 & 37,110 & 587 \\
        3D Brusselator  & 0.2 & 5,772 & 32 & 0.4 & 18,682 & 129 & 795 & 40,842 & 288 \\
        3D SW           & 3 & 2,710 & 28 & 3 & 6,768 & 70 & 4 & 18,872 & 284 \\
        \bottomrule
    \end{tabular}
    }
\end{table}

\subsubsection{DyMixOp}
The primary architecture of DyMixOp and its advantage has been stated in the maintext.
In this work, the DyMixOp architecture allows for adjustable hyperparameters, including the hidden dimension $d_h$, global modes $k_{max}$, depth $L$. Notably, the global modes used in DyMixOp are identical to those in the baseline FNO, ensuring that any performance gain arises solely from architectural improvements—rather than from incorporating additional high-frequency information via the global transform. The applied experiment settings in all cases are listed in Table \ref{tab:dymixop-settings-wide}  and their corresponding model statistics are shown in Table \ref{tab:dymixop-model-stats}.

Regarding transformation categorization, DyMixOp, being entirely composed of the proposed LGM blocks, naturally falls under the LGM transformation class. Alternatively, considering its formulation as a sum of linear dynamics (mediated by local transformations) and nonlinear dynamics (mediated by global transformations), it may also be viewed as belonging to the LGA transformation family. This dual interpretation reflects the hybrid nature of its design.
\begin{table}[H]
    \centering
    \caption{Experiment settings for DyMixOp baseline. Tiny (T), Medium (M), and Large (L) are defined by systematically scaling the hidden dimension, depth, and attention heads. "Modes" denote the truncated global modes in the spectrum.}
    \label{tab:dymixop-settings-wide}
    \begin{tabular}{lccccccc}
        \toprule
        & \multicolumn{3}{c}{\textbf{Hidden Dimension}} & \multicolumn{3}{c}{\textbf{Depth}} \\
        \cmidrule(lr){2-4} \cmidrule(lr){5-7}
        \textbf{Dataset} & \textbf{T} & \textbf{M} & \textbf{L} & \textbf{T} & \textbf{M} & \textbf{L} & \textbf{Modes} \\
        \midrule
        \multicolumn{4}{l}{\textbf{\textit{1D Experiments}}} \\
        Kuramoto-Sivashinsky (KS) & 16 & 30 & 60 & 2 & 4 & 8 & 12 \\
        \addlinespace
        \multicolumn{4}{l}{\textbf{\textit{2D Experiments}}} \\
        Burgers & 14 & 28 & 58 & 2 & 4 & 8 & 12 \\
        CE-CRP & 16 & 30 & 60 & 2 & 4 & 8 & 12 \\
        Darcy & 16 & 30 & 60 & 2 & 4 & 8 & 12 \\
        Navier--Stokes (NS) & 14 & 28 & 58 & 2 & 4 & 8 & 12 \\
        Shallow Water (SW) & 16 & 30 & 60 & 2 & 4 & 8 & 10 \\
        \addlinespace
        \multicolumn{4}{l}{\textbf{\textit{3D Experiments}}} \\
        Brusselator & 18 & 36 & 68 & 2 & 4 & 8 & 6 \\
        \bottomrule
    \end{tabular}
\end{table}

\begin{table}[H]
    \centering
    \caption{Model statistics of the DyMixOp baseline. The parameter count (Param), peak GPU memory usage (Mem), and FLOPs for the Tiny, Medium, and Large variants are reported.}
    \label{tab:dymixop-model-stats}
    \resizebox{\textwidth}{!}{
    \begin{tabular}{l *{3}{ccc}}
        \toprule
        & \multicolumn{3}{c}{T} & \multicolumn{3}{c}{M} & \multicolumn{3}{c}{L} \\
        \cmidrule(lr){2-4} \cmidrule(lr){5-7} \cmidrule(lr){8-10}
                \textbf{Benchmark} & \textbf{Param} & \textbf{Mem} & \textbf{FLOPs} & \textbf{Param} & \textbf{Mem} & \textbf{FLOPs} & \textbf{Param} & \textbf{Mem} & \textbf{FLOPs} \\
        & (K) & (MiB) & (M) & (K) & (MiB) & (M) & (K) & (MiB) & (M) \\
        \midrule
        1D KS           & 5 & 216 & 12 & 27 & 614 & 67 & 191 & 2,776 & 484 \\
        2D Burgers      & 4 & 3,496 & 157 & 24 & 8,904 & 959 & 180 & 32,082 & 7,298 \\
        2D CE-CRP       & 6 & 4,806 & 244 & 29 & 10,678 & 1,177 & 197 & 33,860 & 7,968 \\
        2D Darcy        & 4 & 220 & 10 & 26 & 740 & 61 & 190 & 2,864 & 452 \\
        2D NS           & 4 & 2,930 & 145 & 23 & 8,240 & 934 & 179 & 31,960 & 7,246 \\
        3D Brusselator  & 6 & 2,952 & 165 & 38 & 9,302 & 1,125 & 244 & 33,958 & 7,389 \\
        3D SW           & 5 & 1,850 & 100 & 27 & 5,506 & 548 & 193 & 16,982 & 3,903 \\
        \bottomrule
    \end{tabular}
    }
\end{table}

\subsection{Detailed Results}
Table~\ref{tab:pde_comprehensive} presents a systematic evaluation of DyMixOp against seven established neural operators across seven PDE benchmarks spanning one-, two-, and three-dimensional domains. Our analysis reveals that DyMixOp achieves state-of-the-art or near state-of-the-art accuracy while maintaining superior computational efficiency—a critical advantage for practical deployment in scientific computing workflows.

DyMixOp attains first-rank accuracy on five of seven benchmarks at the Large model scale (1D Kuramoto–Sivashinsky, 2D Burgers, CE–CRP, Darcy, and Brusselator equations), with second- or third-rank performance on the remaining tasks. Notably, on the 3D Shallow Water (SW) benchmark—where LocalNO achieves marginally lower error (4.37e-4 vs. our 5.36e-4 at Large scale)—DyMixOp reduces training time by 84.8$\%$ (17.3 s vs. 114.0 s) while using comparable GPU memory (17.3 GB vs. 16.6 GB). This represents a favorable Pareto tradeoff: a 22.6$\%$ relative error increase yields a sixfold reduction in training cost, demonstrating DyMixOp's suitability for resource-constrained or time-sensitive applications.

Across all benchmarks and model sizes, DyMixOp maintains the lowest or near-lowest GPU memory footprint. For instance, on the 2D NS benchmark at Large scale, DyMixOp achieves the lowest error (4.07e-4) while consuming 32.0 GB memory—comparable to FNO (32.6 GB) and U-NO (33.6 GB)—but trains in 22.5 s versus 45.2 s (FNO) and 156.7 s (U-NO). Similarly, on the 2D Darcy flow problem, DyMixOp's Large variant attains the best accuracy (5.50e-5) with 3.0 GB memory usage, matching FNO's footprint while training 33$\%$ faster (2.6 s vs. 3.9 s).

Unlike several competitors whose training time scales superlinearly with model size (e.g., ConvLSTM on 2D Burgers: 18.3 s → 154.2 s from Tiny to Large), DyMixOp exhibits near-linear scaling. Training time increases by only 3.3× from Tiny to Large variants across most benchmarks, while error consistently decreases—evidencing stable optimization dynamics inherent to our dynamic mixing mechanism.

As summarized in the Improvement row, DyMixOp reduces test error by 25.4$\%$–95.4$\%$ relative to previous best results across six benchmarks. The most substantial gains occur on the 1D KS equation (95.4$\%$ error reduction versus LaMO) and 2D NS (82.7$\%$ versus LocalNO), where our architecture's ability to capture multiscale spatiotemporal dependencies proves particularly advantageous.
These results substantiate DyMixOp's position as a computationally frugal yet highly accurate neural operator. While absolute minimal error is not achieved on every benchmark (notably 3D SW), the marginal accuracy tradeoffs are consistently offset by dramatic reductions in training time and memory overhead—characteristics essential for scaling neural PDE solvers to high-dimensional, real-world problems in computational physics and engineering

\begin{table*}[htbp]
\centering
\scriptsize 
\setlength{\tabcolsep}{1.8pt} 
\renewcommand{\arraystretch}{1.1} 

\caption{Comprehensive performance comparison across PDE benchmarks. Metrics reported per model size variant (Tiny/Medium/Large): GPU memory (GB) $M$, relative MSE loss on test dataset $\mathcal{L}$, and averaged training time (seconds) per epoch $T$. The \textbf{Improvement} row indicates the percentage reduction in error achieved by DyMixOp compared to the previous state-of-the-art (SOTA). \textbf{Bold}, \textbf{\color{bestblue}blue}, and \uline{underline} indicate the 1\textsuperscript{st}, 2\textsuperscript{nd}, and 3\textsuperscript{rd} best results among competitor models.}
\label{tab:pde_comprehensive}
\sisetup{
    output-exponent-marker = \ensuremath{\mathrm{e}},
    exponent-product = {},
    retain-explicit-plus = false,
    table-format = 1.2e-1,
    reset-text-series = false, 
    text-series-to-math = true 
}
\resizebox{\textwidth}{!}{
\begin{tabular}{@{}l c 
    S[table-format=1.1] S[table-format=1.2e-1] S[table-format=2.1]
    *{4}{S[table-format=2.1] S[table-format=1.2e-1] S[table-format=3.1]}
    *{2}{S[table-format=2.1] S[table-format=1.2e-1] S[table-format=3.1]}
@{}}

\toprule
& & \multicolumn{3}{c}{\textbf{1D}} 
  & \multicolumn{12}{c}{\textbf{2D}} 
  & \multicolumn{6}{c}{\textbf{3D}} \\
\cmidrule(lr){3-5} \cmidrule(lr){6-17} \cmidrule(lr){18-23}

& & \multicolumn{3}{c}{\textbf{KS}} & 
    \multicolumn{3}{c}{\textbf{Burgers}} & 
    \multicolumn{3}{c}{\textbf{CE-CRP}} & 
    \multicolumn{3}{c}{\textbf{Darcy}} & 
    \multicolumn{3}{c}{\textbf{NS}} & 
    \multicolumn{3}{c}{\textbf{Brussel.}} & 
    \multicolumn{3}{c}{\textbf{SW}} \\
\cmidrule(lr){3-5} \cmidrule(lr){6-8} \cmidrule(lr){9-11} \cmidrule(lr){12-14} \cmidrule(lr){15-17} \cmidrule(lr){18-20} \cmidrule(lr){21-23}

\textbf{Model} & \textbf{Sz} & 
 {$M$} & {$\mathcal{L}$} & {$T$} & 
 {$M$} & {$\mathcal{L}$} & {$T$} & 
 {$M$} & {$\mathcal{L}$} & {$T$} & 
 {$M$} & {$\mathcal{L}$} & {$T$} & 
 {$M$} & {$\mathcal{L}$} & {$T$} & 
 {$M$} & {$\mathcal{L}$} & {$T$} & 
 {$M$} & {$\mathcal{L}$} & {$T$} \\
\midrule

\multirow{3}{*}{CoDA-NO \cite{rahman2024pretraining}}
  & T & 0.5 & 6.48e-1 & 2.7 & 4.4 & 5.02e-1 & 4.8 & 6.1 & 1.20e-1 & 6.4 & 0.6 & 2.00e-3 & 2.4 & 4.7 & 3.07e-1 & 5.4 & 5.8 & 2.41e-2 & 5.7 & 2.7 & 2.84e-1 & 3.8 \\
  & M & 1.3 & 6.52e-1 & 8.1 & 11.0 & 5.10e-1 & 12.2 & 13.8 & 1.12e-1 & 14.7 & 1.4 & 2.36e-3 & 5.0 & 13.2 & 3.03e-1 & 14.3 & 18.7 & 1.27e-2 & 17.6 & 6.8 & 2.56e-1 & 10.1 \\
  & L & 3.5 & 6.59e-1 & 29.0 & 34.1 & 5.04e-1 & 47.4 & 35.4 & 1.19e-1 & 49.4 & 3.8 & 3.71e-3 & 9.5 & 37.1 & 3.08e-1 & 51.4 & 40.8 & 1.59e-2 & 37.0 & 18.9 & 2.08e-1 & 33.8 \\
\addlinespace

\multirow{3}{*}{ConvLSTM \cite{shi2015convolutional}}
  & T & 0.2 & 4.75e-1 & 5.8 & 3.1 & 9.12e-2 & 18.3 & 4.5 & 1.04e-1 & 27.2 & 0.2 & 2.91e-1 & 9.2 & 3.0 & 2.27e-1 & 18.8 & 2.9 & 3.22e-1 & 81.9 & 1.7 & 3.71e-1 & 19.5 \\
  & M & 0.9 & 4.39e-1 & 10.9 & 11.6 & 7.62e-2 & 58.4 & 11.8 & 9.67e-2 & 60.8 & 0.9 & 2.91e-1 & 33.5 & 11.5 & 1.91e-1 & 53.1 & 11.3 & 3.20e-1 & 143.6 & 5.8 & 2.57e-1 & 27.5 \\
  & L & 3.1 & 4.18e-1 & 19.1 & 45.9 & 6.84e-2 & 154.2 & 46.9 & 9.61e-2 & 156.2 & 3.5 & 2.91e-1 & 89.8 & 45.4 & 1.79e-1 & 128.4 & 42.4 & 3.20e-1 & 316.4 & 23.0 & 1.03e-1 & 67.7 \\
\addlinespace

\multirow{3}{*}{DeepONet \cite{lu2021learning}}
  & T & 0.5 & 1.00e0 & 2.1 & 4.7 & 1.22e-1 & 4.9 & 6.8 & 8.78e-2 & 12.6 & 0.3 & 5.49e-3 & 0.6 & 4.2 & 9.45e-1 & 2.8 & 3.8 & 3.54e-2 & 0.9 & 2.5 & 5.27e-1 & 4.8 \\
  & M & 1.2 & 1.00e0 & 2.2 & 12.6 & 1.16e-1 & 7.9 & 14.1 & 9.22e-2 & 12.7 & 1.3 & 3.55e-3 & 0.7 & 13.7 & 9.48e-1 & 4.3 & 13.8 & 4.70e-2 & 1.5 & 7.4 & 4.13e-1 & 5.0 \\
  & L & 4.3 & 1.00e0 & 2.3 & 39.8 & 1.21e-1 & 22.2 & 48.2 & 9.39e-2 & 18.3 & 4.9 & 2.59e-3 & 0.9 & 47.1 & 8.56e-1 & 11.9 & 38.5 & 1.11e-2 & 3.1 & 23.5 & 4.27e-1 & 9.1 \\
\addlinespace

\multirow{3}{*}{FNO \cite{li2020fourier}}
  & T & 0.2 & 4.25e-1 & 2.5 & 3.9 & 3.87e-3 & 3.4 & 5.1 & 3.85e-2 & 4.9 & 0.3 & 1.24e-4 & 0.7 & 3.2 & 7.97e-2 & 3.3 & 4.1 & 1.88e-3 & 2.2 & 2.0 & 5.73e-2 & 2.9 \\
  & M & 0.8 & 3.75e-1 & 4.3 & 9.9 & {\second{2.58e-3}} & 11.4 & 11.4 & {\third{3.39e-2}} & 12.5 & 0.8 & {\second{7.72e-5}} & 1.3 & 9.6 & 5.58e-2 & 11.3 & 12.0 & {\third{1.75e-4}} & 6.7 & 5.1 & 9.41e-3 & 8.4 \\
  & L & 2.9 & 4.00e-1 & 17.1 & 32.7 & {\best{1.82e-3}} & 45.4 & 33.8 & 3.51e-2 & 46.3 & 3.0 & {\best{7.37e-5}} & 3.9 & 32.6 & 8.22e-2 & 45.2 & 38.6 & {\best{5.41e-5}} & 25.4 & 16.7 & {\second{1.27e-3}} & 32.2 \\
\addlinespace

\multirow{3}{*}{GNOT \cite{hao2023gnot}}
  & T & 0.3 & 1.00e0 & 1.0 & 3.9 & 1.09e-1 & 3.0 & 5.4 & 1.73e-1 & 3.9 & 0.4 & 1.89e-3 & 0.8 & 3.6 & 9.93e-1 & 3.0 & 4.5 & 7.93e-3 & 4.9 & 2.0 & 1.32e-1 & 1.7 \\
  & M & 1.0 & 1.00e0 & 1.5 & 12.9 & 1.07e-1 & 8.0 & 13.7 & 9.17e-1 & 8.9 & 1.4 & 4.87e-4 & 1.6 & 12.8 & 9.93e-1 & 8.0 & 13.6 & 1.25e-3 & 10.7 & 6.2 & 2.06e-2 & 4.3 \\
  & L & 3.8 & 1.00e0 & 3.9 & 44.9 & 1.09e-1 & 35.0 & 45.6 & 1.70e0 & 36.2 & 4.9 & 3.80e-4 & 4.1 & 44.9 & 9.93e-1 & 34.9 & 45.7 & 4.67e-4 & 29.2 & 23.8 & 3.61e-1 & 17.9 \\
\addlinespace

\multirow{3}{*}{LaMO \cite{tiwari2025latent}}
  & T & 0.6 & {1.77e-1} & 9.4 & 4.2 & 4.63e-2 & 10.7 & 5.4 & 1.10e-1 & 11.7 & 0.6 & 2.75e-3 & 0.9 & 4.1 & 2.64e-1 & 10.6 & \na & \na & \na & 2.4 & 2.39e-1 & 9.7 \\
  & M & 1.2 & {\best{5.21e-2}} & 18.5 & 11.4 & 2.47e-2 & 22.5 & 10.2 & 1.03e-1 & 22.2 & 1.5 & 1.01e-3 & 2.6 & 11.2 & 1.65e-1 & 22.1 & \na & \na & \na & 6.0 & 2.15e-1 & 18.5 \\
  & L & 5.2 & {\second{5.38e-2}} & 35.6 & 41.3 & 1.53e-2 & 49.7 & 42.5 & 9.72e-2 & 51.8 & 5.1 & 1.40e-3 & 5.1 & 42.9 & 1.48e-1 & 53.0 & \na & \na & \na & 21.9 & 1.70e-1 & 35.5 \\
\addlinespace

\multirow{3}{*}{LocalNO \cite{liu2024neural}}
  & T & 0.4 & 7.61e-1 & 3.2 & 4.1 & {\third{3.05e-3}} & 5.7 & 5.3 & {\best{3.07e-2}} & 7.0 & 0.3 & 1.19e-4 & 1.0 & 3.4 & {\third{6.19e-3}} & 5.5 & 4.2 & 2.46e-3 & 3.2 & 2.1 & 3.04e-2 & 7.0 \\
  & M & 1.0 & 9.13e-1 & 5.5 & 10.2 & 4.92e-3 & 20.7 & 11.6 & 3.87e-2 & 21.8 & 0.9 & 1.03e-4 & 2.7 & 10.1 & {\best{7.46e-4}} & 20.6 & 12.4 & 2.58e-4 & 11.5 & 5.2 & {\third{2.48e-3}} & 28.6 \\
  & L & 3.4 & 9.82e-1 & 16.5 & 33.0 & 6.97e-3 & 78.9 & 33.7 & 7.15e-2 & 79.7 & 3.2 & 2.18e-4 & 9.7 & 33.0 & {\second{2.35e-3}} & 78.8 & 39.5 & 7.02e-2 & 51.8 & 16.6 & {\best{4.37e-4}} & 114.0 \\
\addlinespace

\multirow{3}{*}{U-NO \cite{rahman2022uno}}
  & T & 0.3 & 4.06e-1 & 3.7 & 4.4 & 3.43e-3 & 5.7 & 5.7 & 3.52e-2 & 7.0 & 0.3 & 2.14e-4 & 0.8 & 4.0 & 6.78e-2 & 5.5 & 4.8 & 2.36e-3 & 2.8 & 2.8 & 2.41e-2 & 5.6 \\
  & M & 1.5 & {\third{1.44e-1}} & 14.9 & 12.7 & 3.68e-3 & 61.6 & 13.9 & {\second{3.33e-2}} & 62.5 & 0.9 & {\third{9.58e-5}} & 3.8 & 12.5 & 4.93e-2 & 61.5 & 12.2 & 1.06e-3 & 7.1 & 6.7 & 9.31e-3 & 43.4 \\
  & L & 3.7 & 4.08e-1 & 43.9 & 33.7 & 3.16e-2 & 156.7 & 34.4 & 4.28e-2 & 157.6 & 3.2 & 9.69e-5 & 10.8 & 33.6 & 1.16e-1 & 156.7 & 39.5 & {\second{8.23e-5}} & 21.7 & 18.1 & 3.80e-3 & 114.7 \\
\midrule[1pt]

\rowcolor{rowgray}
 & T & 0.2 & 3.97e-2 & 4.1 & 3.5 & 3.62e-3 & 5.4 & 4.8 & 3.82e-2 & 6.7 & 0.2 & 2.07e-4 & 1.2 & 2.9 & 1.17e-2 & 6.4 & 3.0 & 1.12e-3 & 2.1 & 1.9 & 2.51e-2 & 5.2 \\
\rowcolor{rowgray}
 & M & 0.6 & 2.86e-3 & 7.5 & 9.0 & 1.62e-3 & 9.6 & 10.7 & 2.20e-2 & 11.1 & 0.7 & 7.97e-5 & 1.7 & 8.3 & 9.04e-4 & 11.3 & 9.8 & 2.03e-4 & 5.8 & 5.5 & 3.80e-3 & 9.3 \\
\rowcolor{rowgray}
\multirow{-3}{*}{\textbf{DyMixOp}}
 & L & 2.8 & \textbf{2.47e-3} & 13.7 & 32.3 & \textbf{9.18e-4} & 23.0 & 34.1 & \textbf{1.47e-2} & 22.9 & 3.0 & \textbf{5.50e-5} & 2.6 & 32.0 & \textbf{4.07e-4} & 22.5 & 36.0 & \textbf{3.61e-5} & 21.7 & 17.3 & 5.36e-4 & 17.3 \\

\multicolumn{2}{@{}l}{\textbf{\textit{Improvement vs. SOTA}}} 
 & & \imp{95.4} & & & \imp{49.6} & & & \imp{58.1} & & & \imp{25.4} & & & \imp{82.7} & & & \imp{33.3} & & & \noimp & \\
\bottomrule
\end{tabular}
}
\end{table*}


\subsection{Heuristic Derivation Inspired by Inertial Manifold Theory}
Here provides a heuristic derivation of the reduced dynamical form.
The derivation follows the general philosophy of inertial manifold theory but does not assume the existence of a rigorous inertial manifold nor verify the associated spectral gap conditions.
Instead, the goal is to offer intuitive insight into how effective finite-dimensional dynamics of the form
\begin{align}\label{compact_dynamics_of_c}
\frac{\partial c(t)}{\partial t} \approx \mathcal{L}_c c+\mathcal{A}[\mathcal{N}_c(c)]=\mathscr{F}(c),
\end{align}
can arise from projecting infinite(high)-dimensional dynamics onto a reduced space.

Inertial manifold theory allows the reduction of complex, infinite-dimensional systems to finite-dimensional systems that capture the essential long-term dynamics while retaining necessary nonlinear terms, providing a promising avenue to tackle the above two challenges. 
An inertial manifold \( \mathcal{M} \) is a finite-dimensional, Lipschitz continuous manifold embedded within a Hilbert space $\mathcal{H}$, expressed as \( \mathcal{M} = \{\hat{h} + \Phi(\hat{h}) : \hat{h} \in \mathcal{H}_r = \mathscr{P}_m \mathcal{H}\} \), where \( \mathscr{P}_m: \mathcal{V}(D; \mathbb{R}^{d_v}) \to \mathcal{V}_r(D; \mathbb{R}^{d_m}) \) is a projection operator (typically spectral projection onto the first \( d_m \) eigenmodes in a projected space $\mathcal{V}_r=\mathscr{P}_m \mathcal{V}$), and $\Phi$ is a Lipschitz continuous function mapping from the low-mode component \( \hat{h} \) to its corresponding high-mode component in the orthogonal complement space \( \mathcal{H}_r^\perp \). 
Theoretically, the dimension $d_m$ is often inherently determined by the dynamics $\tilde{F}$, such as by identifying a sufficient gap in its spectrum. 
Even though the existence of inertial manifold for general dynamic systems still waits for being proved, there is a belief that they possess global attractors in their low-dimensional space. This belief stems from rigorous demonstrations in several PDEs, showing that the dynamics exponentially converge to a global attractor within a low-dimensional subset of the phase space \cite{foias1988inertial}, \cite{temam2012infinite}. 

Assuming the high-dimensional dynamics $\tilde{F}(v)=\mathcal{L}(v)+\mathcal{N}(v)$, where $\mathcal{L}:\mathcal{V}(D; \mathbb{R}^{d_v}) \to \mathcal{V}(D; \mathbb{R}^{d_v})$ represents a linear operator and $\mathcal{N}:\mathcal{V}(D; \mathbb{R}^{d_v}) \to \mathcal{V}(D; \mathbb{R}^{d_v})$ a nonlinear operator, and the existence of an inertial manifold $\mathcal{M}$ such that $v\in \mathcal{M}$. 
Then reduced dynamics of $\hat{v}=\mathscr{P}_mv$ on the projected space $\mathcal{V}_r$ are governed by the equation: 
\begin{align}
    \frac{\partial \hat{v}(t)}{\partial t} &= \mathscr{P}_m\frac{\partial v }{\partial t} \\
    &=\mathscr{P}_m\mathcal{L}(v)+\mathscr{P}_m\mathcal{N}(v)\\
    &=\mathcal{L}\hat{v}+\mathscr{P}_m\mathcal{N}[\hat{v}+\Phi(\hat{v})].
\end{align}
Thanks to the existence of graph mapping $\Phi$ inherent to the inertial manifold $\mathcal{M}$, the necessary nonlinear interactions influencing the finite-dimensional reduced state $\hat{v}$ can be implicitly captured by $\Phi(\hat{v})$ where $\hat{v}=\mathscr{P}_mv$ locates at the projected space $\mathcal{V}_r=\mathscr{P}_m \mathcal{V}$.
Since the nonlinear effort $\mathcal{N}(\hat{v}+\Phi(\hat{v}))$ comes from $\mathcal{N}[\hat{v}, \hat{v}]$, $\mathcal{N}[\hat{v}, \Phi(\hat{v})]$, $\mathcal{N}[\Phi(\hat{v}), \hat{v}]$ and $\mathcal{N}[\Phi(\hat{v}), \Phi(\hat{v})]$, one can speculate that $\mathcal{N}[\Phi(\hat{v}), \Phi(\hat{v})]$, $\mathcal{N}[\hat{v}, \Phi(\hat{v})]$, $\mathcal{N}[\Phi(\hat{v}), \hat{v}]$ are adequately smaller than $\mathcal{N}[\hat{v}, \hat{v}]=\mathcal{N}(\hat{v})$, such that assume
\begin{equation}
    \mathcal{N}[\hat{v}+\Phi(\hat{v})]=\mathcal{N}(\hat{v})+\mathcal{R}[\mathcal{N}(\hat{v})]
\end{equation} 
where $\mathcal{R}:\mathcal{V}(D; \mathbb{R}^{d_v}) \to \mathcal{V}(D; \mathbb{R}^{d_v})$ is a residual operator.
Then reduced dynamics of $\hat{v}$ are governed by the equation: 
\begin{align}\label{dynamics_only_associated_with_reduced_state}
 \frac{\partial \hat{v}(t)}{\partial t}
=\mathcal{L}\hat{v}+\mathscr{P}_m\mathcal{N}(\hat{v})+\mathscr{P}_m\mathcal{R}[\mathcal{N}(\hat{v})].
\end{align}
and $\mathcal{A}$ the operator mapping the low-mode component to the full-mode component.

Although the reduced dynamics is built, it still evolves in the $d_v$-dimensional space and is difficult to perform calculation in practice. By expanding the reduced state $\hat{v}$ onto the spectral space, it has:
\begin{equation}
    \hat{v} = \sum^{d_m}_{i=1} c_iw_i,
\end{equation}
where $w_i \in \mathbb{R}^{d_v}$ is one spectral basis of the first $d_m$ eigenmodes and $c_i \in \mathbb{R}$ is its corresponding spectral coefficient. Substituting this expansion into Eq. \ref{dynamics_only_associated_with_reduced_state}:
\begin{equation}
    \sum_{i=1}^{d_m} \frac{\partial c_i}{\partial t} w_i=\mathcal{L}\left(\sum_{i=1}^{d_m} c_i w_i\right)+\mathscr{P}_m \mathcal{N}\left(\sum_{j=1}^{d_m} c_j w_j\right)+\mathscr{P}_m \mathcal{R}\Bigg[\mathcal{N}\left(\sum_{j=1}^m c_j w_j\right)\Bigg]
\end{equation}
Using the linearity of $\mathcal{L}$, the orthonormality $\left\langle w_i, w_k\right\rangle=\delta_{i k}$ and the inner product with $w_k$ for $k=1,…,d_m$:
\begin{equation}
    \frac{\partial c_k}{\partial t}=\sum_{i=1}^{d_m} c_i\left\langle\mathcal{L} w_i, w_k\right\rangle+\left\langle\mathcal{N}\left(\sum_{j=1}^{d_m} c_j w_j\right), \mathscr{P}_m w_k\right\rangle+\left\langle\mathcal{R}\Bigg[\mathcal{N}(\sum_{j=1}^m c_j w_j)\Bigg], \mathscr{P}_m w_k\right\rangle
\end{equation}
Since $w_k$ is in the range of $\mathscr{P}_m$, $\mathscr{P}_mw_k=w_k$. Finally, we can give the dynamics of coefficient $c \in \mathcal{C}(D; \mathbb{R}^{d_m})$, where $\mathcal{C}(D; \mathbb{R}^{d_m})$ is a Banach space, in a compact vector form:
\begin{align}\label{dynamics_of_c}
 \frac{\partial c(t)}{\partial t}
 =\mathcal{L}_c c+\mathcal{N}_c(c)+\mathcal{R}_c[\mathcal{N}(c)],
\end{align}
where $\mathcal{L}_c:\mathcal{C}(D; \mathbb{R}^{d_m}) \to \mathcal{C}(D; \mathbb{R}^{d_m})$ is a linear operator for $c$ with its $k$-th component $[\mathcal{L}_c]_k= \langle\mathcal{L} w_m, w_k\rangle$,  $\mathcal{N}_c:\mathcal{C}(D; \mathbb{R}^{d_m}) \to \mathcal{C}(D; \mathbb{R}^{d_m})$ is a nonlinear operator for $c$ with its $k$-th component $[\mathcal{N}_c]_k=\left\langle\mathcal{N}\left(\sum_{j=1}^m c_j w_j\right), \mathscr{P}_m w_k\right\rangle$, and $\mathcal{R}_c:\mathcal{C}(D; \mathbb{R}^{d_m}) \to \mathcal{C}(D; \mathbb{R}^{d_m})$ is a nonlinear operator for $c$ with its $k$-th component $[\mathcal{R}_c]_k=\left\langle\mathcal{R}[\mathcal{N}(\sum_{j=1}^m c_j w_j)], \mathscr{P}_m w_k\right\rangle$.
Assume $\mathcal{R}_c[\mathcal{N}_c(c)] \approx \mathcal{R}_c[\mathcal{N}(c)]$ and $\mathcal{A}[\mathcal{N}_c(c)]=\mathcal{N}_c(c)+\mathcal{R}_c[\mathcal{N}_c(c)]$, Eq. \ref{dynamics_of_c} becomes a more compact form:
\begin{align}\label{compact_dynamics_of_c}
 \frac{\partial c(t)}{\partial t}
\approx \mathcal{L}_c c+\mathcal{A}[\mathcal{N}_c(c)]=\mathscr{F}(c),
\end{align}
where $\mathcal{A}:\mathcal{C}(D; \mathbb{R}^{d_m}) \to \mathcal{C}(D; \mathbb{R}^{d_m})$ is an operator mapping the low-mode component to the full-mode component, $\mathscr{F}:\mathcal{C}(D; \mathbb{R}^{d_m}) \to \mathcal{C}(D; \mathbb{R}^{d_m})$ is a nonlinear operator that acts on $c$ and represents the dynamics of $c$.

\subsection{Dynamics-informed architecture}
We need to assume that the transformation $\mathscr{T}$ induces a linear time transformation $t_v=g(t_u)$ and the projection $\mathscr{P}_m$ maintains the time scale $t_v=t_c$. It indicates a relationship between states $\mathscr{P}_m\mathscr{T}(u)(t_u) = \mathscr{P}_mv(t_v)=c\big(t_c\big)$, and their dynamics: 
\begin{align}\label{new_relationship_infinite_u_and_v}
    \frac{\partial \mathscr{P}_m\mathscr{T}\big(u\big)(t_u)}{\partial t_u} 
    = \frac{\partial \mathscr{P}_mv\big(g(t_u)\big)}{\partial g(t_u)} \frac{\partial g(t_u)}{\partial t_u}=\zeta\frac{\partial c(t_c)}{\partial t_c}=\zeta \mathscr{F},
\end{align}
where $\zeta=\frac{\partial g(t_u)}{\partial t_c}$ is the scaling factor, transforming the time scale $t_u$ to $t_c$.
When dynamics are represented by neural layers, let we assume the scaling factor is the number of stacked LGM layers, $\zeta=L_d$, and Eq. \ref{new_relationship_infinite_u_and_v} can be parameterized and expanded as
\begin{align}\label{expanded_form_of_c_dynamics}
    \zeta  \mathscr{F}_\theta(c) =  \mathscr{F}_{\theta_1} \big(c\big) + \cdots +  \mathscr{F}_{\theta_{L_d}} \big(c \big),
\end{align}
where $ \mathscr{F}_{\theta_l}, l=1, ..., L_d$ is parameterized by neural layer $l$. 
Assuming an initial reduced latent state $c_0$ and an evolutionary step $\Delta t_{c}=\Delta t_u / L_d$ to the target reduced latent state $c_{L_d}$, the form of Eq. \ref{expanded_form_of_c_dynamics} suggest two alternatives to constitute the network architecture: a parallel way, stacking all layers which consider the same input and is formulated as
\begin{equation}
        c_{L_d}(x)=c_{0}(x)+\Delta t_{c}\sum^{L_d}_{l=1}\mathscr{F}_{\theta_{l}} (c_0),
\end{equation}
or a hierarchical way, connecting all layers in order, each of which takes the output from the previous one as the input and is formulated as
\begin{equation}
     c_{l}(x) = c_{l-1}(x)+ \Delta t_{c} \mathscr{F}_{\theta_{l}} (c_{l-1}), \quad l=1, ..., L_d.
\end{equation}
In the parallel way, it facilitates easier gradient propagation to the preceding neural layers, mitigating the issues of gradient vanishing or exploding while treating all parameterized dynamics with similar importance. 
In the hierarchical way, it builds finer dynamics by leveraging the previous reduced latent state $c_{l-1}$. This process progressively refines the model’s accuracy, as each successive layer adjusts the representation established by the prior dynamics and finally form the successive trajectory from the initial state to the terminal state. 

Alternatively, taking full advantages of two ways, a combined connection version is naturally proposed. 
In this version, we break the limitation that each reduced latent dynamics evolves with a fixed step $\Delta t_{c}$, and parameterize the evolutionary step $\Delta t_{c_{\theta_l}}$, providing the flexibility to represent the reduced latent dynamics of $c$. Consequently, we can get the following formula of the output:
\begin{align}
    c_{L_d}
    =c_{0}+ \sum_{l=1}^{L_d} \Delta t_{c_{\theta_l}} \mathscr{F}_{\theta_{l}} (c_{l-1} ) 
\end{align}
where the input to each layer $c_{l-1}$ is recursively defined as
\begin{equation}
        c_{l-1} = c_{l-2}+ \mathscr{F}_{\theta_{l-1}} (c_{l-2}), l=2, ..., L_d.
\end{equation}
Notably, $\{c_{l}\}_{l=1,...,L_d-1}$ are computed using a default evolutionary step $\Delta t_{v}=1$. When it comes to the calculation of the final output $c_{L_d}$, the parameterized evolutionary step  $\Delta t_{c_{\theta_l}}$ is incorporated.

\newpage
\subsection{Visualization of baselines across benchmarks}
\subsubsection{1D Kuramoto-Sivashinsky}
\begin{figure}[htbp]
    \centering
    \includegraphics[width=0.98\linewidth]{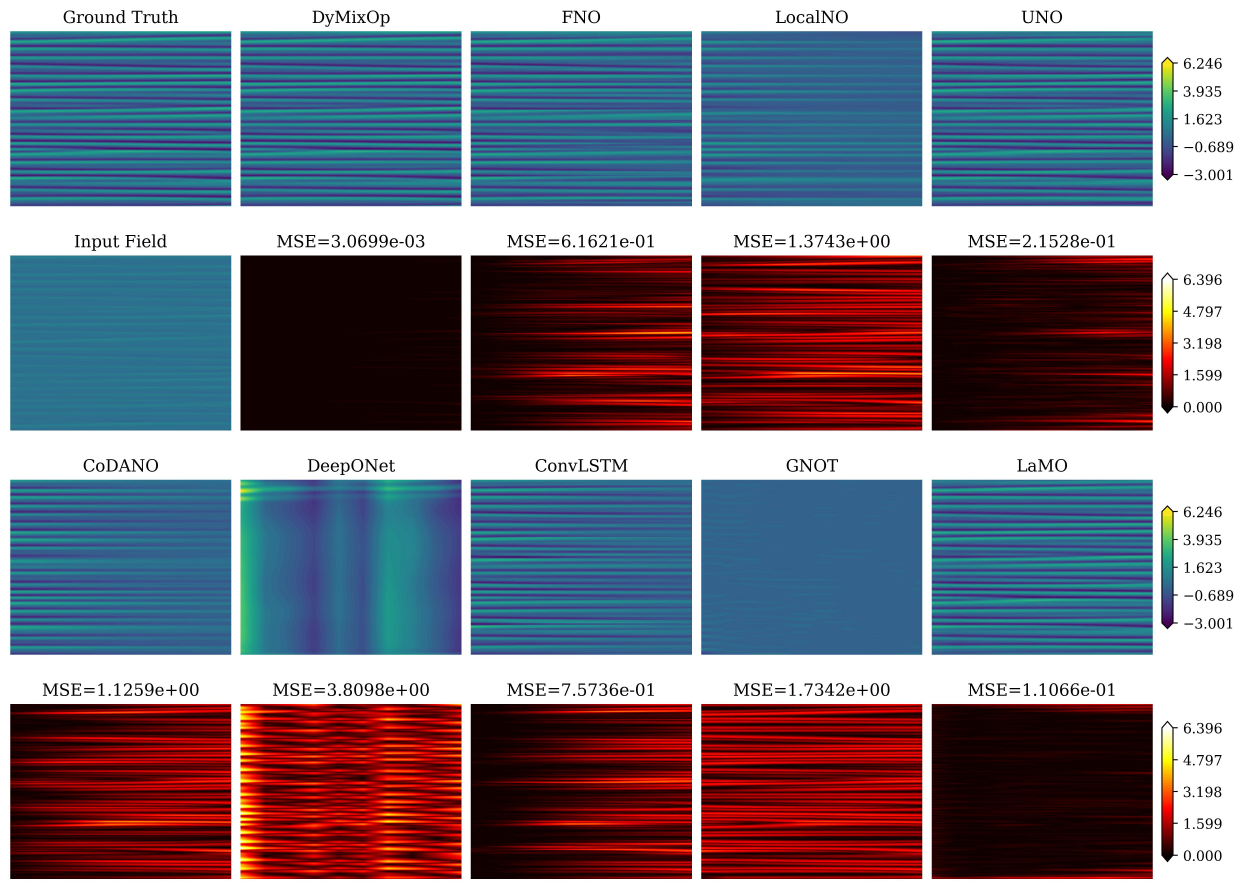}
    \caption{Visual comparison on 1D Kuramoto-Sivashinsky for the first sample. Each panel displays the evolution of the field $u(x,t)$ over space (horizontal axis) and time (vertical axis). }
    \label{fig:1dks_batch0}
\end{figure}

\begin{figure}[htbp]
    \centering
    \includegraphics[width=0.98\linewidth]{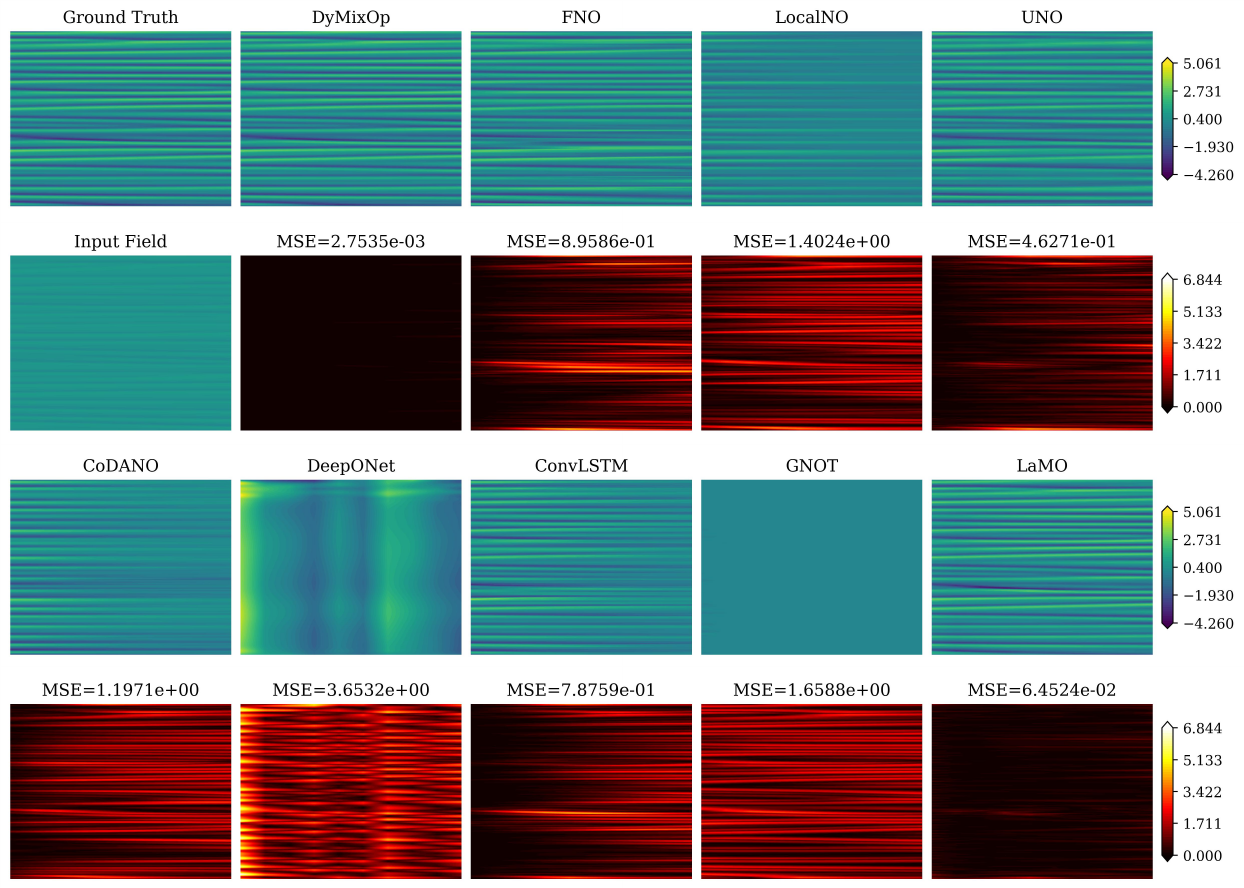}
    \caption{Visual comparison on 1D Kuramoto-Sivashinsky for the last sample. Each panel displays the evolution of the field $u(x,t)$ over space (horizontal axis) and time (vertical axis). }
    \label{fig:1dks_batch1}
\end{figure}

The visualizations of all baselines and DyMixOp for the spatiotemporal dynamics of the first and last samples in the test dataset are shown in Fig.~\ref{fig:1dks_batch0} and Fig.~\ref{fig:1dks_batch1}. Each panel displays the evolution of the field $u(t, x)$ over space (vertical axis) and time (horizontal axis), with color encoding the scalar value. The ground-truth solutions exhibit characteristic chaotic structures: coherent, localized wave-like disturbances propagating obliquely across the domain, interspersed with intermittent bursts of high-amplitude activity—consistent with the known spatiotemporal intermittency of the 1D Kuramoto–Sivashinsky (KS) equation.
The KS equation governs the weakly nonlinear evolution of a flame front or thin film interface, where the competition between destabilizing second-order diffusion ($-u_{xx}$) and stabilizing fourth-order dispersion ($-u_{xxxx}$), together with the nonlinear advection term ($-uu_x$), yields spatiotemporal chaos. The observed diagonal striations in the ground truth correspond to traveling wave packets whose speed is determined by the balance of these terms; their irregular spacing and occasional merging/splitting reflect the underlying deterministic chaos.

Baselines such as GNOT and DeepONet produce overly smooth, spatially homogeneous fields (near-uniform teal), indicating severe under-resolution of the high-wavenumber structures and loss of dynamical intermittency—suggesting insufficient capacity to capture the nonlinear cascade.
LocalNO and CoDANO retain some stripe-like structure but exhibit phase drift and amplitude attenuation over time, manifested as misaligned or faded diagonal features in the error plots (red intensity indicates local MSE). This points to cumulative phase errors in the temporal evolution.
FNO, UNO, and ConvLSTM show improved fidelity in preserving spatial coherence but still suffer from spurious high-frequency artifacts (localized red streaks in error maps), implying instability in resolving sharp gradients near wavefronts.
LaMO achieves relatively low global MSE (e.g., $6.45\times10^{-2}$ in batch 0) and maintains visible diagonal structures, though subtle phase deviations persist, particularly in later time steps.
DyMixOp consistently yields the most visually faithful reconstructions across both samples: the predicted fields preserve not only the orientation and spacing of the dominant wave packets but also the intermittent high-amplitude events (e.g., bright yellow/red spots near $t\approx0.6$ in Fig.~\ref{fig:1dks_batch0}), with the smallest error magnitude (MSE $\sim 10^{-3}$) and minimal structured residuals in the error maps (nearly uniform dark regions).

Notably, the error patterns reveal that failure modes are often structured. For instance, FNO and UNO exhibit periodic vertical bands of elevated error, suggesting aliasing or mode truncation at specific wavenumbers; whereas DyMixOp’s error is largely uncorrelated and low-amplitude, indicative of stochastic-like approximation noise rather than systematic dynamical bias. This suggests DyMixOp better respects the underlying symmetries and conservation properties (e.g., Galilean invariance, energy cascade scaling) inherent to the KS dynamics, without imposing excessive smoothing or artificial damping.

\subsubsection{2D Darcy}
\begin{figure}[htbp]
    \centering
    \includegraphics[width=0.98\linewidth]{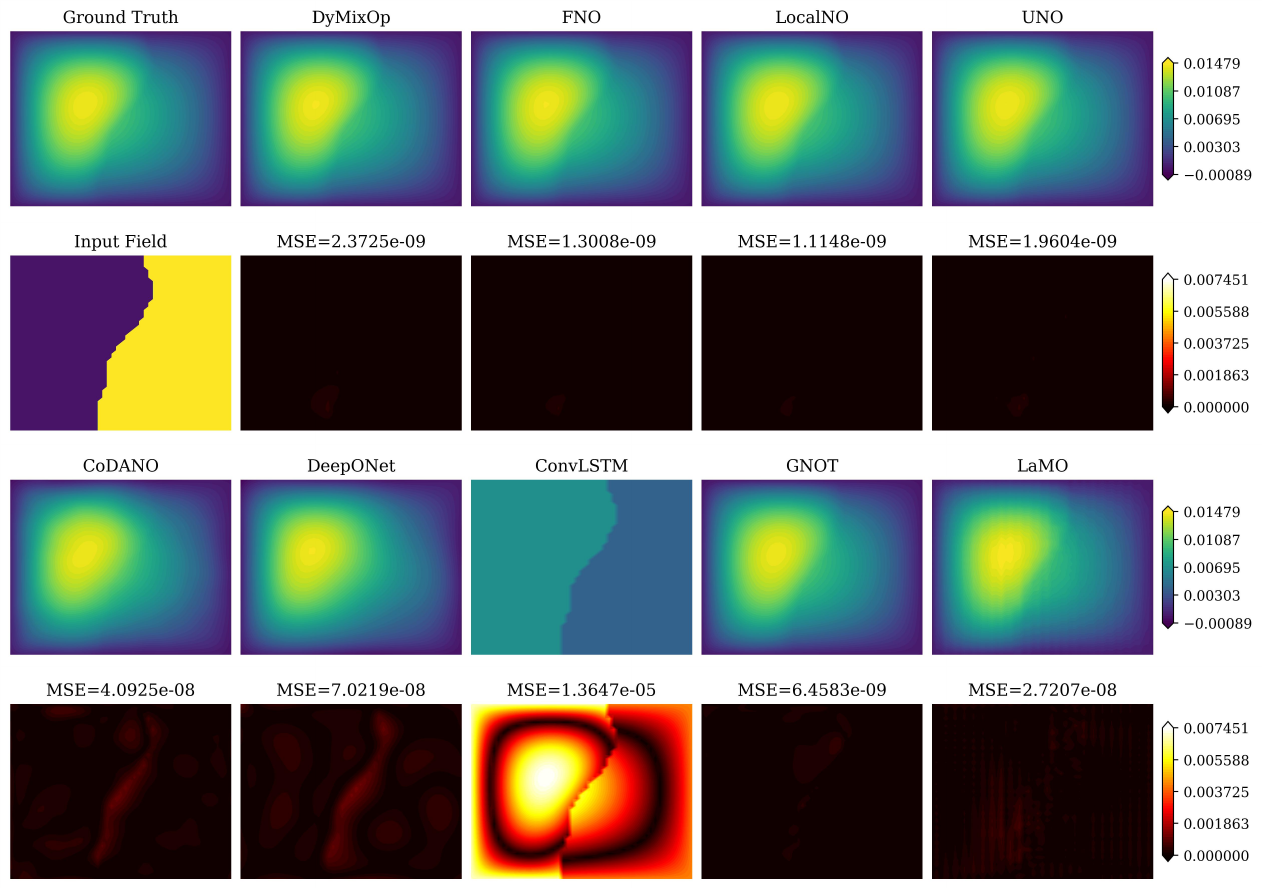}
    \caption{Visual comparison on 2D Darcy for the first sample. Each panel displays a 2D scalar field (e.g., pressure or concentration) over a square domain}
    \label{fig:2ddarcy_batch0}
\end{figure}

\begin{figure}[htbp]
    \centering
    \includegraphics[width=0.98\linewidth]{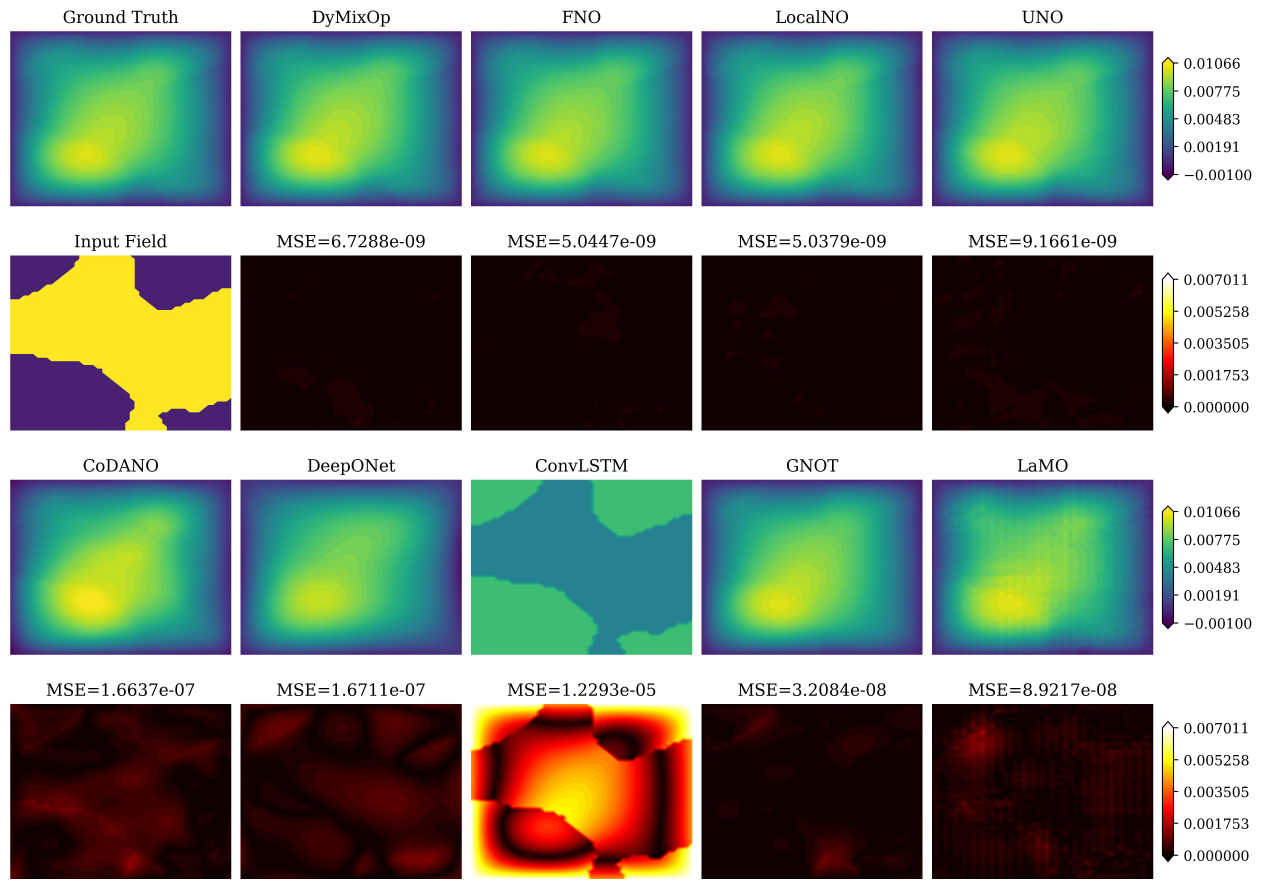}
    \caption{Visual comparison on 2D Darcy for the last sample. Each panel displays a 2D scalar field (e.g., pressure or concentration) over a square domain}
    \label{fig:2ddarcy_batch1}
\end{figure}

The visualizations of all baselines and DyMixOp for the states of the first and last samples in the test dataset are shown in Fig.~\ref{fig:2ddarcy_batch0} and Fig.~\ref{fig:2ddarcy_batch1}, respectively. Each row corresponds to a distinct model, with the top row displaying the predicted solution fields and the bottom row presenting the absolute error distributions (i.e., $|u_{\text{pred}} - u_{\text{true}}|$), where the color intensity reflects the magnitude of local error.
The ground-truth solutions exhibit smooth, elliptic-like concentration profiles—characteristic of steady-state Darcy flow under heterogeneous permeability—where high values (yellow–green) correspond to regions of elevated pressure or hydraulic head, and low values (blue–purple) indicate low-pressure zones. In this mapping, the input data (binary-valued permeability fields) define sharp material interfaces; the solution must satisfy continuity of flux across these discontinuities, leading to nontrivial gradients near the boundaries.

FNO, LocalNO, UNO, and DyMixOp produce qualitatively smooth output fields that closely match the shape, orientation, and centroid of the ground truth. In contrast, DeepONet and GNOT exhibit localized distortions: e.g., in Fig.~\ref{fig:2ddarcy_batch0}, DeepONet’s prediction shows an elongated, asymmetric peak, while GNOT introduces spurious secondary lobes near the interface region. ConvLSTM consistently fails to resolve the interior gradient structure, producing a blocky, piecewise-constant field aligned with the input geometry rather than the physics-driven solution.
The error maps reveal that models with poor spatial generalization (e.g., ConvLSTM, DeepONet, GNOT, CoDANO) concentrate errors along the permeability discontinuity and in regions of high curvature of the solution (e.g., near the “neck” of the high-permeability inclusion). This suggests insufficient capacity to enforce the jump conditions inherent in Darcy’s law across heterogeneous media. Conversely, DyMixOp, FNO, and LocalNO show minimal error in those critical regions, with residual errors confined to the far-field low-gradient zones—indicating better enforcement of global conservation and local continuity.
Comparing Fig.~\ref{fig:2ddarcy_batch0} (simpler, single-inclusion geometry) and Fig.~\ref{fig:2ddarcy_batch1} (more complex, multi-component inclusion), we observe that error magnitudes increase for most models, but not uniformly. For instance, UNO’s MSE rises from $1.96\times10^{-9}$ to $9.17\times10^{-9}$ ($\approx$4.7×), whereas DyMixOp’s increases only from $2.37\times10^{-9}$ to $6.73\times10^{-9}$ ($\approx$2.8×), suggesting relatively better robustness to geometric complexity. Notably, ConvLSTM’s error remains large ($\sim 10^{-5}$) in both cases, indicating structural limitations rather than overfitting to specific geometries.
The error patterns of DyMixOp, FNO, and LocalNO are predominantly diffuse and low-amplitude, lacking sharp edges or oscillatory artifacts—consistent with well-posed elliptic PDE solvers. In contrast, DeepONet and CoDANO display filamentary or patchy high-error structures that correlate with regions of strong solution curvature, hinting at inadequate resolution of second-order derivatives (i.e., Laplacian terms) in the learned operator.

These observations collectively suggest that superior performance correlates with the model’s ability to respect the underlying elliptic regularity and interface conditions of the Darcy problem—not merely with expressive capacity, but with architectural alignment to the physics-constrained solution manifold.

\subsubsection{2D Burgers}
\begin{figure}[htbp]
    \centering
    \includegraphics[width=0.98\linewidth]{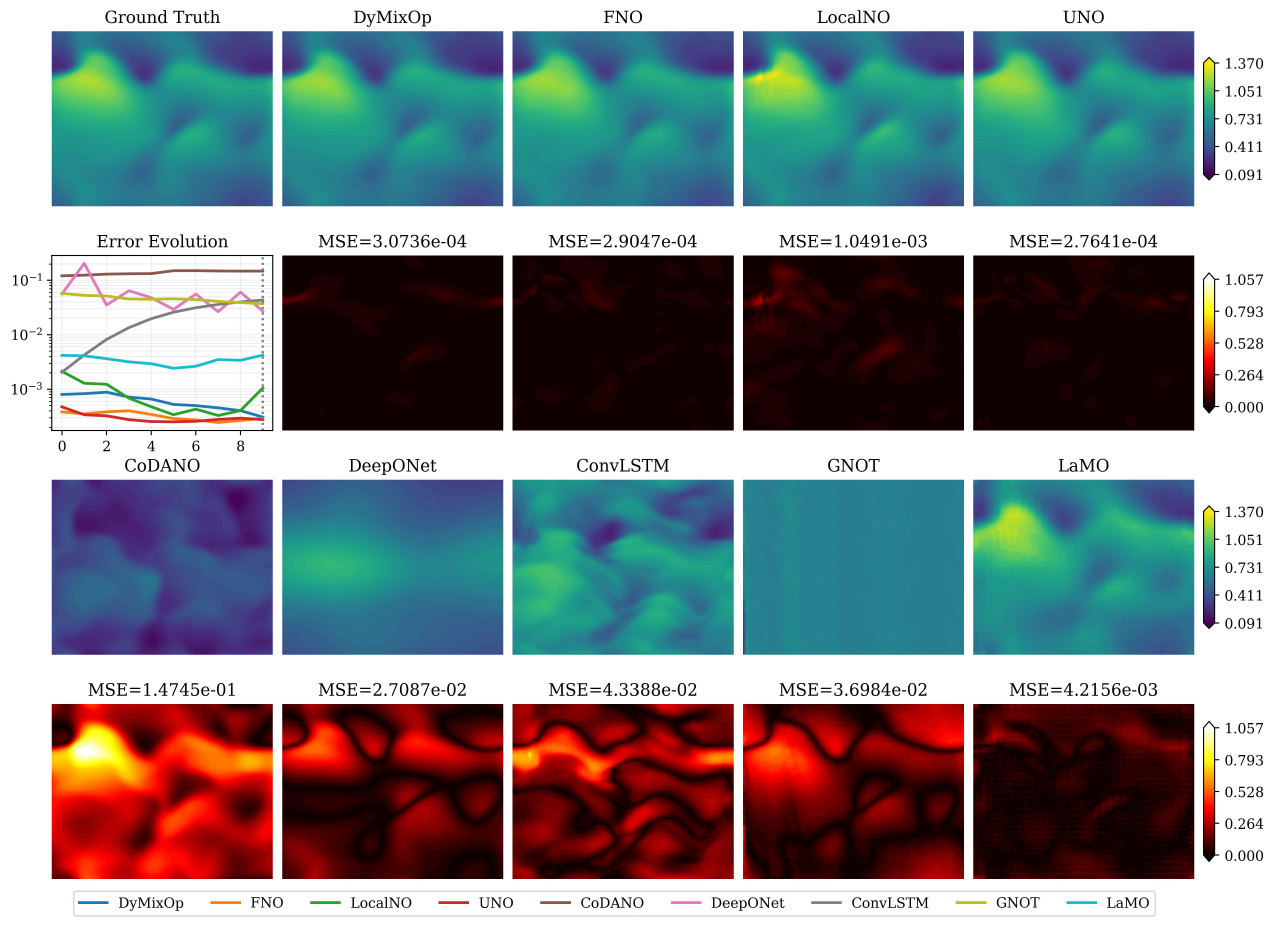}
    \caption{Visual comparison on 2D Burgers for the first sample at the final step. The variable $u$ is presented here.}
    \label{fig:2dbrugers_batch0}
\end{figure}

\begin{figure}[htbp]
    \centering
    \includegraphics[width=0.98\linewidth]{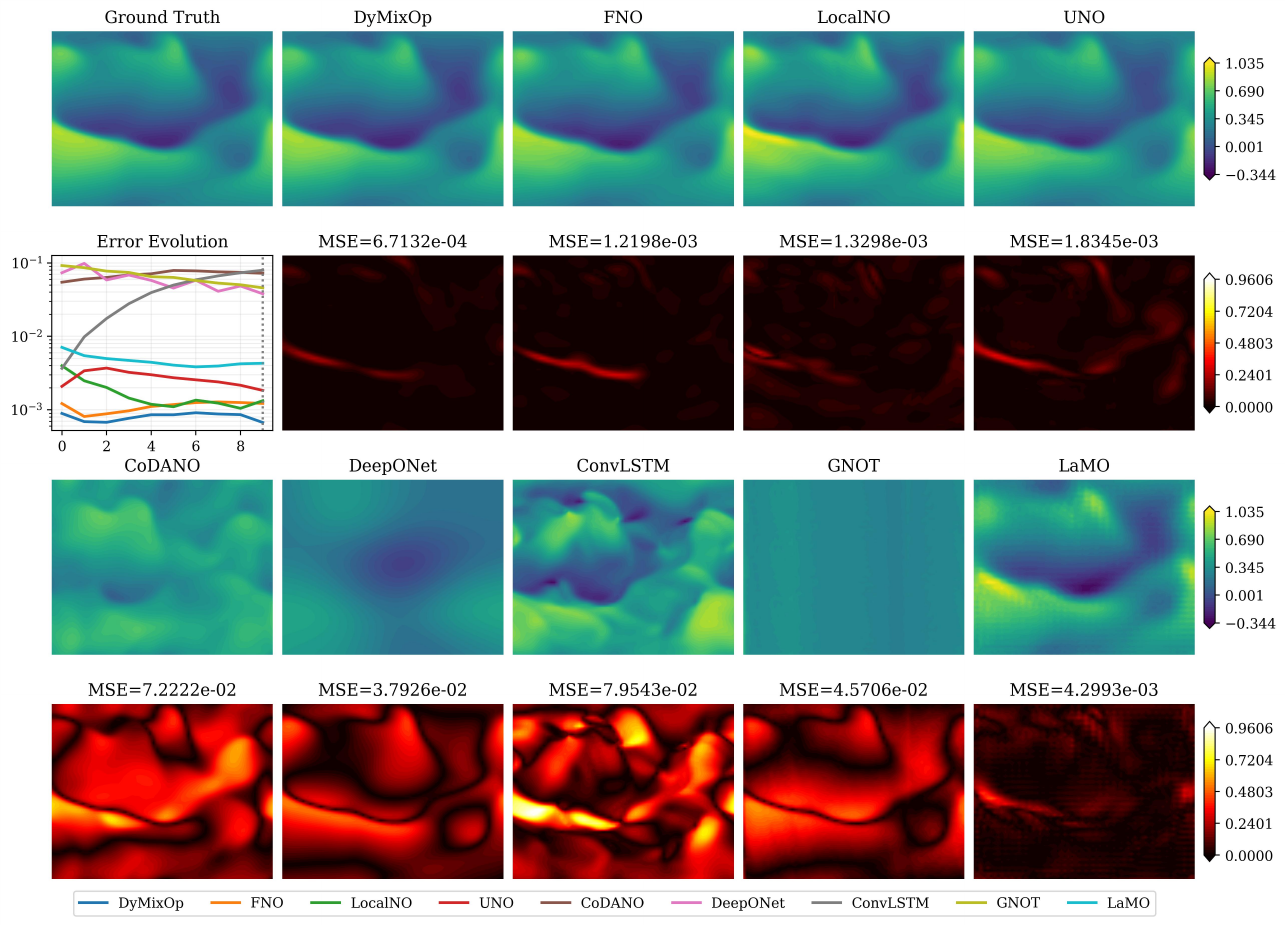}
    \caption{Visual comparison on 2D Burgers for the last sample at the final step. The variable $u$ is presented here.}
    \label{fig:2dburgers_batch1}
\end{figure}

The visualizations of all baselines and DyMixOp for the scalar field $u$ at the final time step of the first and last samples in the test dataset are shown in Fig.~\ref{fig:2dbrugers_batch0} and Fig.~\ref{fig:2dburgers_batch1}, respectively. Each panel displays (top row) the predicted solution field (the ground truth) and (bottom row) the absolute error distribution $|u_{\text{pred}} - u_{\text{true}}|$ (or the error evolution), with color intensity indicating magnitude. 
Recalling that the 2D Burgers equation $\partial_t u + u \nabla u = \nu \Delta u$ governs advective transport with possible shock formation. The ground-truth solutions exhibit coherent, elongated structures resulting from the stretching and folding of initial vorticity-like features under self-advection. Sharp gradients develop along curved fronts, but remain regularized by viscosity ($\nu > 0$ in the dataset), preventing true discontinuities while preserving high local gradients.
In this mapping, the input data (initial condition $u_0(\mathbf{x})$) is a smooth, multi-modal field with localized high-amplitude regions embedded in a lower-background field, characteristic of the inviscid or weakly viscous 2D Burgers equation where nonlinear advection dominates.

FNO, LocalNO, and DyMixOp produce output fields that retain the spatial topology of the ground truth: the number, orientation, and relative amplitude of dominant lobes are well-preserved. In contrast, DeepONet and GNOT yield overly smoothed fields (e.g., Fig.~\ref{fig:2dbrugers_batch0}, DeepONet: near-uniform teal; GNOT: featureless gradient), indicating excessive dissipation or insufficient resolution of advective nonlinearity. ConvLSTM captures some structure but introduces spurious oscillatory bands (e.g., horizontal striations in Fig.~\ref{fig:2dbrugers_batch1}), suggesting temporal instability or inadequate handling of spatial coupling.
The error maps reveal that models failing to resolve sharp fronts (DeepONet, GNOT, CoDANO) exhibit large errors precisely along the high-gradient interfaces of the solution—regions where $\|\nabla u\|$ is maximal. For instance, in Fig.~\ref{fig:2dbrugers_batch0}, CoDANO’s error map shows intense red bands coinciding with the curved shear layers in the ground truth, consistent with inability to model the nonlinear term $u \nabla u$ accurately. Conversely, DyMixOp, LocalNO, and FNO show minimal error along these fronts, with residuals confined to low-gradient background regions—indicating better preservation of the advective transport dynamics.

The ground truth contains distinct high-amplitude peaks (yellow) surrounded by steep decay into background levels. DyMixOp and LocalNO reproduce these peak magnitudes faithfully (visually comparable saturation), whereas UNO and LaMO systematically underestimate peak amplitudes (e.g., Fig.~\ref{fig:2dbrugers_batch1}, UNO: peaks appear green rather than yellow), suggesting energy dissipation beyond physical viscosity. Notably, ConvLSTM overestimates peak values in localized zones (e.g., bright yellow spots in error map of Fig.~\ref{fig:2dbrugers_batch0}).

Comparing the two test samples, the qualitative failure modes are consistent across models: DeepONet and GNOT remain uniformly blurred; CoDANO consistently generates fragmented, disconnected high-error patches; DyMixOp maintains structural coherence in both cases, though with slightly increased error magnitude in the more complex second sample (MSE: $3.07e-4$ vs. $6.71e-4$). This suggests DyMixOp’s operator representation is less sensitive to variations in initial spectral content than alternatives reliant on fixed-grid convolutions or pointwise MLPs.

In summary, the figures indicate that accurate modeling of the 2D Burgers dynamics requires not only global pattern matching but precise resolution of local gradient alignment and nonlinear advection-induced deformation. Models whose error distributions concentrate along evolving fronts—and whose predictions deviate in peak amplitude or topological connectivity—are physically inconsistent with the expected behavior of a weakly viscous advective system.

\subsubsection{2D CE-CRP}
\begin{figure}[htbp]
    \centering
    \includegraphics[width=0.98\linewidth]{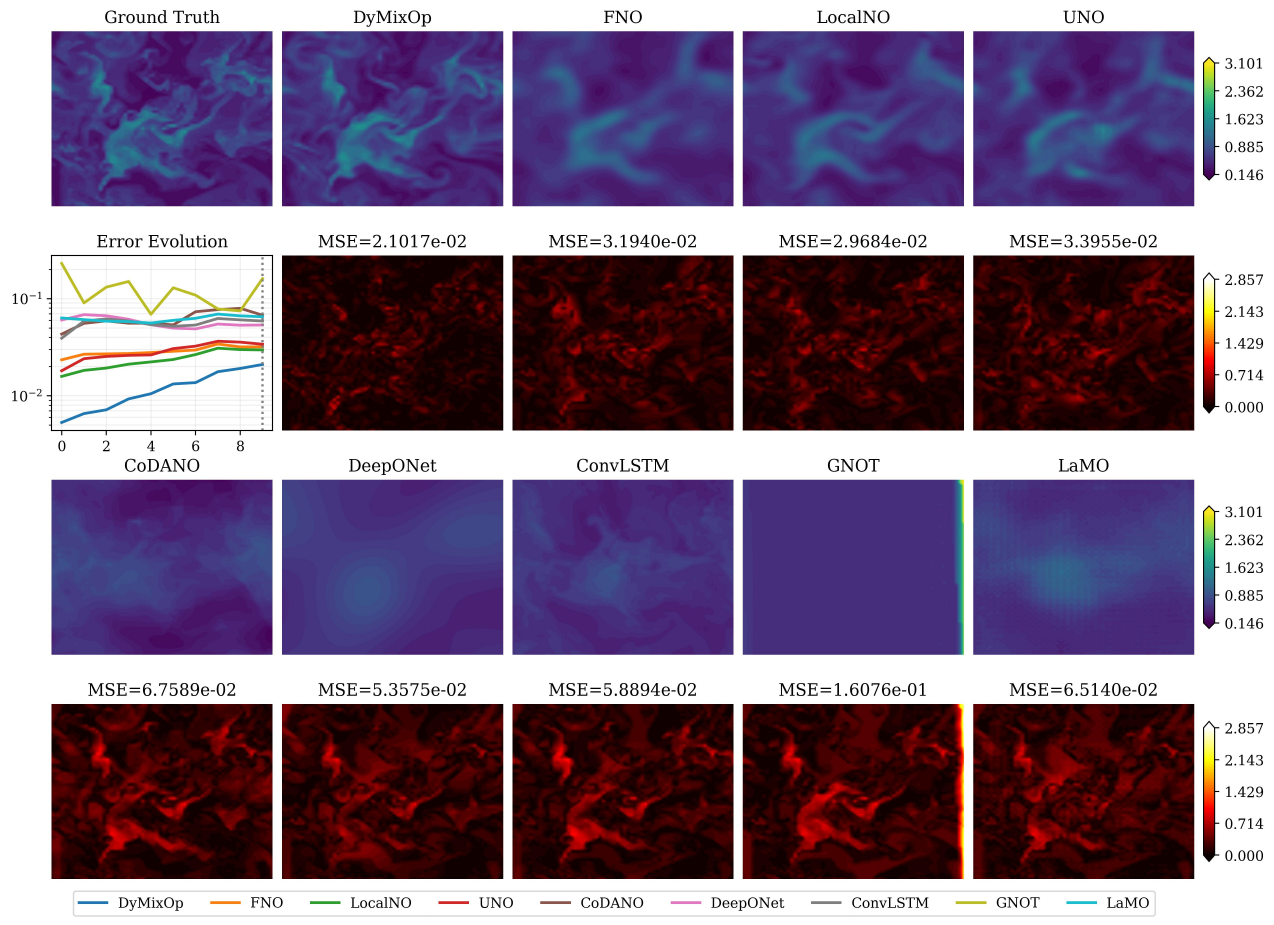}
    \caption{Visual comparison on 2D CE-CRP for the first sample at the final step. The velocity variable $u$ is presented here.}
    \label{fig:2dce-crp_batch0}
\end{figure}

\begin{figure}[htbp]
    \centering
    \includegraphics[width=0.98\linewidth]{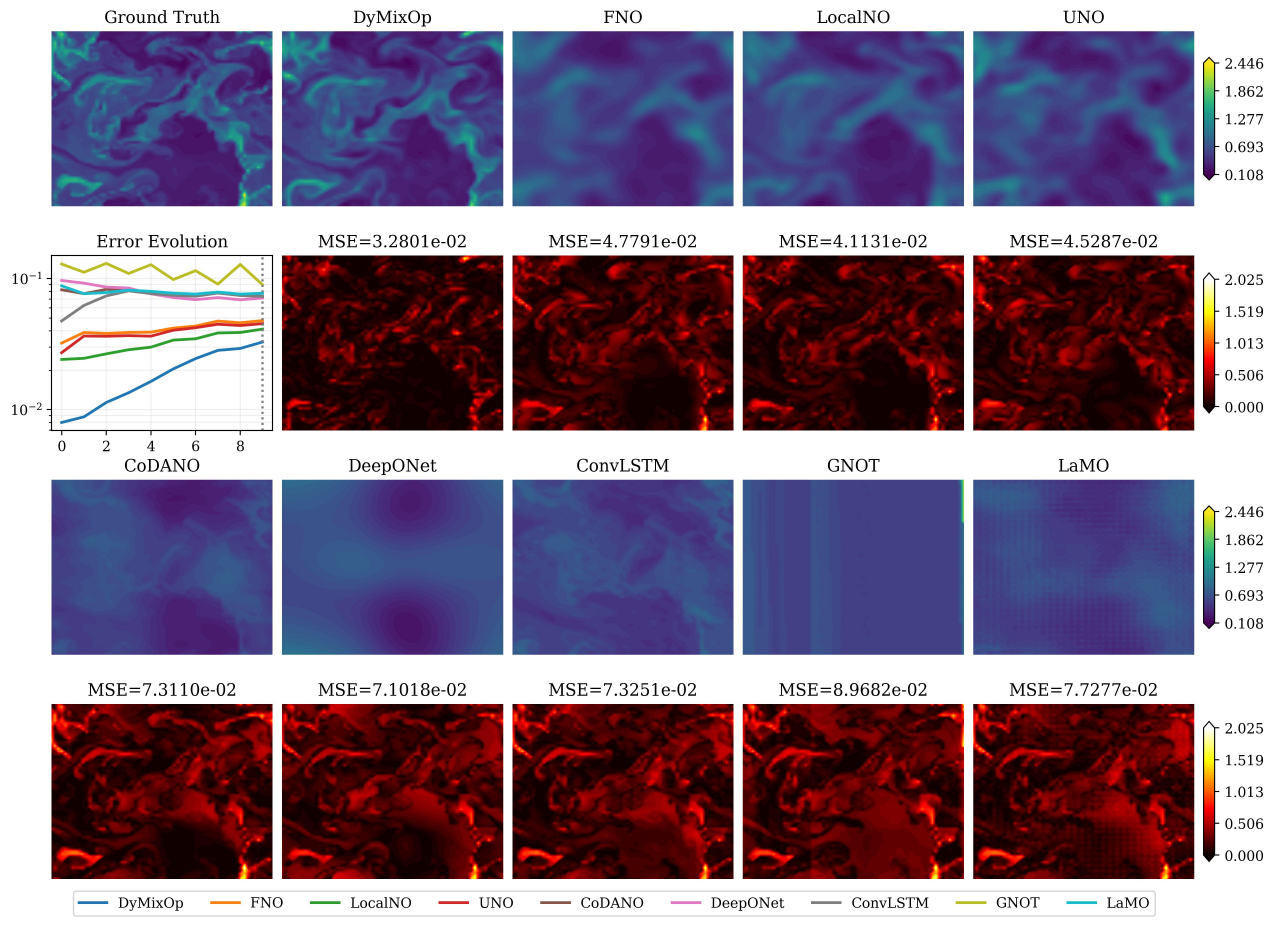}
    \caption{Visual comparison on 2D CE-CRP for the last sample at the final step. The velocity variable $u$ is presented here.}
    \label{fig:2dce-crp_batch1}
\end{figure}

The visualizations of all baselines and DyMixOp for the horizontal velocity component $u$ at the final prediction step of the first and last samples in the test dataset are shown in Fig.~\ref{fig:2dce-crp_batch0} and Fig.~\ref{fig:2dce-crp_batch1}, respectively. Each panel displays (top row) the predicted solution field (the ground truth) and (bottom row) the absolute error distribution $|u_{\text{pred}} - u_{\text{true}}|$ (or the error evolution), with color intensity indicating magnitude. 
The 2D CE-CRP problem models inviscid compressible flow governed by the Euler equations, which admit discontinuous solutions—including shocks, contact discontinuities, and shear layers—due to the absence of viscosity. The initial condition consists of four curved subdomains with piecewise-constant states.
Upon evolution, these interfaces interact nonlinearly, generating complex wave patterns: rarefaction fans, oblique shocks, and vortical roll-up structures visible as fine-scale filamentary features in the ground truth (e.g., Fig.~\ref{fig:2dce-crp_batch0}, ground truth: intricate green–teal filaments embedded in purple background). The presence of sharp gradients and multi-scale coherent structures reflects the hyperbolic nature of the system and the sensitivity to initial Riemann data.

The ground truth exhibits thin, high-contrast transition layers (e.g., sharp green–blue boundaries) corresponding to shock or contact surfaces. DyMixOp, FNO, and LocalNO reproduce these layers with reasonable fidelity: their predictions retain the spatial location, curvature, and connectivity of major filaments. In contrast, DeepONet, GNOT, and ConvLSTM produce overly diffused fields—e.g., DeepONet (Fig.~\ref{fig:2dce-crp_batch0}) shows only broad, low-amplitude undulations without localized fronts; GNOT yields near-uniform fields (dominant purple) with minimal structure; ConvLSTM preserves some topology but similar with the input data. This suggests inadequate resolution of jump conditions and characteristic propagation.
The error maps confirm that large errors (bright red regions) concentrate precisely along the high-gradient interfaces present in the ground truth—not in smooth interior regions. For instance, in Fig.~\ref{fig:2dce-crp_batch0}, CoDANO and UNO show intense red streaks aligned with the curved filamentary structures, indicating failure to capture the correct wave speed. DyMixOp’s error map, while non-zero, exhibits significantly weaker and more localized red patches, often confined to secondary small-scale features rather than primary shock fronts—suggesting better adherence to the underlying hyperbolic transport physics.
The ground truth contains both large-scale coherent structures (e.g., dominant green lobes) and fine-scale turbulent-like filaments. Baselines such as LaMO and CoDANO suppress the fine-scale features entirely (LaMO: nearly uniform field; CoDANO: coarse, blocky patterns), implying excessive numerical dissipation or insufficient representational capacity for high-wavenumber content. Conversely, DyMixOp and LocalNO retain both scales: the former preserves the global morphology and injects plausible small-scale detail, while the latter maintains sharper edges and higher contrast in filament regions—consistent with operator-based architectures that respect the spectral structure of hyperbolic evolution.
Comparing the two samples, DyMixOp consistently resolves the dominant curved interfaces, though with slight variation in filament density. 

In summary, the visual evidence indicates that successful modeling of the 2D CE-CRP requires architectures capable of resolving discontinuity propagation and multi-scale wave interaction without artificial diffusion or dispersion. The superior performance of DyMixOp—evidenced by structurally faithful predictions and error localized away from primary wave fronts—is consistent with its design for mixed global-local operator learning, which aligns with the characteristic-based solution mechanism of hyperbolic systems.

\subsubsection{2D Navier-Stokes}
\begin{figure}[htbp]
    \centering
    \includegraphics[width=0.98\linewidth]{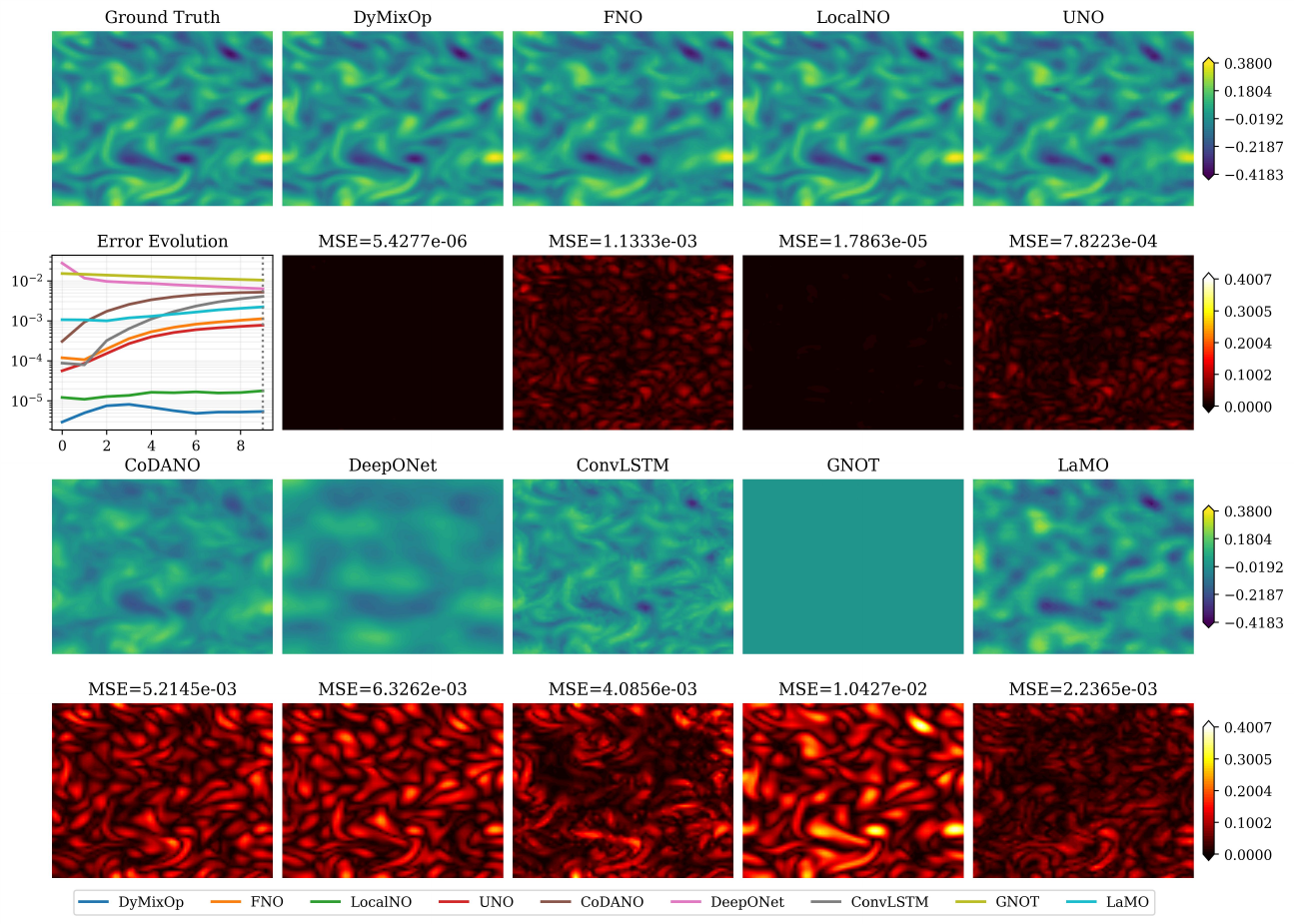}
    \caption{Visual comparison on 2D Navier-Stokes for the first sample at the final step. The vorticity variable $\omega$ is presented here.}
    \label{fig:2dns_batch0}
\end{figure}

\begin{figure}[htbp]
    \centering
    \includegraphics[width=0.98\linewidth]{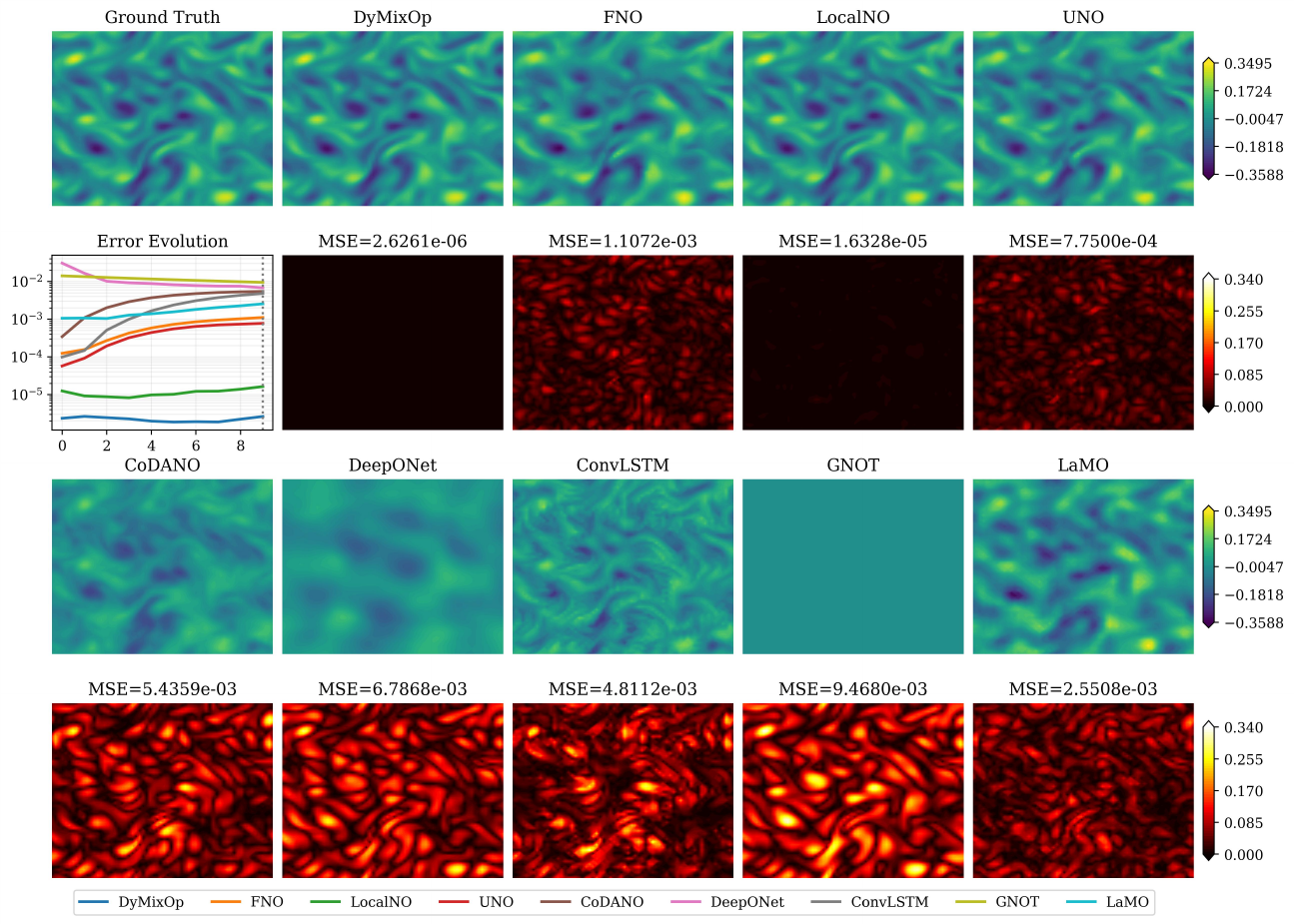}
    \caption{Visual comparison on 2D Navier-Stokes for the last sample at the final step. The vorticity variable $\omega$ is presented here.}
    \label{fig:2dns_batch1}
\end{figure}

The visualizations of all baselines and DyMixOp for the vorticity field $\omega$ at the final prediction step of the first and last samples in the test dataset are shown in Fig.~\ref{fig:2dns_batch0} and Fig.~\ref{fig:2dns_batch1}, respectively. Each panel displays (top row) the predicted solution field (the ground truth) and (bottom row) the absolute error distribution $|\omega_{\text{pred}} - \omega_{\text{true}}|$ (or the error evolution), with color intensity indicating magnitude.
The 2D Navier–Stokes equation in vorticity form describes the competition between nonlinear advection—where vorticity is stretched and folded by the self-induced velocity field—and viscous diffusion, which smooths gradients over time. With $\nu = 10^{-5}$, the flow remains nearly inviscid over the simulated interval, permitting the formation of coherent structures such as vortex dipoles, filaments, and small-scale intermittency, while still allowing gradual enstrophy dissipation at the smallest resolved scales. The ground-truth fields exhibit rich multi-scale dynamics: elongated vorticity filaments, tightly wound cores (bright yellow/red spots), and intricate interfacial layers where sign changes occur—features characteristic of 2D turbulence under weak dissipation.

The ground truth displays thin, high-contrast vorticity filaments that connect or separate regions of opposite sign (e.g., Fig.~\ref{fig:2dns_batch0}, ground truth: green–yellow ribbons embedded in purple background, with sharp transitions). DyMixOp, FNO, and LocalNO reproduce these filaments with correct topology, curvature, and sign distribution: the relative locations of positive and negative patches match closely, and fine-scale connectivity is maintained. In contrast, DeepONet and GNOT produce overly smoothed fields—e.g., DeepONet yields near-uniform teal with no discernible filaments; GNOT shows only broad, low-amplitude undulations and entirely fails to predict the dynamics.
The error maps reveal that models failing to resolve sharp gradients (LaMO, UNO, CoDANO, ConvLSTM) concentrate errors precisely along the thin vorticity interfaces and near vortex cores—regions where $|\nabla \omega|$ is largest. For instance, in Fig.~\ref{fig:2dns_batch0}, CoDANO’s error map shows intense red streaks aligned with the filament boundaries, suggesting inaccurate advection of vorticity contours. DyMixOp and LocalNO, by contrast, exhibit minimal error in these critical regions. Their residual errors are predominantly dark (near-zero) along interfaces, with only faint red patches in low-gradient background zones—consistent with faithful enforcement of the transport term $(\mathbf{u} \cdot \nabla)\omega$.
Comparing the two samples, the qualitative performance ranking remains stable. DyMixOp consistently yields the most structurally faithful predictions; DeepONet and GNOT remain featureless; LocalNO maintains intermediate fidelity with slightly more diffusion than DyMixOp. Notably, DyMixOp’s error magnitude is markedly lower (MSE $\sim 5.4\times10^{-6}$) than all others (most > $10^{-3}$), and its error map is visually darkest—suggesting superior resolution of the nonlinear advective coupling between vorticity and velocity, which is central to 2D NS dynamics.

In summary, the figures demonstrate that accurate modeling of 2D Navier–Stokes vorticity evolution requires architectures capable of resolving sharp vorticity gradients, sign-conserving transport, and multi-scale filamentation without artificial smoothing. The superior visual fidelity of DyMixOp, particularly its retention of fine-scale coherent structures and minimal interface-localized error, is physically consistent with a model that better approximates the nonlocal, velocity-mediated advection inherent in the vorticity equation.

\subsubsection{3D Shallow Water}
\begin{figure}[htbp]
    \centering
    \includegraphics[width=0.98\linewidth]{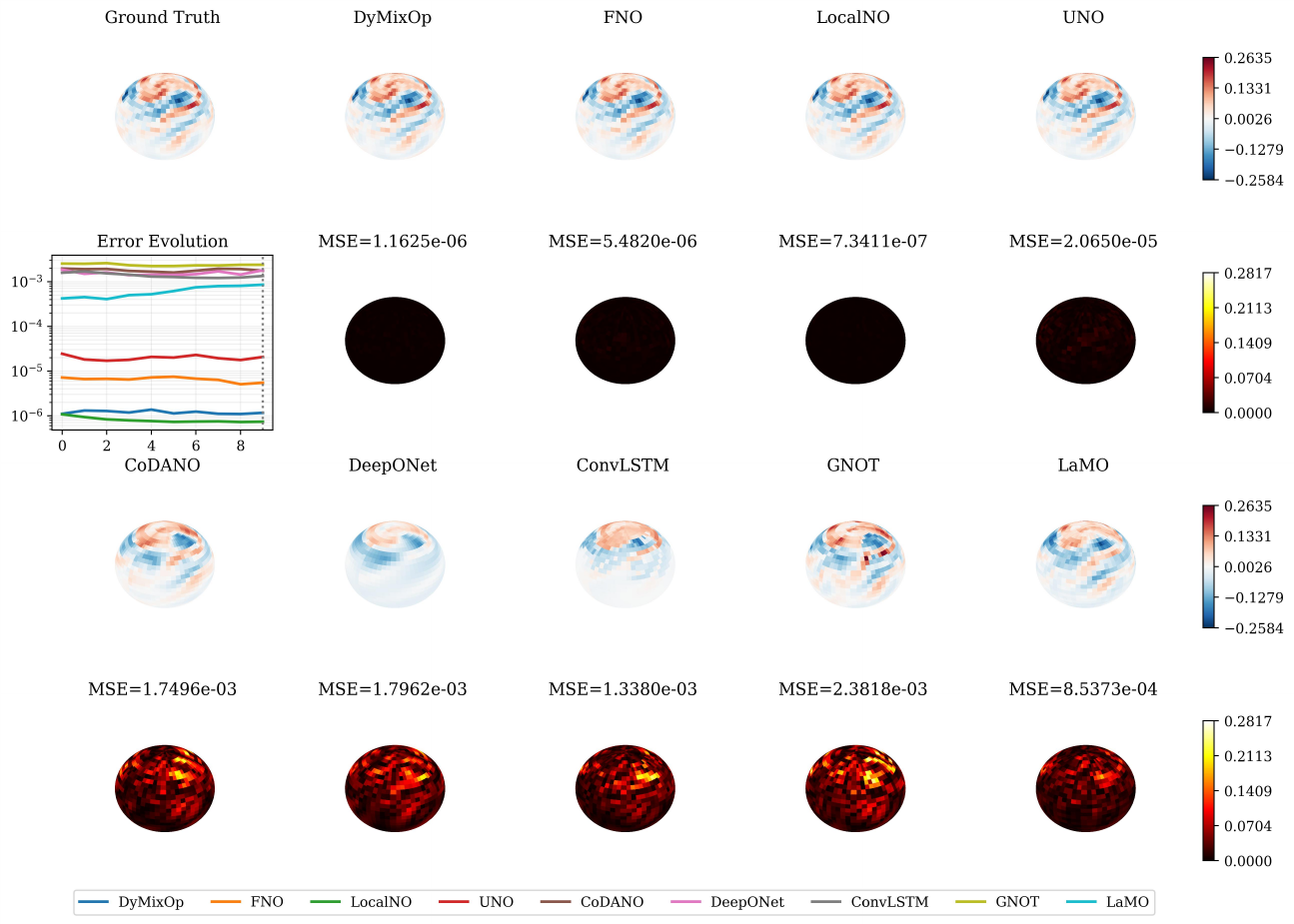}
    \caption{Visual comparison on 3D Shallow Water for the first sample at the final step. The vorticity variable $\omega$ is presented here.}
    \label{fig:3dsw_batch0}
\end{figure}

\begin{figure}[htbp]
    \centering
    \includegraphics[width=0.98\linewidth]{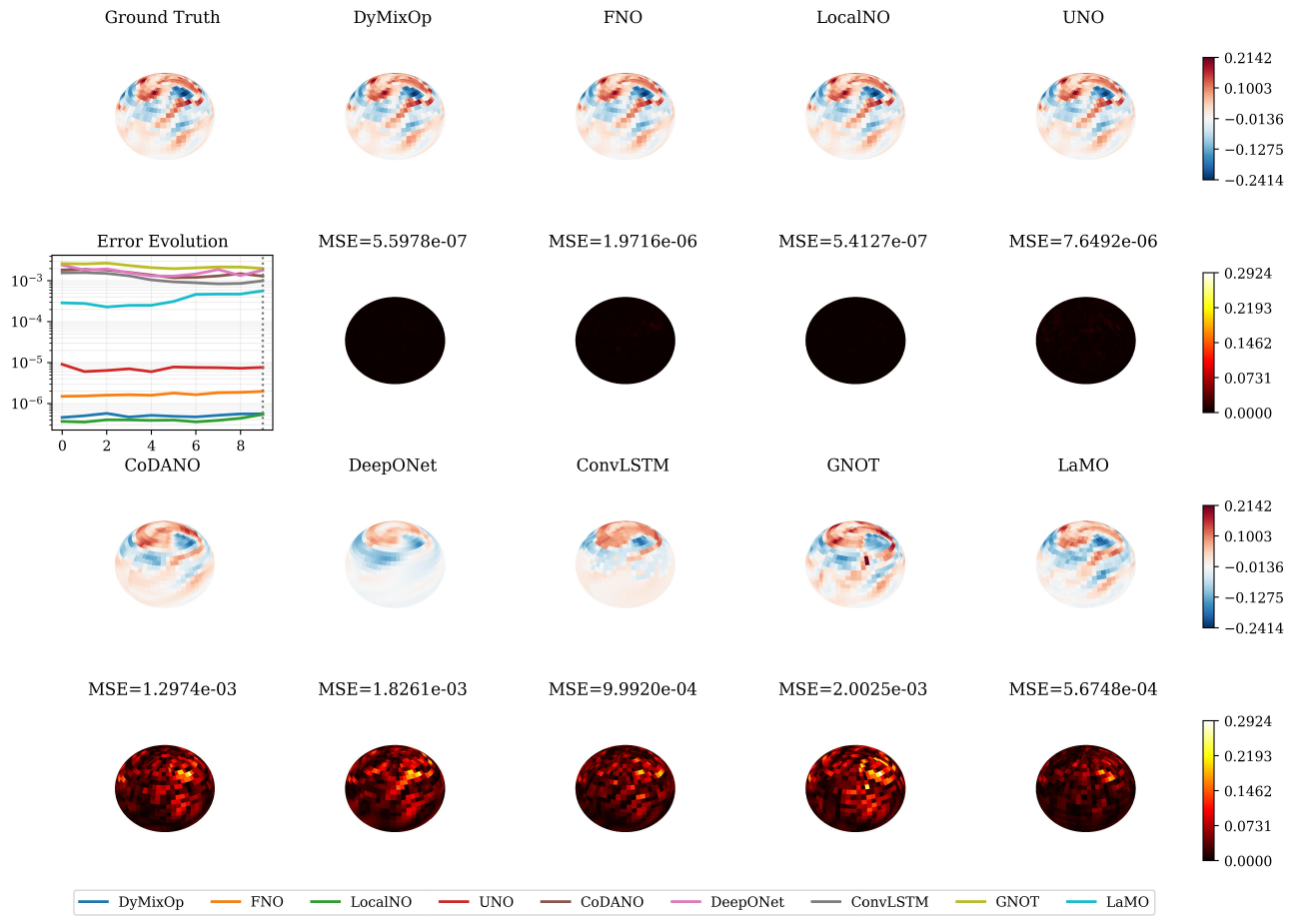}
    \caption{Visual comparison on 3D Shallow Water for the last sample at the final step. The vorticity variable $\omega$ is presented here.}
    \label{fig:3dsw_batch1}
\end{figure}

The visualizations of all baselines and DyMixOp for the vorticity field $\omega$ at the final prediction step of the first and last samples in the test dataset are shown in Fig.~\ref{fig:3dsw_batch0} and Fig.~\ref{fig:3dsw_batch1}, respectively. Each panel displays (top row) the predicted solution field (the ground truth) and (bottom row) the absolute error distribution $|\omega_{\text{pred}} - \omega_{\text{true}}|$ (or the error evolution), with color intensity indicating magnitude. 
The 3D shallow-water system models barotropically unstable mid-latitude jets on a rotating sphere. 
In this mapping, the initial condition consists of a zonally symmetric jet (implied by the smooth meridional shear in the input vorticity) perturbed by a localized Gaussian disturbance, which triggers Rossby wave radiation and nonlinear roll-up into coherent eddies—a process governed by the competition between planetary vorticity gradient ($\beta$-effect), advection, and weak hyperdiffusion. 
The ground-truth solutions exhibit characteristic features of barotropic instability: (i) a dominant dipolar or tripolar structure aligned along the jet core (mid-latitudes), (ii) westward-propagating Rossby wave trains extending into the subtropics and polar regions, and (iii) fine-scale filamentation near the edges of the main vortices, reflecting the cascade of enstrophy toward smaller scales under weak dissipation.

CoDANO, LaMO, DeepONet, GNOT, and ConvLSTM produce qualitatively smooth vorticity fields that retain the large-scale sign distribution but lack the fine-scale filamentary structures present in the ground truth. Their error maps reveal that high-error regions concentrate along the boundaries between red and blue lobes, indicating under-resolved vorticity gradients; UNO show moderate fidelity but introduce spurious small-scale oscillations (e.g., Fig.~\ref{fig:3dsw_batch0}, UNO: checkerboard-like artifacts in mid-latitudes)
DyMixOp, FNO, and LocalNO reproduce this global dipole structure accurately—including the relative amplitude, latitudinal extent, and longitudinal phasing of the wave trains.
Comparing the two samples, the qualitative behavior of each model remains consistent. DyMixOp, FNO, UNO and LocalNO maintains structural coherence across both; Other models remain featureless regardless of initial perturbation scale.

In summary, the figures indicate that successful modeling of 3D shallow-water dynamics requires architectures capable of resolving large-scale balanced structures, Rossby wave dispersion, and nonlinear vortex roll-up without artificial symmetry imposition or excessive diffusion. DyMixOp’s superior visual agreement—with preserved dipole asymmetry, correct wave train geometry, and minimal interface-localized error—is physically consistent with a representation that respects the nonlocal, rotation-dominated nature of barotropic instability on the sphere.

\subsubsection{3D Brusselator}
\begin{figure}[htbp]
    \centering
    \includegraphics[width=0.98\linewidth]{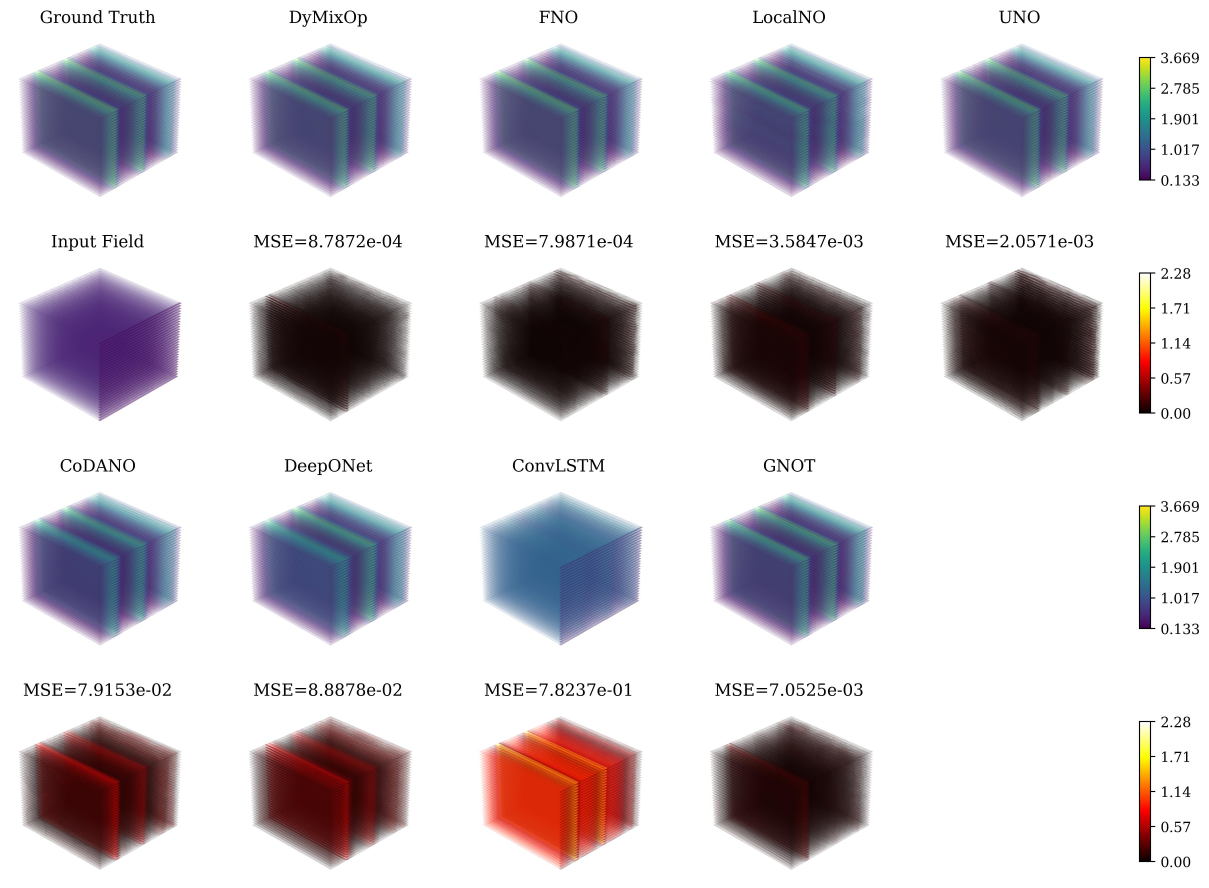}
    \caption{Visual comparison on 3D Shallow Water for the first sample at the final step. The spatiotemporal trajectory of concentration $u$ is presented here.}
    \label{fig:3dbrusselator_batch0}
\end{figure}

\begin{figure}[htbp]
    \centering
    \includegraphics[width=0.98\linewidth]{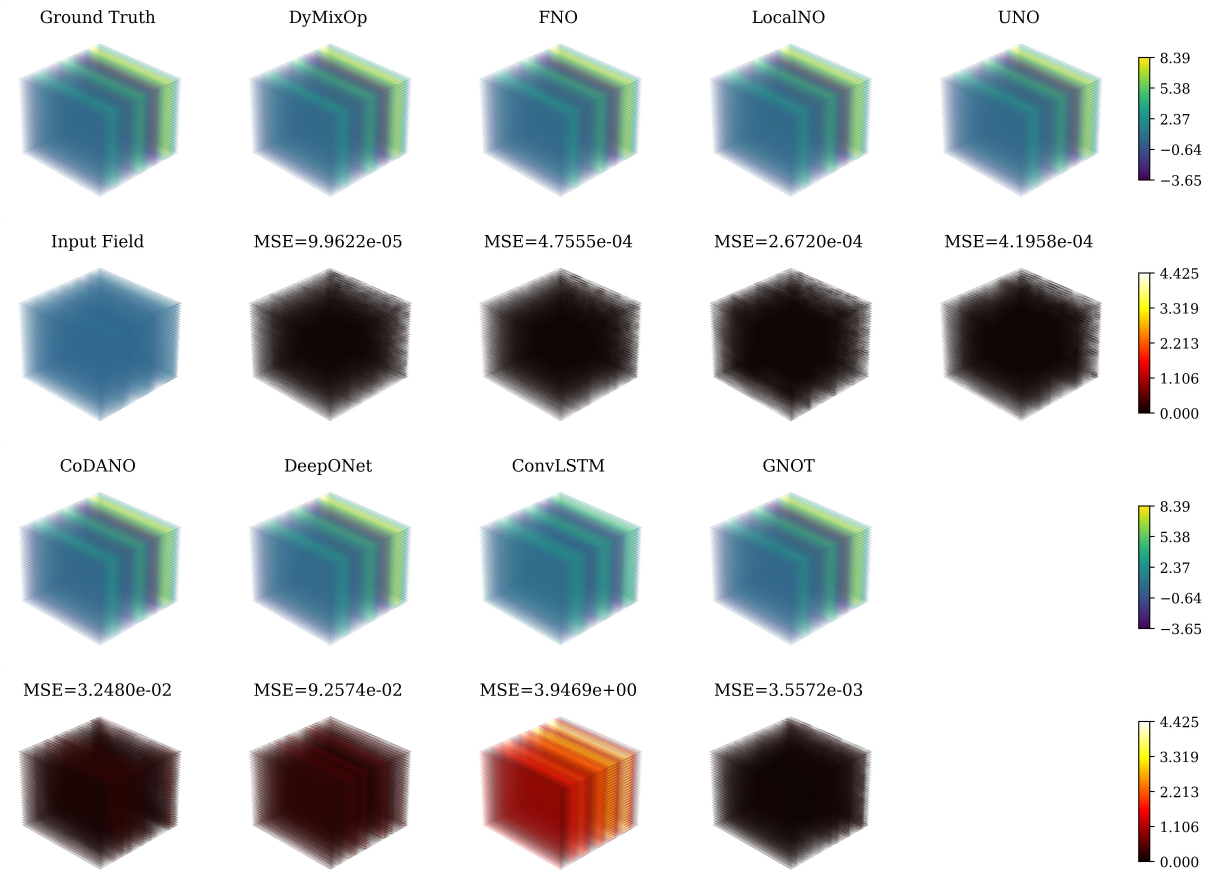}
    \caption{Visual comparison on 3D Shallow Water for the last sample at the final step. The spatiotemporal trajectory of concentration $u$ is presented here.}
    \label{fig:3dbrusselator_batch1}
\end{figure}

The visualizations of all baselines and DyMixOp for the concentration field $u(\mathbf{x}, t)$ of the first and last samples in the test dataset are shown in Fig.~\ref{fig:3dbrusselator_batch0} and Fig.~\ref{fig:3dbrusselator_batch1}, respectively. Each panel presents a 3D isosurface rendering of the spatiotemporal field, with color encoding the scalar value of $u$. The top row shows the the ground truth (the prediction) and the bottom row the initial condition (the absolute error $|u_{\text{pred}} - u_{\text{true}}|$ where red denotes large local deviations).
The 3D Brusselator system models an autocatalytic reaction–diffusion process with $D_0 > D_1$, implying that species $u$ diffuses faster than $v$. This disparity enables Turing-type pattern formation: small spatial perturbations in $v$ (introduced via $\epsilon(\mathbf{x})$) are amplified through the nonlinear feedback $u^2 v$, while differential diffusion stabilizes emerging structures against homogeneous decay. The time-dependent forcing $f(t)$ modulates the base production rate, inducing transient oscillations and influencing the phase and amplitude of emerging patterns. The ground-truth solutions exhibit characteristic striped or layered spatiotemporal structures—i.e., quasi-1D bands aligned along one spatial axis, with smooth transitions in $u$ across bands and clear temporal evolution in band spacing and intensity—consistent with known Turing instabilities in the Brusselator under periodic boundary conditions.

The ground truth displays well-defined, approximately parallel bands of alternating high/low $u$ concentration, extending uniformly in the temporal dimension (visible as coherent “sheets” in the 3D renderings). DyMixOp, FNO, UNO,and LocalNO reproduce this banding structure with correct orientation, spacing, and continuity across time; the bands remain straight, evenly spaced, and retain their lateral coherence. In contrast, ConvLSTM produce highly smoothed or distorted fields and generates blocky, stair-like artifacts (especially evident in Fig.~\ref{fig:3dbrusselator_batch1}), suggesting failure to resolve the spatially coherent, diffusion-mediated pattern selection inherent in Turing systems.
The error maps reveal that models with poor pattern fidelity (DeepONet, CoDANO) concentrate errors precisely along the sharp transitions between bands—regions where $\|\nabla u\|$ is maximal and where the reaction–diffusion balance is most sensitive. For instance, in Fig.~\ref{fig:3dbrusselator_batch0}, CoDANO’s error map shows intense red sheets coinciding with the inter-band boundaries, indicating inaccurate modeling of the nonlinear coupling $u^2 v$ and/or mismatched effective diffusion. DyMixOp, FNO and LocalNO, by contrast, exhibit minimal error along these interfaces; their residuals are predominantly dark (near-zero) in band interiors. They preserve the temporal coherence that their predicted bands maintain alignment across the time axis. ConvLSTM, however, show phase drift or band dislocation (e.g., Fig.~\ref{fig:3dbrusselator_batch1}, implying incorrect coupling between spatial and temporal dynamics—likely due to insufficient modeling of the nonlocal memory induced by diffusion and the time-dependent forcing $f(t)$.
Comparing the two test samples, the qualitative performance ranking remains stable. DyMixOp consistently yields the structurally faithful predictions.

In summary, the figures demonstrate that accurate modeling of the 3D Brusselator requires architectures capable of resolving diffusion-driven spatial coherence, nonlinear reaction-induced band selection, and temporally consistent phase evolution. DyMixOp’s superior visual fidelity is physically consistent with a representation that respects the underlying Turing mechanism, where pattern formation arises from the interplay of local reaction kinetics and nonlocal diffusion.

\subsection{Zero-shot Super-resolution Results}
Zero-shot super-resolution evaluates whether a neural operator learned on a \,\emph{single} training discretization can be applied to a \,\emph{different} (typically finer) mesh at test time, without any additional finetuning. This capability is closely related to the defining property of neural operators---learning mappings between \,\emph{function spaces} rather than between fixed-dimensional vectors (e.g. \emph{mesh/discretization invariance}). In practice, we train each model on a base resolution and then directly query the learned operator on progressively refined grids, by providing the input field sampled on the target grid.

Table~\ref{tab:unified-results-refined} summarizes the relative MSE under several refinement factors for representative PDEs. Following the same set of dynamics-informed architectural variants used in our ablation study (i.e., Adaptive Evolution, Parallel, and Hybrid), we report their zero-shot super-resolution performance under both settings \emph{with} and \emph{without} differential kernels (marked as ``diff.''). The spectral FNO baseline exhibits limited zero-shot capability to unseen resolutions: its error remains high for the KS system and increases with refinement for nontrivial geometries (e.g., Darcy), reflecting a discretization mismatch between training and testing grids. In contrast, DyMixOp variants without differential kernels maintain consistently low errors across refinement levels, demonstrating strong mesh invariance. Notably, although introducing differential kernels can degrade zero-shot performance (especially when used in multiple consecutive layers), our proposed components---the local--global mixing (LGM) transformations, the inertial-manifold-inspired latent construction, and the dynamics-informed aggregation module (DyIA)---largely preserve the discretization robustness and still outperform FNO in most cases.

The performance drop caused by differential kernels aligns with the observation in the original differential/integral-kernel neural operator work~\cite{liu2024neural}: differential layers may overfit to the training discretization, incur discretization errors when the training resolution does not sufficiently resolve local details, and such errors can rapidly propagate when differential convolutions are stacked. This suggests that differential operators typically require higher-resolution training data (or more careful placement of differential layers) to realize their potential in super-resolution, whereas DyMixOp's multiplicative local--global mixing and dynamics-consistent design provide a more robust route to mesh-invariant generalization.
\begin{table}[htbp]
  \centering
  \small
  \caption{\textbf{Zero-shot super-resolution results.} Relative MSE when a model trained on a base discretization is directly evaluated on finer meshes (no finetuning). ``Mapping'' reports the refinement factor and the corresponding target resolution. We compare FNO with DyMixOp variants that adopt the same dynamics-informed architectures as in the ablation study (Adaptive Evolution / Parallel / Hybrid), and ``diff.'' denotes the presence of differential kernels. Best results are \textbf{bolded}.}
  \label{tab:unified-results-refined}
  \vspace{2mm}
  \resizebox{\textwidth}{!}{
  \begin{tabular}{l c c c c c}
    \toprule
    \multirow{2}{*}{\textbf{Baselines}} & \multicolumn{5}{c}{\textbf{Resolution}} \\
    \cmidrule(lr){2-6}
    & \textbf{1} & \textbf{2} & \textbf{3} & \textbf{4} & \textbf{5} \\
    \midrule

    \rowcolor[gray]{0.9} \multicolumn{6}{l}{\textbf{1D Kuramoto-Sivashinsky (KS)}} \\
    \textit{Mapping} & $\times 1.0 \to (256)$ & $\times 2.0 \to (512)$ & $\times 4.0 \to (1024)$ & $\times 8.0 \to (2048)$ & $\times 16.0 \to (4096)$ \\
    \cmidrule(lr){1-6}
    FNO & 0.4192 & 0.4231 & 0.4208 & 0.4206 & 0.4205 \\
    Adaptive Evolution (w/o-diff.) & \textbf{0.0088} & \textbf{0.0102} & \textbf{0.0102} & \textbf{0.0102} & \textbf{0.0104} \\
    Parallel Arch. (w/o-diff.) & 0.0099 & 0.0114 & 0.0114 & 0.0114 & 0.0115 \\
    Hybrid Arch. (w/o-diff.) & 0.0424 & 0.0448 & 0.0447 & 0.0447 & 0.0448 \\
    Hybrd Arch. (w/-diff.) & 0.0252 & 1.0963 & 0.8463 & 0.8579 & 0.8420 \\
    \addlinespace[3mm]

    \rowcolor[gray]{0.9} \multicolumn{6}{l}{\textbf{2D Darcy Flow}} \\
    \textit{Mapping} & $\times 1.0 \to (49^2)$ & $\approx \times 1.3 \to (121^2)$ & $\approx \times 1.7 \to (81^2)$ & $\approx \times 2.5 \to (121^2)$ & $\approx \times 4.9 \to (241^2)$ \\
    \cmidrule(lr){1-6}
    FNO & 7.24e-5 & 1.45e-4 & 3.79e-4 & 7.66e-4 & 1.30e-3 \\
    Adaptive Evolution (w/o-diff.) & \textbf{4.95e-5} & \textbf{1.23e-4} & \textbf{3.43e-4} & \textbf{7.03e-4} & \textbf{1.19e-3} \\
    Parallel Arch. (w/o-diff.) & 6.62e-5 & 1.38e-4 & 3.60e-4 & 7.22e-4 & 1.22e-3 \\
    Hybrid Arch. (w/o-diff.) & 6.64e-5 & 1.39e-4 & 3.62e-4 & 7.25e-4 & 1.22e-3 \\
    Hybrd Arch. (w/-diff.) & 5.75e-5 & 8.79e-4 & 3.87e-3 & 1.01e-2 & 2.09e-2 \\
    \addlinespace[3mm]

    \rowcolor[gray]{0.9} \multicolumn{6}{l}{\textbf{2D CE-CRP}} \\
    \textit{Mapping} & $\times 1.0 \to (64^2)$ & $\times 2.0 \to (128^2)$ & -- & -- & -- \\
    \cmidrule(lr){1-6}
    FNO & 0.0349 & \textbf{0.0954} & -- & -- & -- \\
    Adaptive Evolution (w/o-diff.) & 0.0350 & 0.1016 & -- & -- & -- \\
    Parallel Arch. (w/o-diff.) & 0.0235 & 0.1022 & -- & -- & -- \\
    Hybrd Arch. (w/o-diff.) & 0.0372 & 0.0933 & -- & -- & -- \\
    Hybrd Arch. (w/-diff.) & \textbf{0.0147} & 0.1228 & -- & -- & -- \\
    \bottomrule
  \end{tabular}
  }
\end{table}

\newpage
\subsection{Limitation and discussion}
\textbf{Universal Approximation Theorem}
While our method is firmly grounded in inertial manifold theory, offering a principled reduction of infinite-dimensional PDE dynamics into a finite latent space, we acknowledge a theoretical gap and our proposed architecture does not yet come with a formal UAT guarantee. But we can still have a look at the realizability of this theorem. We embed a sufficiently expressive operator network FNO block inside our latent dynamics loop. Since such blocks are UAT‑capable, the overall architecture would inherit universality in principle. This suggests that, by constructing our latent-update functions and LGM transforms such that they can emulate any continuous mapping on the latent space, the entire operator pipeline could satisfy an approximation theorem in the limit of sufficiently high capacity. Similar reasoning is often used in works like FNO and DeepONet papers to infer universality from the constituent blocks \cite{kovachki2021universal, lu2019deeponet}.
While empirical performance across varied PDE benchmarks strongly supports its effectiveness, a formal universal approximation guarantee for the complete architecture remains to be established in future theoretical work.

\textbf{Irregular grid}
In this implemented DyMixOp, the global transformation is the parameterized Fourier transformation, which require the uniform grid and may encounter the difficulty of a direct implementation in the irregular grid. However, there still exits some methods to solve this problem in the existing literature, for example, transforming the irregular grid in the physical space to the regular grid in the computational space \cite{li2023fourier, li2023geometry}. Alternatively, the global transformation can be specified as the transformer architecture, which regards the input as the point cloud, then DyMixOp can naturally deal with the irregular grid problems.


\newpage
\bibliographystyle{plain}
\bibliography{references}
\newpage